\documentclass[journal]{IEEEtran}
\usepackage{epsfig}
\usepackage{graphicx}
\usepackage{amsmath}
\usepackage{amssymb}
\usepackage{algorithm, algorithmic}

\usepackage{diagbox}
\usepackage{float}
\usepackage{afterpage}
\usepackage{bm}
\usepackage{subfig}

\usepackage{multirow}
\usepackage{color}
\usepackage{tablefootnote}
\usepackage{adjustbox}
\usepackage{soul}

\usepackage{xcolor,colortbl}

\usepackage[pagebackref=true,breaklinks=true,colorlinks,bookmarks=false]{hyperref}

\newcommand{\rev}[1]{\textcolor{black}{#1}}
\newcommand{\revf}[1]{\textcolor{black}{#1}}

\def\etal{\textit{et al}.}
\def\ie{\textit{i.e.}}
\def\eg{\textit{e.g.}}

\begin{document}

\title{Self-supervised Matting-specific\\ Portrait Enhancement and Generation}

\author{{Yangyang~Xu,~Zeyang Zhou, and~Shengfeng He,~\IEEEmembership{Senior Member,~IEEE}}

\thanks{This project is supported by the National Natural Science Foundation of China (No. 61972162); Guangdong International Science and Technology Cooperation Project (No. 2021A0505030009); Guangdong Natural Science Foundation (No. 2021A1515012625); Guangzhou Basic and Applied Research Project (No. 202102021074); and CCF-Tencent Open Research fund (CCF-Tencent RAGR20210114). Corresponding author: Shengfeng He.}
\thanks{Yangyang Xu, Zeyang Zhou, and Shengfeng He are with the School of Computer Science and Engineering, South China University of Technology, Guangzhou, China. E-mail: cnnlstm@gmail.com, zzygit@gmail.com, hesfe@scut.edu.cn.}}

\markboth{IEEE TRANSACTIONS ON IMAGE PROCESSING}%
{Shell \MakeLowercase{\textit{Xu et al.}}: Self-supervised Matting-specific Portrait Enhancement and Generation}

\IEEEtitleabstractindextext{%
\begin{abstract}
We resolve the ill-posed alpha matting problem from a completely different perspective. Given an input portrait image, instead of estimating the corresponding alpha matte, we focus on the other end, to subtly enhance this input so that the alpha matte can be easily estimated by any existing matting models. This is accomplished by exploring the latent space of GAN models. It is demonstrated that interpretable directions can be found in the latent space and they correspond to semantic image transformations. We further explore this property in alpha matting. Particularly, we invert an input portrait into the latent code of StyleGAN, and our aim is to discover whether there is an enhanced version in the latent space which is more compatible with a reference matting model. We optimize multi-scale latent vectors in the latent spaces under four tailored losses, ensuring matting-specificity and subtle modifications on the portrait. We demonstrate that the proposed method can refine real portrait images for arbitrary matting models, boosting the performance of automatic alpha matting by a large margin. In addition, we leverage the generative property of StyleGAN, and propose to generate enhanced portrait data which can be treated as the pseudo GT. It addresses the problem of expensive alpha matte annotation, further augmenting the matting performance of existing models. Code is available at~\url{https://github.com/cnnlstm/StyleGAN_Matting}.
\end{abstract}

}

\maketitle

\IEEEdisplaynontitleabstractindextext

\IEEEpeerreviewmaketitle


{\section{Introduction}}
Self-portrait photographs, \ie, selfies, have been popular on social media, and its related portrait enhancement applications are widely developed to fulfill different aesthetic requirements. A fundamental technique that drives these applications is image matting, which is to estimate the foreground transparency of portrait images. Image matting assumes an image is composed of the linear combination of the foreground $F$ and background $B$ with the alpha matte $\alpha$ indicating the foreground transparency. It can be formulated as
\begin{equation}
   I_{i}= \alpha _{i} F_{i}+ (1-\alpha _{i})B_{i},  \alpha _{i} \in \left [ 0, 1\right ],
   \label{eq1}
\end{equation}
where $i = (x, y)$ denotes the pixel position in the image.

It can be seen that image matting is an ill-posed problem given only 3 known values but 7 values need to be solved for each pixel. Therefore, most matting methods~\cite{4547428,aksoy2017designing,yang2018active,li2020natural,xu2017deep,chen2018semantic} introduce an additional user-specified scribble or trimap to reduce the dimension of the solution space. This kind of auxiliary input is a confidence map that gives each pixel of the image a probability of belonging to the foreground, background or a mixed area. These works have been proven to achieve a reliable result. However, they highly rely on a well-defined trimap, and this precise prior knowledge is always hard to obtain. Their results deteriorate rapidly if the trimap is defective, which is a practical problem in real-world scenarios. For this reason, automatic portrait matting has been proposed in recent years~\cite{shen2016deep,xu2017deep,orrite2019portrait,chen2018semantic}.

\begin{figure}[t]
\centering
\subfloat[Original]{\label{fig:teaser_a}
  \begin{minipage}{0.2\linewidth}
  \includegraphics[width=\linewidth]{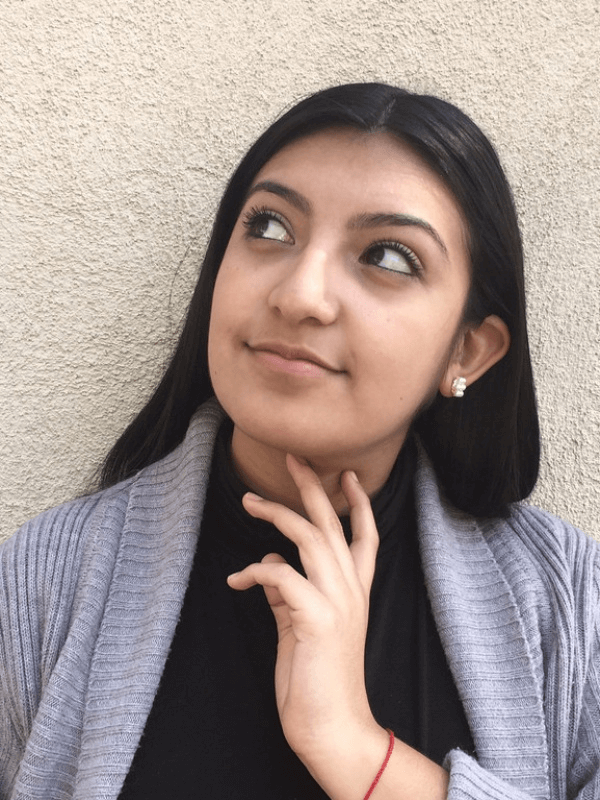}
  \includegraphics[width=\linewidth]{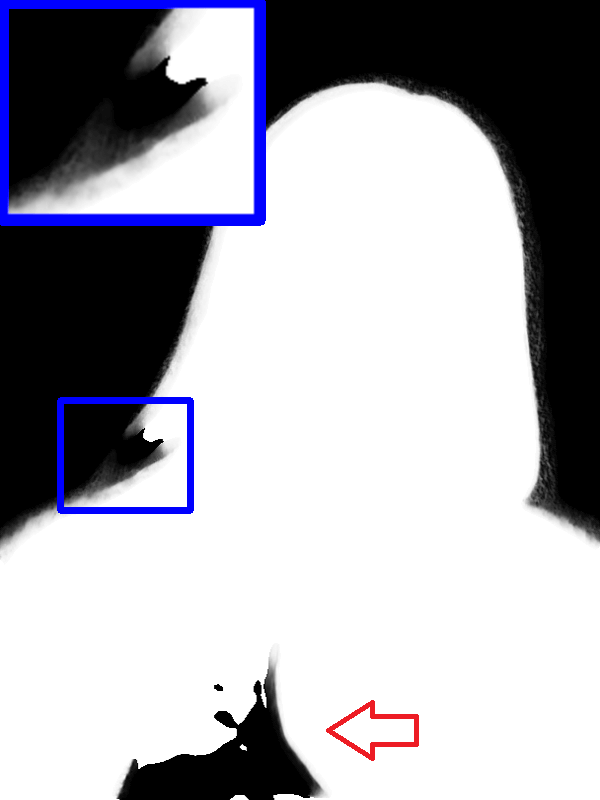}
  \end{minipage}\hspace{-2mm}
  }
\subfloat[Enhanced]{\label{fig:teaser_b}
  \begin{minipage}{0.2\linewidth}
  \includegraphics[width=\linewidth]{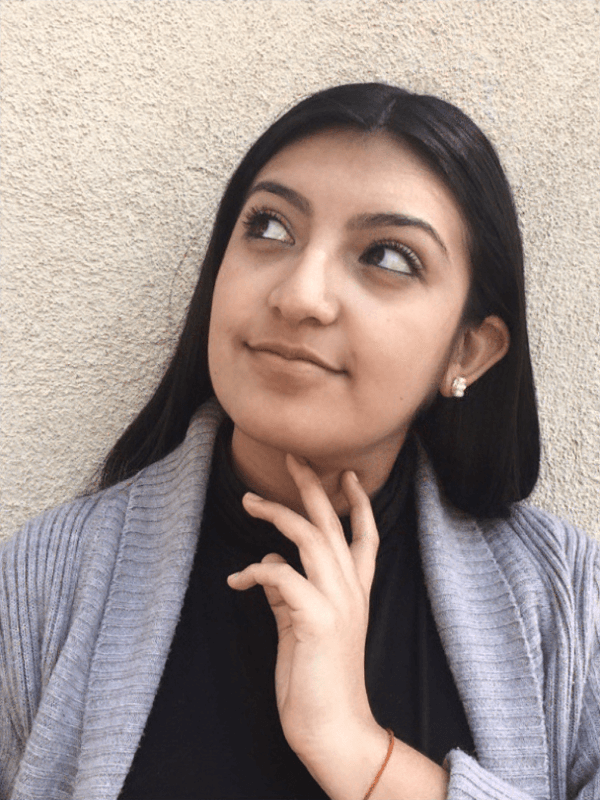}
  \includegraphics[width=\linewidth]{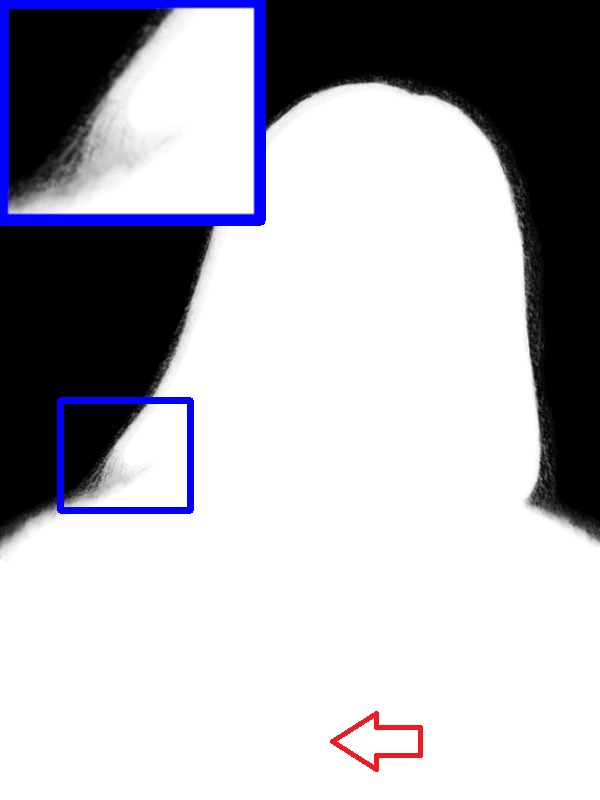}
  \end{minipage}\hspace{-2mm}
}
\subfloat[{Pert.\& GT}\vspace{-3mm}]{\label{fig:teaser_c}
  \begin{minipage}{0.2\linewidth}
  \includegraphics[width=\linewidth]{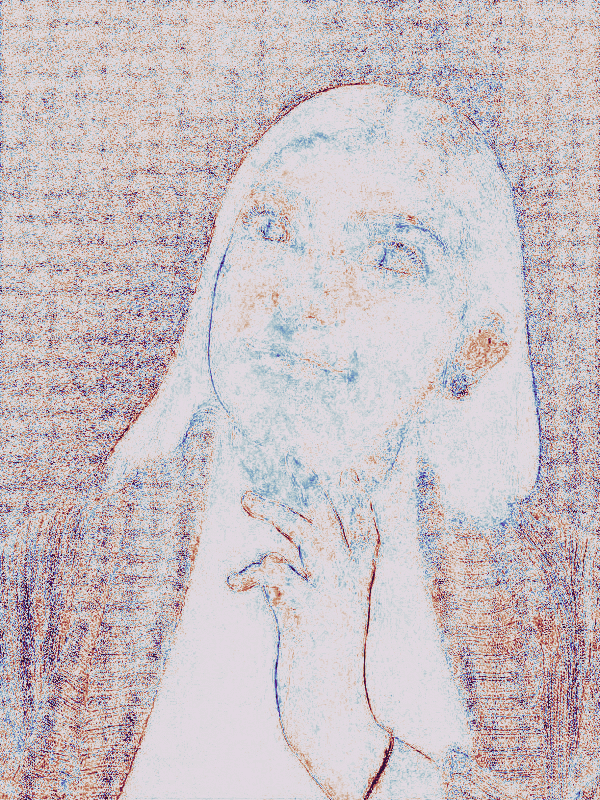}
  \includegraphics[width=\linewidth]{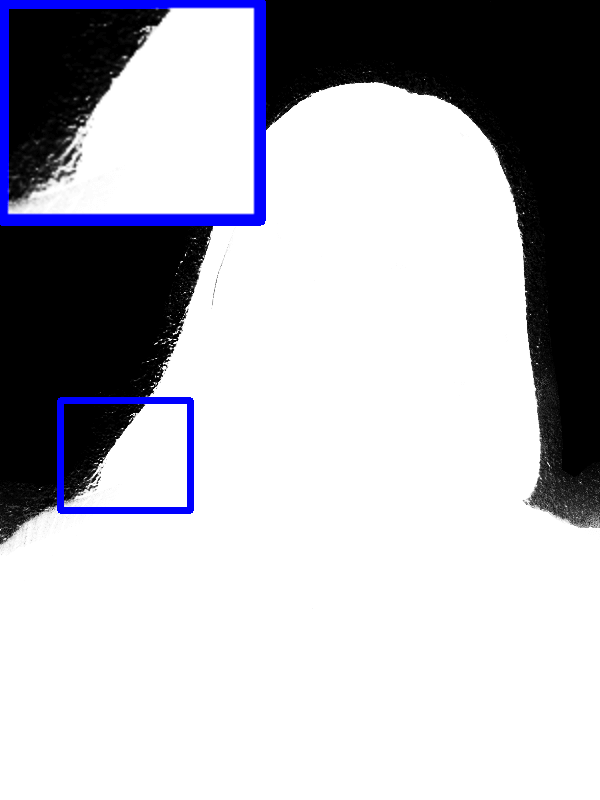}
  \end{minipage}
  }
\subfloat[Synthesized Data]{
  \begin{minipage}{0.2\linewidth}
  \includegraphics[width=\linewidth]{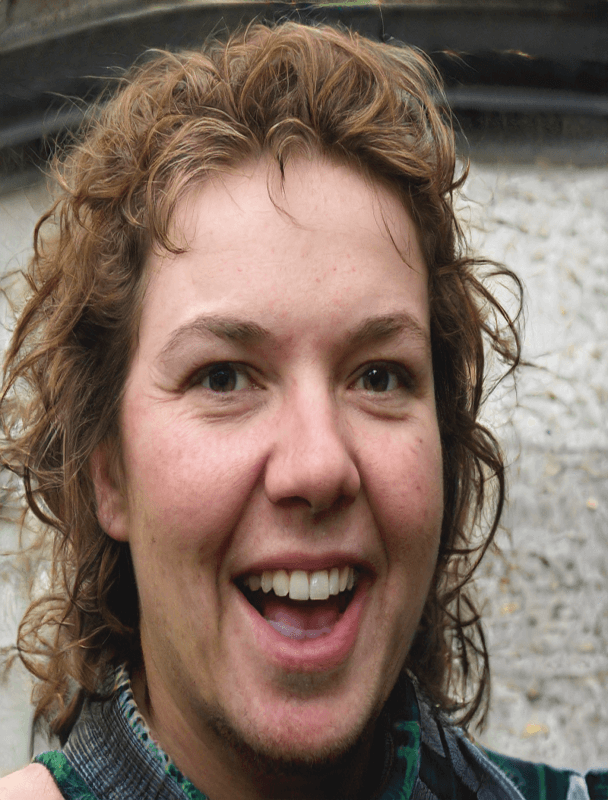}
  \includegraphics[width=\linewidth]{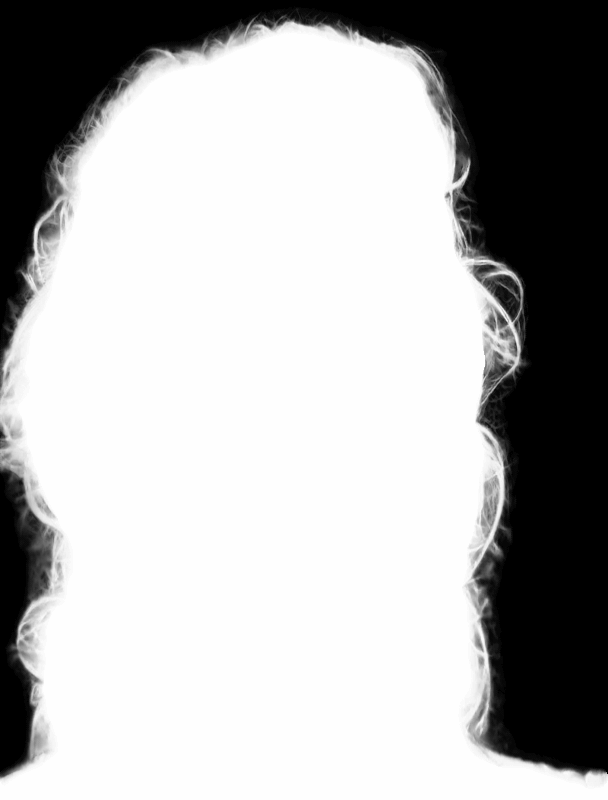}
  \end{minipage}
  \hspace{-2mm}
  \begin{minipage}{0.2\linewidth}
  \includegraphics[width=\linewidth]{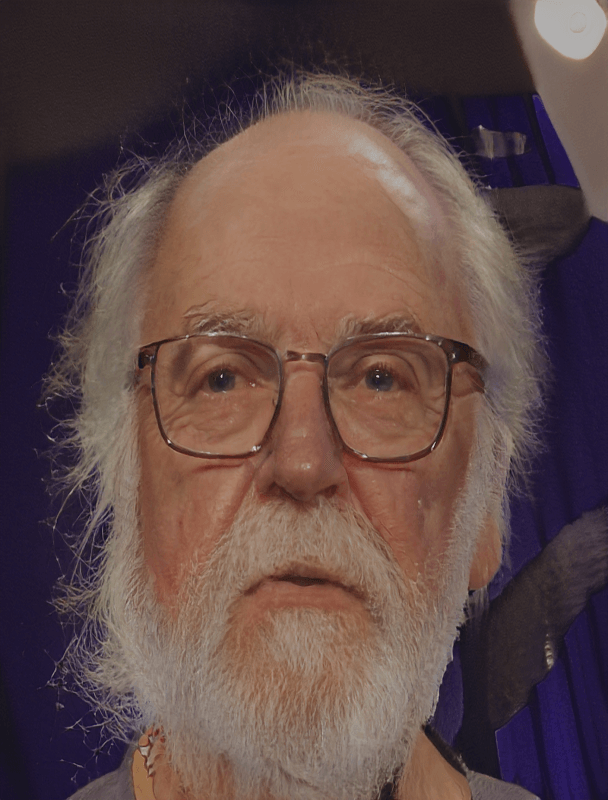}
  \includegraphics[width=\linewidth]{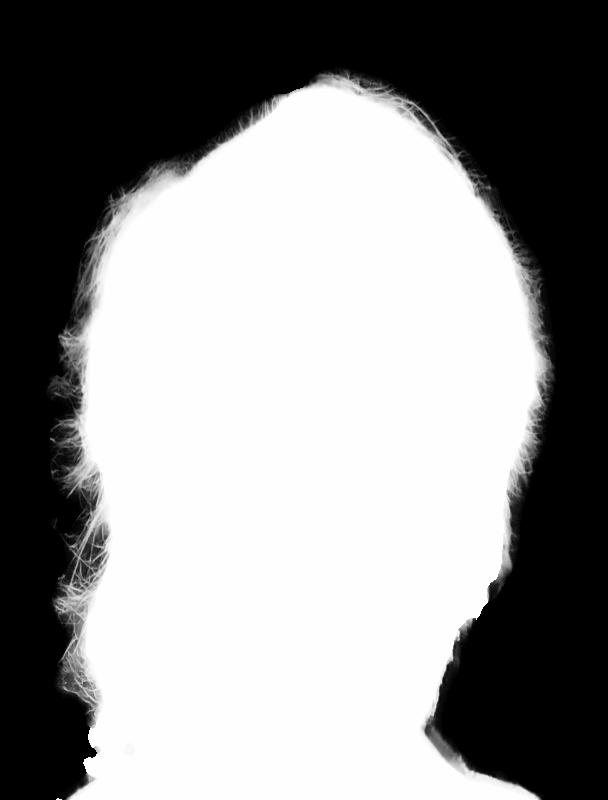}
  \end{minipage}
  }
\caption{We propose to generate matting-specific portraits by exploring the latent space of GANs. (a), (b) show that subtle changes in the portrait can significantly enhance the output alpha matte (see blue boxes and red arrows). (c) shows the learned noise perturbations (rescaled and remapped) are automatically attached in high-frequency image contents so that they are not sensitive to human perception. (d) shows that the proposed method can be also used to synthesize pseudo GT that augments existing matting models.}
\label{fig:teaser}
\end{figure}

Notwithstanding the demonstrated success, automatic portrait matting still suffers from ambiguous semantic and discrepant alpha values. These problems can be hardly addressed without prior knowledge due to the entanglement between foreground and background colors. However, we observe that foreground and background may be disentangled if small changes in the image are introduced (see Fig.~\ref{fig:teaser_a}, \ref{fig:teaser_b}, and \ref{fig:teaser_c}), as the alpha matte is estimated from a global perspective. This finding is inspired by sampling-based methods~\cite{he2011global,tang2019learning}, as changing the foreground or background slightly may provide a completely different set of support samples for alpha value estimation.

Based on the above observation, we propose an alternative solution for resolving the ill-posed matting problem. Unlike traditional image matting that devoted to alpha matte prediction, we focus on how an input image affects the output alpha matte. However, connecting image editing with alpha matting is not trivial. Naively cascading two modules lacks interpretable control of image editing, and therefore cannot guarantee subtle changes applied in the portrait. In addition, this requires a conditional Generative Adversarial Network (GAN) structure which cannot reconstruct high-resolution and high-fidelity portrait.

In this paper, we achieve matting-specific portrait enhancement by exploring the latent space of GAN models. Very recent literature~\cite{shen2019interpreting,Plumerault2020ControllingGM,voynov2020unsupervised} show that meaningful directions can be discovered in the latent space of GAN models that correspond to human interpretable image transformation. Except for editing randomly generated images, latest works~\cite{abdal2019image2stylegan,abdal2019image2stylegan2,gu2019image,zhu2020domain} propose to invert a real image back to the latent code of the GAN model for reconstruction. All these prior works motivate us to explore the following question: is there an enhanced portrait that exists in the latent space that is more compatible with a reference matting model? However, the main barrier is that there is no labeled data for alpha matting to regress to during testing. Our principle is to convert the regression problem to a classification problem so that it can be optimized in a self-supervised manner. To this end, we first decompose the intractable automatic portrait matting problem into two parts, ternary trimap prediction and alpha matte estimation. The former serves as an optimizer while the latter serves as an assessor. In this way, we can optimize the latent code governed by a reference matting model with four tailored losses, and our objective is to discover a new latent code that minimizes the prediction entropy and corrects the generative distribution in a self-supervised manner. To ensure matting-specificity as well as subtle modifications on the portrait, we search the optimal latent code in the noise latent space $\mathcal{N}$ and the intermediate latent space $\mathcal{W+}$ of StyleGAN.

The input latent code can be either inverted from a real image or randomly-generated, each of which enables two different applications (see Fig.~\ref{fig:teaser}). The former one is used to enhance a real portrait for a better matting performance, while the latter can create a massive amount of pseudo GT data for augmenting the performance of arbitrary matting models. Extensive experiments demonstrate that the proposed two alternative solutions can enhance a portrait with minimum modifications or produce matting-specific portrait data, boosting state-of-the-art models by a large margin.

Our contributions are summarized as follows:
\begin{itemize}
    \item \rev{Different from existing image matting works that predict an alpha matte based on the fixed input image, we focus on how an input image affects the output alpha matte. We seek an enhanced image by exploring the latent space of GAN that is more compatible with a referenced matting model in a self-supervised manner.}
    \item We present a portrait enhancement approach that can refine an input portrait with minimum changes in the image. State-of-the-art models can be boosted by a large margin on two datasets.
    \item We propose to generate matting-specific portrait data, addressing the expensive annotation problem of alpha matte. The synthesized data can augment state-of-the-art matting models.
\end{itemize}

\section{{Related Work}}
\subsection{Natural Image Matting}
Traditional image matting can be categorized into sampling-based and affinity-based approaches. {Sampling-based approaches}~\cite{he2011global,karacan2015image, chuang2001bayesian,wang2007optimized} assume that the alpha values for two neighboring pixels are similar if their colors are similar. Affinity-based approaches~\cite{aksoy2017designing,chen2013knn,levin2007closed} propagate alpha values from the known regions into unknown ones by the affinities of neighboring pixels. Several deep learning-based approaches are proposed for image matting to leverage the powerful representation learning ability. Xu~\etal~\cite{xu2017deep} propose a large scale dataset and the first attempt to apply a neural network for nature image matting. Li~\etal~\cite{li2020natural} present the Guided Contextual Attention to propagate high-level opacity information based on the learned low-level affinity. Lu~\etal~\cite{lu2019indices} develop an IndexNet to generate the indices on the feature map adaptively. SampleNet~\cite{tang2019learning} estimates the background and foreground colors of pixels in unknown regions using deep neural networks prior. Cai~\etal~\cite{cai2019disentangled} aim to refine the trimap by an adaptation module in the matting network. However, these works rely heavily on the user-designed trimap. When the trimap is unavailable, a typical solution is to predict the trimap by a semantic segmentation model~\cite{shen2016deep,chen2018semantic}. However, compared with the user-designed trimap, the predicted trimap is insufficiently accurate. Background matting~\cite{sengupta2020background,lin2021real,deng2022background} can extract alpha mattes without the need of trimap, but it requires an additional input of clean background image, which confines the flexibility. Unlike previous approaches, we propose an alternative solution for automatic portrait matting that focuses on how an input image affects the output alpha matte by exploring the latent space of GAN models.

\subsection{Image Editing via Latent Space Exploration}

Generative models show great potential in synthesizing various images by taking random latent codes as inputs. It's evidenced that a pre-trained generator encodes rich knowledge prior, which benefits many downstream works, such as face swapping, super-resolution, semantic editing, and so on~\cite{zhong2022faithful,xu2022high,zhou2022pro,shen2019interpreting,gu2019image}. Particular, there are many image editing methods explore the latent space of a pre-trained generative model (\eg, StyleGAN) in supervised or unsupervised manners~\cite{shen2019interpreting,gu2019image,zhu2020domain,goetschalckx2019ganalyze,lample2017fader,yang2021discovering}. \rev{Different from existing works that learn the end-to-end models for image enhancement or restoration~\cite{gou2020clearer,li2021you,deng2019deep}, GAN-based editing methods need to invert a target image to the latent code in advance, which is called GAN inversion~\cite{abdal2019image2stylegan,abdal2019image2stylegan2,Xu2021ICCV,wang2022HFGI,tov2021designing}.} GAN inversion works discover a latent code that can produce the closest result to the input real image. In particular, Abdal~\etal~\cite{abdal2019image2stylegan} propose an optimization-based algorithm that optimizes the latent code in the~$\mathcal{W+}$~of a fixed StyleGAN governed by pixel-wise reconstruction. Their subsequent work~\cite{abdal2019image2stylegan2} introduces noise optimization as a complement to the~$\mathcal{W+}$~latent space for restoring the high-frequency details in images. Zhu~\etal~\cite{zhu2020domain} propose to learn an encoder that ensures the inverted code in the semantic domain of the original latent space and explores many semantic image editing tasks. Compared to these works, we have a different objective that aiming at producing matting-specific portrait. We emphasize editing portraits with the minimum changes while making them the most ``predictable'' by a matting model.

\begin{figure}[t]
\centering
\includegraphics[width=0.75\linewidth]{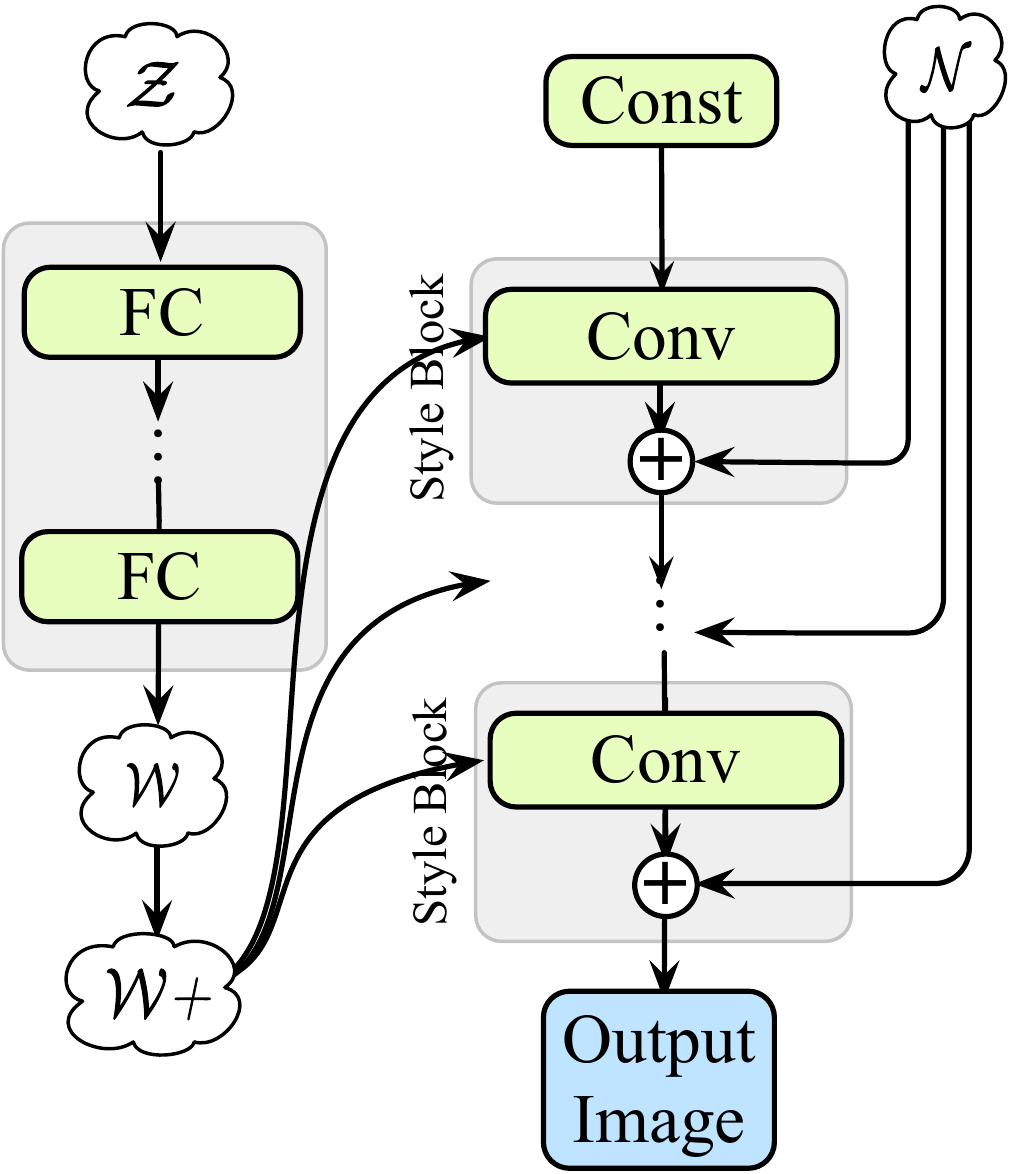}
\caption{\rev{Illustration of different latent spaces in StyleGAN.}}
\label{fig:spaces}
\end{figure}

\section{Methodology}


\subsection{Problem Formulation}
Given an input portrait~$I_o$, {a reference matting model~$M_{\alpha}$, and a pre-trained trimap model~$M_{t}$}, our goal is to obtain an enhanced portrait produced by the generator $G$ of StyleGAN~\cite{karras2019style,karras2019analyzing}. \rev{Unlike traditional GANs, the output of StyleGAN is controlled by two latent codes from $\mathcal{Z} $ and $\mathcal{N}$ spaces. As shown in Fig.~\ref{fig:spaces}, StyleGAN maps the latent code from $Z$ space to intermediate latent space $\mathcal{W}$ space with few fully-connected layers for a better semantic disentanglement~\cite{karras2019style,karras2019analyzing}. Abdal~\etal~\cite{abdal2019image2stylegan} further extend the $\mathcal{W}$ space to $\mathcal{W+}$ space for a better representative ability, and many GAN inversion works invert images in this space~\cite{Xu2021ICCV,wang2022HFGI,tov2021designing}. Besides, noise latent space $\mathcal{N}$ provides high-frequency information to the generator. It produces the noise latent code to each scale of the synthesis network. As shown in~\cite{karras2019style,abdal2019image2stylegan2}, this space controls subtle details of the synthesis.}

\begin{figure*}[t]
\centering
\includegraphics[width=\linewidth]{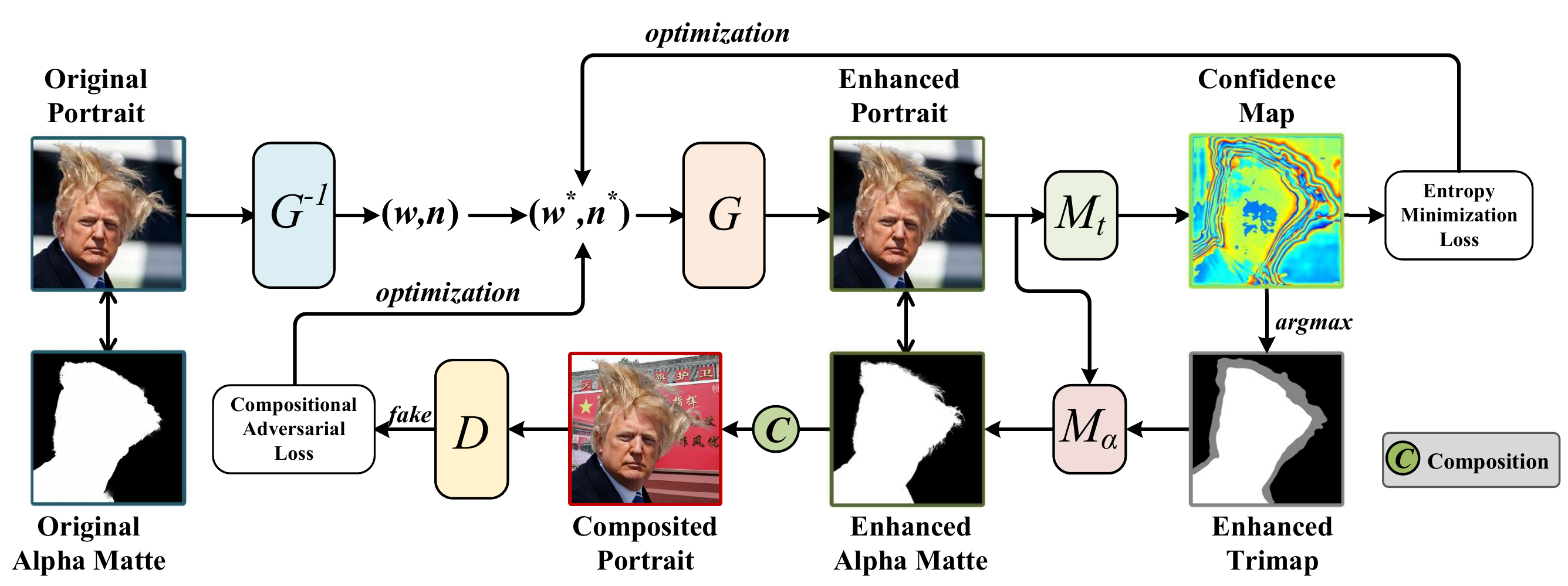}
\caption{Overview of the optimization on the latent codes~$(w,n)$. Here~$(w,n)$~could be acquired by GAN inversion of a given portrait or randomly generation. The optimization process is governed by combined losses. Entropy minimization loss and compositional adversarial loss are proposed to ease the matting difficulty and alpha matte quality respectively. An enhanced portrait is obtained by feeding the optimal latent codes~$(w^*,n^*)$~to the generator $G$ of StyleGAN. $M_{t}$ and $M_{\alpha}$ denotes the pre-trained trimap model and alpha matte estimation model respectively. $D$ is a pre-trained discriminator. We omit the portrait consistency loss and perceptual loss for simplicity.}
\label{fig:overview}\vspace{2mm}
\end{figure*}

We take two latent vectors $(w_o,n_o)$ in $\mathcal{W+}$ and $\mathcal{N}$ spaces to represent the input portrait~$I_o$, and these vectors can be the inverted codes using GAN inversion methods~\cite{abdal2019image2stylegan,abdal2019image2stylegan2} or randomly generated ones. Our objective is to obtain the optimal latent vectors $(w^*,n^*)$ as follows
\begin{equation}
\label{eq2}
(w^*, n^*)= \arg \min _{(w,n)} \lambda_{1} \mathcal{L}_{em}+ \lambda_{2} \mathcal{L}_{ca}+ \lambda_{3} \mathcal{L}_{pc}+ \lambda_{4} \mathcal{L}_{pp},
\end{equation}
where $\mathcal{L}_{em}$, $\mathcal{L}_{ca}$, $\mathcal{L}_{pc}$, $\mathcal{L}_{pp}$ denote the entropy minimization loss, compositional adversarial loss, portrait consistency loss, and perceptual loss respectively. $\lambda_{x}$ is the balancing weight. These losses constrain the searching of the enhanced portrait $G(w^*,n^*)$ to be matting-specific and face-preserved. Fig.~\ref{fig:overview} shows the pipeline of our optimization.

\subsubsection{Entropy Minimization Loss}

The first goal of our portrait enhancement is to alleviate the difficulty of separating the foreground and background. As there is no ground-truth for training the regression problem of matting, we instead propose to minimize the prediction entropy. The rationale behind this loss is to edit the portrait so that the matting model predicts a high-confidence alpha matte, \ie, refining ambiguous regions of the alpha matte. However, alpha matte estimation is a regression problem without explicit confidences during inference. To resolve this problem, we shift our optimization target to a classification problem, which indeed equals to the essential trimap prediction (\ie, classifying foreground, background, and unknown regions) in the pipeline. The new optimization task shares the same goal of obtaining the enhanced portrait, as the quality of a trimap is proportional to the quality of the predicted alpha matte. Following the entropy minimization principle~\cite{long2016unsupervised,long2018conditional,grandvalet2005semi,tbd_wang2019tada}, we encourage the low-density separation between classes by minimizing the entropy of confidence map, which can be represented as
\begin{equation}
\begin{aligned}
\label{eq6}
\mathcal{L}_{em}= H(M_{t}(G(w,n))),
\end{aligned}
\end{equation}
where $H(x)=-\sum x \log (x)$ is the entropy function, and~$M_{t}(G(w,n))$~represents the confidence map of~$G(w,n)$~predicted by the trimap model $M_{t}$. In this way, we refine the portrait $G(w,n)$ so that a more certain confidence map can be obtained.

\subsubsection{Compositional Adversarial Loss}
As the entropy minimization loss is not directly applied to the alpha matte, it is intractable to evaluate whether the enhanced alpha matte works better than the original one, especially without ground-truth data.

We further introduce a compositional adversarial loss to evaluate the alpha matte. Instead of directly examining the alpha matte, we use the enhanced alpha matte, with a new background image randomly selected from MS COCO dataset~\cite{lin2014microsoft} to compose a new portrait image for evaluation. The proposed compositional adversarial loss aligns the distance between the distributions of real portraits and the composite ones, yielding a better alpha matte. Specifically, we utilize a pre-trained discriminator~$D$ of StyleGAN to distinguish the composite portrait as a fake sample, which is defined as follows
\begin{equation}
\mathcal{L}_{ca}= -\log (D(C(F,B_r,\alpha_{e}))),
\label{cal}
\end{equation}
where~$C(\cdot)$~is a composition function that composes a new portrait with the enhanced alpha matte~$\alpha_{e}$, foreground portrait~$F$, and a random background image~$B_r$. 

\rev{We use Closed-form Matting~\cite{levin2007closed} for estimating the foreground portrait~$F$ based on the enhanced matte $\alpha_{e}$. Particularly, given an alpha matte $\alpha$, Closed-form Matting~\cite{levin2007closed} estimates the foreground $F$ and background $B$ by minimizing a cost function for each pixel $i$ and color channel $c$:}

\begin{equation}
\begin{aligned}
\underset{\text { global }}{\operatorname{cost}}(F,B)=\sum_{i \in I} \sum_{c} & 
 {\left[\alpha_{i} F_{i}^{c}+\left(1-\alpha_{i}\right) B_{i}^{c}-I_{i}^{c}\right]^{2} } \\
+\left|\alpha_{i_{x}}\right| &
 {\left[\left(F_{i_{x}}^{c}\right)^{2}+\left(B_{i_{x}}^{c}\right)^{2}\right] } \\
+\left|\alpha_{i_{y}}\right| & 
{\left[\left(F_{i_{y}}^{c}\right)^{2}+\left(B_{i_{y}}^{c}\right)^{2}\right] }.
\end{aligned}
\label{fg_est}
\end{equation}

\rev{The first term aligns the distance between the composed and original images, and the rest two terms reduce the magnitude of color gradients $F_{i_{x}}$, $F_{i_{y}}$, $B_{i_{x}}$ and $B_{i_{y}}$ in regions of large $\alpha$-gradients $\left|\alpha_{i_{x}}\right|$ and $\left|\alpha_{i_{y}}\right|$ for preserving the texture information.}

\subsubsection{Portrait Consistency Loss}
In the scenario of using a real portrait, maintaining the original appearance is of great practical importance. Therefore, we propose to minimize the pixel-wise $L_2$ distance between the enhanced portrait image~$G(w^*,n^*)$~and the original one~$I_o$ to constrain the editing degree. Portrait consistency loss is defined as follows
\begin{equation}
\label{eq3}
\mathcal{L}_{pc} = \|G(w^*,n^*)-I_o\|_{2}.
\end{equation}

{\subsubsection{Perceptual Loss}
Besides the portrait consistency loss, we also utilize a perceptual loss~\cite{johnson2016perceptual} for maintaining the subtle facial details between two portraits. We computes the $L_2$ distance between features produced by the enhanced portrait image $G(w^*,n^*)$ and the original one $I_o$ after feeding to a pre-trained VGG-16 network. Perceptual loss is defined as
\begin{equation}
\label{eq3}
\mathcal{L}_{pp} = \| \Phi_{VGG}(G(w^*,n^*))-\Phi_{VGG}(I_o) \|_{2},
\end{equation}
where $\Phi_{VGG}(\cdot)$ is a pre-trained VGG-16 network and we select the features produced by \texttt{conv4\_2} layer for calculating the loss.

\setlength{\textfloatsep}{5pt}
\renewcommand{\algorithmicrequire}{\textbf{Input:}} 
\renewcommand{\algorithmicensure}{\textbf{Output:}} 
\begin{algorithm}[t]
    \caption{Optimizing the latent codes for the enhanced portrait and alpha matte}
    \label{alg:alg1}
    \begin{algorithmic}[1]
    \REQUIRE Original portrait image~$I_{o}$, random background image~$I_{B}$, {pre-trained trimap model~$M_{t}$, pre-trained matting model~$M_{\alpha}$}, pre-trained StyleGAN~$G$~and~$D$, iterative times~$T$.
    \ENSURE  Enhanced portrait~$G(w^*,n^*)$~and~ alpha matte~${\alpha}_{e}$.
    \STATE \textbf{begin}
            \STATE Invert the portrait image~$I_{o}$~to the latent vectors~$(w_o,n_o)$ under the supervision of $\mathcal{L}_{pc}$ and $\mathcal{L}_{pp}$, and $(w,n) = (w_o, n_o)$.
            \FOR {\emph{steps} = 1, 2, ..., T }
                \STATE Generate an enhanced portrait image~$G(w,n)$~by $(w,n)$ to~$G$.
                {
                \STATE Predict the trimap of~$G(w,n)$~by~$M_{t}$.
                \STATE Estimate the enhanced alpha matte~${\alpha}_{e}$~by~$M_{\alpha}$.
                }
                \STATE \rev{Estimate the foreground layer $F$ using the enhanced alpha matte~${\alpha}_{e}$ by minimizing cost function $\underset{\text { global }}{\operatorname{cost}}(F,B)$.}
                \STATE \rev{Compose a new portrait with a random background image~$B_r$, foreground image~$F$, and~${\alpha}_{e}$~based on Eq.~\eqref{eq1}}.
                \STATE Optimize the latent codes~$(w,n)$~with Eq.~\eqref{eq2}.
            \ENDFOR
        \STATE $(w*,n*) = (w, n)$.
        \RETURN $G(w^*,n^*)$~and~${\alpha}_e$
        \STATE \textbf{end begin}
    \end{algorithmic}
\end{algorithm}

\subsection{Portrait Enhancement}
\label{enhancement}
Although the proposed optimization is performed on the latent space of a generative model, it can be applied for real image enhancement. Given a portrait image~$I_{o}$, we apply GAN inversion method~\cite{abdal2019image2stylegan2} to map the portrait image~$I_{o}$~to the latent vectors~$(w_o,n_o)$~in the intermediate space $\mathcal{W+}$ and the noise space~$\mathcal{N}$~of a pre-trained StyleGAN. Once the optimal latent codes $(w^*,n^*)$ are properly trained, it is fed to the generator $G$ again to obtain the enhanced portrait $G(w^*,n^*)$. Algorithm~\ref{alg:alg1} summarizes the pseudo-code for optimizing the latent vectors of a real input portrait.

\subsection{Portrait Generation}
An important property of using GANs is the flexibility of generating data. We first leverage StyleGAN to generate a large number of portrait images with random latent vectors. Then these images are enhanced by discovering the optimal latent codes using Eq.~\eqref{eq2}. Note that the portrait consistency loss~$\mathcal{L}_{pc}$ and perceptual loss~$\mathcal{L}_{pp}$ are removed as no original image is needed, and it can also relax the optimization with a large searching space. In this way, we can generate arbitrary numbers of matting-specific portraits. All the predicted alpha mattes are of high-confidence that can be treated as pseudo GT data. \revf{Note here we use the initial generated portraits, together with the enhanced alpha mattes to form the pseudo GT pairs. Here we use the initial portraits instead of the enhanced portraits, as the enhanced portraits are likely to be overfitted to the pre-trained model, and provide less supervision information.}


\section{Experiments}

\begin{table*}[!h]
  \renewcommand\arraystretch{1.2}
     \caption{Quantitative evaluation on portrait enhancement with respect to \rev{six} matting models on two datasets. The lower the better for all metrics. Top 2 improved methods (in percentage) for each metric are marked in \textcolor[rgb]{0.00,0.07,1.00}{blue} and \textcolor[rgb]{0.00,0.59,0.00}{green} respectively.}
     \begin{center}
     \setlength{\tabcolsep}{0.125cm}{
     \begin{tabular}{c|c|c|c|c|c|c|c|c|c|c|c|c}
         \hline
         \multirow{2}{*}{\diagbox{Models}{Metrics}}     &\multicolumn{3}{c|}{Grad$\downarrow$}  &\multicolumn{3}{c|}{Conn$\downarrow$}  &\multicolumn{3}{c|}{MSE$\downarrow$}  &\multicolumn{3}{c}{SAD$\downarrow$} \\
         \cline{2-13}
                                              &Ori.         &Enh.    & $\Delta$ / $|$\%$\Delta$$|$
                                              &Ori.         &Enh.    & $\Delta$ / $|$\%$\Delta$$|$
                                              &Ori.         &Enh.    & $\Delta$ / $|$\%$\Delta$$|$
                                              &Ori.         &Enh.    & $\Delta$ / $|$\%$\Delta$$|$ \\
         \hline
         \multicolumn{13}{c}{Portrait Matting Dataset~\cite{shen2016deep}} \\
         \hline
         DIM~\cite{xu2017deep}                                        &8.699    &7.459    &\textcolor[rgb]{0.00,0.59,0.00}{-1.240/14.25\%}
                                                                      &12.291   &11.374   &{-0.917/7.46\%}
                                                                      &0.017    &0.015    &\textcolor[rgb]{0.00,0.59,0.00}{-0.002/11.76\%}
                                                                      &13.269   &12.464   &-0.805/6.07\% \\

         \rowcolor{gray!20} IndexNet~\cite{lu2019indices}             &7.405    &6.168    &\textcolor[rgb]{0.00,0.07,1.00}{-1.237/16.70\%}
                                                                      &11.345   &10.302   &\textcolor[rgb]{0.00,0.59,0.00}{-1.043/9.19\%}
                                                                      &0.016    &0.014    &\textcolor[rgb]{0.00,0.07,1.00}{-0.002/12.5\%}
                                                                      &11.826   &10.799   &\textcolor[rgb]{0.00,0.07,1.00}{-1.027/8.68\%} \\

         GCA~\cite{li2020natural}                                     &9.170    &8.115    &-1.055/11.50\%
                                                                      &11.382   &10.877   &-0.505/4.43\%
                                                                      &0.017    &0.016    &-0.001/5.88\%
                                                                      &11.751   &11.384   &-0.367/3.12\%\\

         \rowcolor{gray!20} AdaMatting~\cite{cai2019disentangled}     &21.882   &21.186   &-0.696/3.19\%
                                                                      &18.291   &17.458   &-0.833/4.55\%
                                                                      &0.029    &0.028    &-0.001/3.44\%
                                                                      &18.225   &17.664   &-0.561/3.08\%\\

         SampleNet~\cite{tang2019learning}                            &8.511    &7.703    &-0.808/9.49\%
                                                                      &11.119   &10.132   &\textcolor[rgb]{0.00,0.07,1.00}{-0.987/8.87\%}
                                                                      &0.017    &0.016    &-0.001/5.88\%
                                                                      &11.503   &10.668   &\textcolor[rgb]{0.00,0.59,0.00}{-0.835/7.26\%}\\

        \rowcolor{gray!20} \rev{MODNet~\cite{ke2020modnet}}           &\rev{22.436}   &\rev{21.552}   &\rev{-0.884/3.94\%}
                                                                      &\rev{17.523}   &\rev{17.227}   &\rev{-0.296/1.69\%}
                                                                      &\rev{0.028}    &\rev{0.027}    &\rev{-0.001/3.57\%}
                                                                      &\rev{19.359}   &\rev{18.907}   &\rev{-0.452/2.33\%}\\
                                                           
         \hline
         \multicolumn{13}{c}{Adobe Image Matting Dataset~\cite{xu2017deep}} \\
         \hline
         DIM~\cite{xu2017deep}                                        &21.718   &15.589   &\textcolor[rgb]{0.00,0.59,0.00}{-6.129/28.22\%}
                                                                      &23.364   &21.366   &\textcolor[rgb]{0.00,0.07,1.00}{-1.998/8.55\%}
                                                                      &0.038    &0.037    &\textcolor[rgb]{0.00,0.59,0.00}{-0.001/2.63\%}
                                                                      &23.417   &22.766   &\textcolor[rgb]{0.00,0.00,0.00}{-0.651/2.78\%} \\

         \rowcolor{gray!20} IndexNet~\cite{lu2019indices}             &21.268   &18.617   &\textcolor[rgb]{0.00,0.07,0.00}{-2.651/12.46\%}
                                                                      &22.541   &21.144   &\textcolor[rgb]{0.00,0.00,0.00}{-1.397/6.20\%}
                                                                      &0.038    &0.037    &{-0.001/2.63\%}
                                                                      &22.701   &22.323   &\textcolor[rgb]{0.00,0.07,0.00}{-0.378/1.67\%}\\

         GCA~\cite{li2020natural}                                     &22.489   &17.261   &-5.228/23.25\%
                                                                      &22.106   &21.165   &-0.941/4.26\%
                                                                      &0.038    &0.037    &{-0.001/2.63\%}
                                                                      &22.205   &21.445   &\textcolor[rgb]{0.00,0.59,0.00}{-0.760/3.42\%} \\

         \rowcolor{gray!20} AdaMatting~\cite{cai2019disentangled}     &24.934   &19.577   &\textcolor[rgb]{0.00,0.07,1.00}{-5.357/23.35\%}
                                                                      &23.274   &21.953   &\textcolor[rgb]{0.00,0.59,0.00}{-1.981/8.51\%}
                                                                      &0.041    &0.039    &\textcolor[rgb]{0.00,0.07,1.00}{-0.002/4.88\%}
                                                                      &25.613   &24.655   &\textcolor[rgb]{0.00,0.07,1.00}{-0.958/0.374\%}\\

         SampleNet~\cite{tang2019learning}                            &21.806   &18.729   &-3.077/14.11\%
                                                                      &23.029   &21.527   &\textcolor[rgb]{0.00,0.00,0.00}{-1.502/6.52\%}
                                                                      &0.038    &0.037    &{-0.001/2.63\%}
                                                                      &22.345   &22.003   &\textcolor[rgb]{0.00,0.00,0.00}{-0.342/1.53\%}\\

        \rowcolor{gray!20} \rev{MODNet~\cite{ke2020modnet}}           &\rev{25.783}   &\rev{25.042}   &\rev{-0.741/2.96\%}
                                                                      &\rev{24.433}   &\rev{23.903}   &\rev{-0.530/2.17\%}
                                                                      &\rev{0.045}    &\rev{0.042}    &\rev{-0.003/6.67\%}
                                                                      &\rev{27.335}   &\rev{27.029}   &\rev{-0.306/1.11\%}\\

        \hline
     \end{tabular}
     }
     \end{center}
     \label{table:1}
\end{table*}

\subsection{Implementation Details}

We implement the proposed method in Pytorch on a PC with an Nvidia GeForce RTX 2080Ti. The proposed method takes about 3 mins to optimize the latent codes for each image.
We utilize the generator and discriminator of StyleGAN2~\cite{karras2019analyzing} pre-trained on the FFHQ dataset~\cite{karras2019style}. Since the output resolution of StyleGAN2 is $1024\times 1024$, we resize the original portraits to the resolution of $1024\times 1024$ for portrait enhancement. We optimize the latent codes by Adam optimizer~\cite{kingma2014adam}. We use 500 gradient descent steps with a learning rate of 0.0001 and a weight decay of 0.0005. We empirically set the balancing weights in Eq.~\eqref{eq2} as $\lambda_{1}=1$, $\lambda_{2}=1$, $\lambda_{3}=10$, and $\lambda_{4}=10$.

\subsection{Dataset and Evaluation Metrics}

We conduct our experiments on two datasets, the first one is the largest portrait matting dataset~\cite{shen2016deep}. It was collected from Flickr with a good variety of age, color, clothing, and hairstyle. Alpha mattes are annotated with intensive user interactions. It consists of 2,000 samples with a resolution of $600\times 800$, and we follow \cite{shen2016deep} to divide images into training and testing sets with 1,700 and 300 images, respectively.

Although the above dataset is the largest portrait matting dataset, it is not fully labeled by human. To further evaluate our method on high-quality data, we select portrait images from the test set of Adobe Image Matting Dataset~\cite{xu2017deep}, resulting a new dataset that consists of 140 images.


We use four quantitative metrics to evaluate the alpha matting results, namely, the mean square error (MSE), the sum of absolute differences (SAD), the gradient error (Grad), and connectivity errors (Conn)~\cite{rhemann2009perceptually}. MSE and SAD mainly focus on the statistical differences between two alpha mattes, but they may not correlate to the visual quality perceived by human~\cite{rhemann2009perceptually}. Compared with MSE and SAD, Conn and Grad are more related to human perception. Specifically, Conn evaluates the disconnected foreground objects and Grad mainly concentrates on the over-smoothed or erroneous discontinuities in the alpha matte.

\subsection{Trimap and Matting Models} 
DeeplabV3~\cite{chen2017deeplab}~with a backbone of ResNet50~\cite{he2016deep} pre-trained on the CoCo-Stuff Dataset~\cite{caesar2018coco} is employed as our {trimap model}. Following the work~\cite{cai2019disentangled}, we also use alpha mattes as the ground-truth for fine-tuning the trimap model. We take \rev{six} state-of-the-art matting methods, Deep Image Matting (DIM)~\cite{xu2017deep}, IndexNet~\cite{lu2019indices}, Guided Contextual Attention (GCA)~\cite{li2020natural}, AdaMatting~\cite{cai2019disentangled}, SampleNet~\cite{tang2019learning} and \rev{MODNet~\cite{ke2020modnet}} as the reference matting models. \rev{Note that all methods need trimap as the extra input except MODNet~\cite{ke2020modnet}. MODNet is a trimap-free matting method, and only the loss of $\mathcal{L}_{ca}$ can be applied to this model.} All of them are pre-trained on the Adobe Image Matting Dataset~\cite{xu2017deep}~and then fine-tuned on the portrait matting dataset~\cite{shen2016deep} except SampleNet~\cite{tang2019learning}. Because it requires two inputs of foreground and background images, therefore we cannot fine-tune their model using the portrait matting dataset~\cite{shen2016deep} or our generated portraits. Instead we directly use the model pre-trained on Adobe Image Matting Dataset~\cite{xu2017deep}~for enhancement, which also demonstrates the generalization ability of our method.

\begin{figure*}[!h]
  \centering
  \subfloat[\scriptsize{Portrait}]{\label{fig:analysis_a}
    \rotatebox[origin=c]{270}{Original \hspace{7mm} Enhanced \hspace{7mm} Original \hspace{7mm} Enhanced \hspace{7mm} Original \hspace{7mm} Enhanced \hspace{7mm} Original \hspace{7mm} Enhanced}
    \begin{minipage}{0.09\linewidth}
    \includegraphics[width=\linewidth]{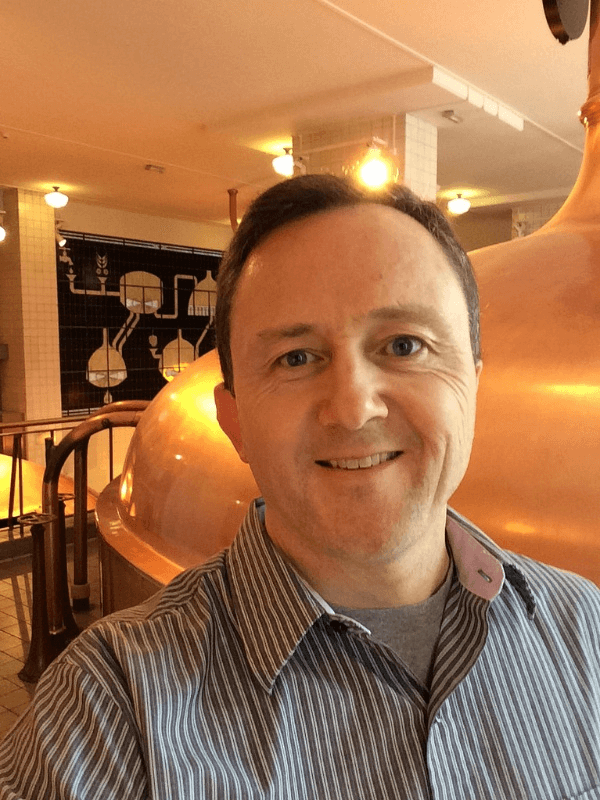}
    \includegraphics[width=\linewidth]{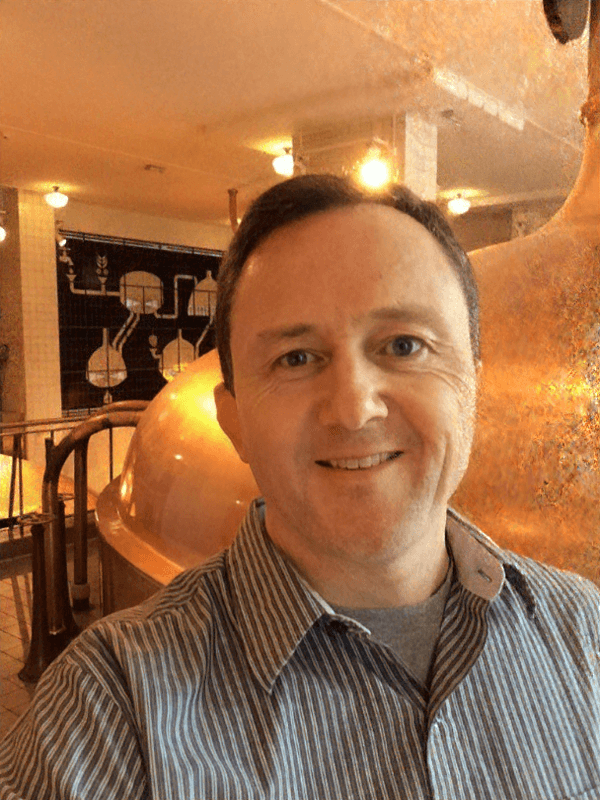}
    \includegraphics[width=\linewidth]{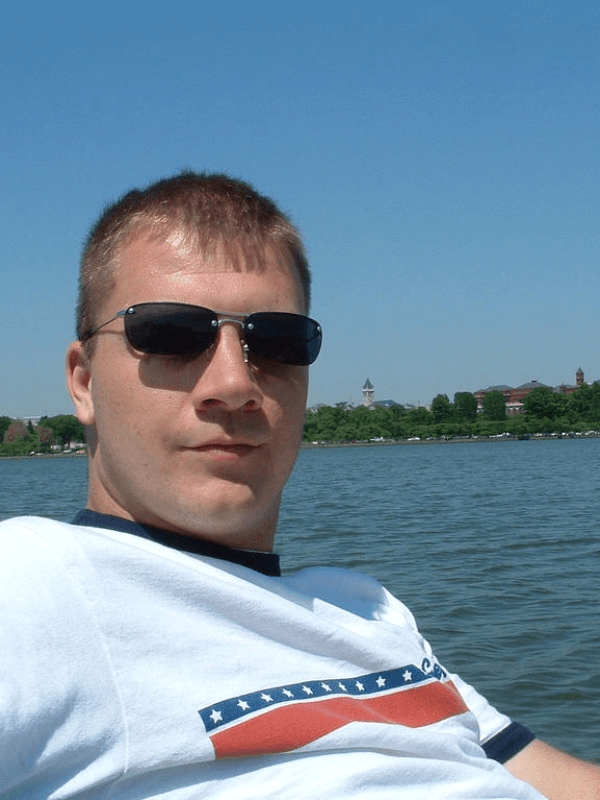}
    \includegraphics[width=\linewidth]{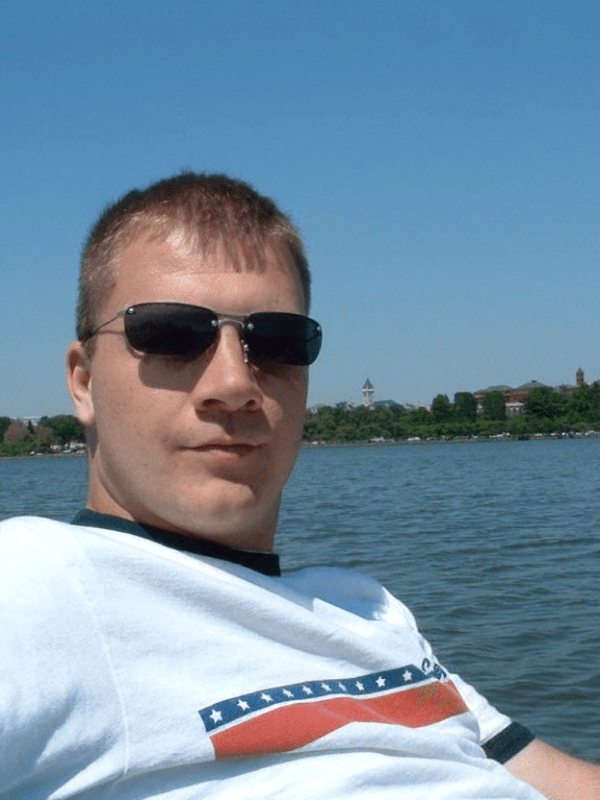}
    \includegraphics[width=\linewidth]{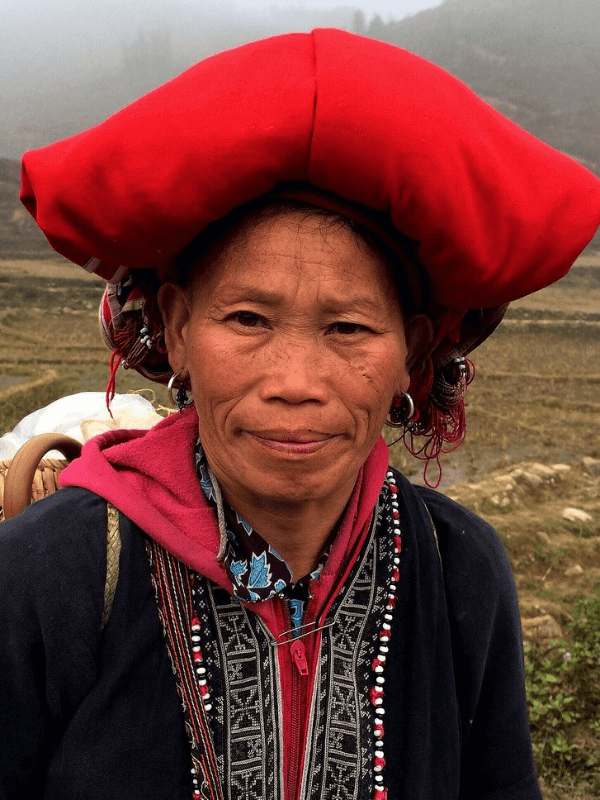}
    \includegraphics[width=\linewidth]{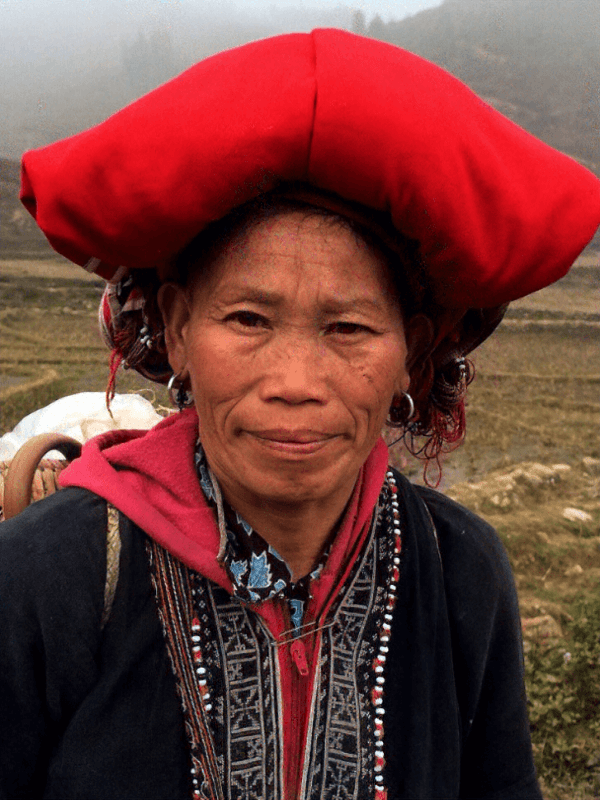}
    \includegraphics[width=\linewidth]{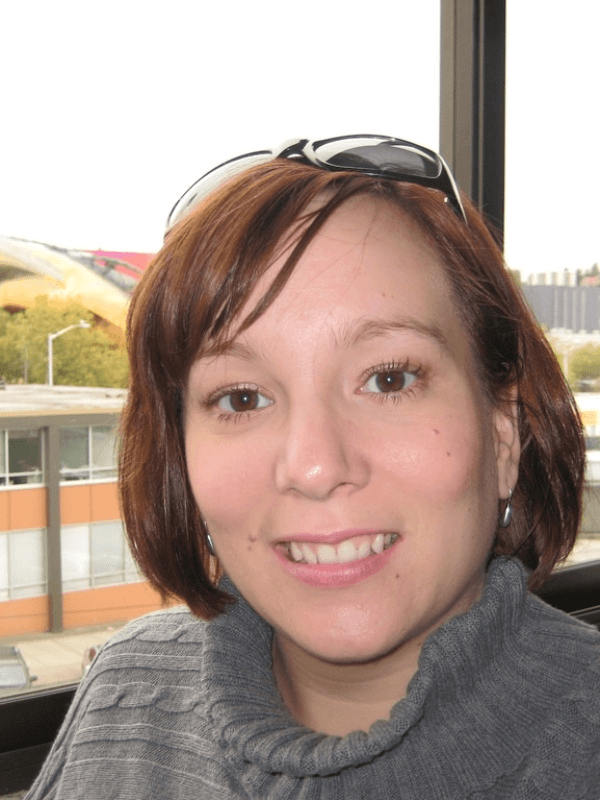}
    \includegraphics[width=\linewidth]{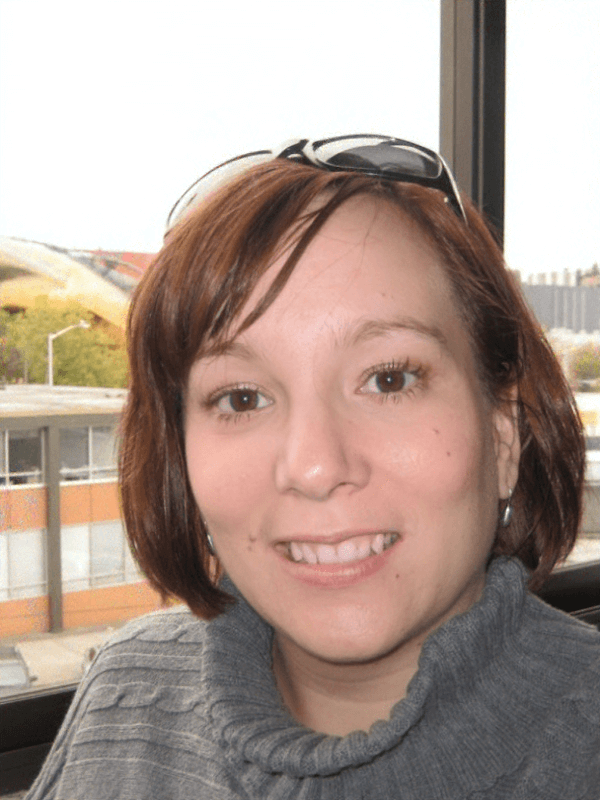}
    \end{minipage}
    }
  \hspace{-3.5mm}
  \subfloat[\scriptsize{Trimap}]{\label{fig:analysis_b}
    \begin{minipage}{0.09\linewidth}
    \includegraphics[width=\linewidth]{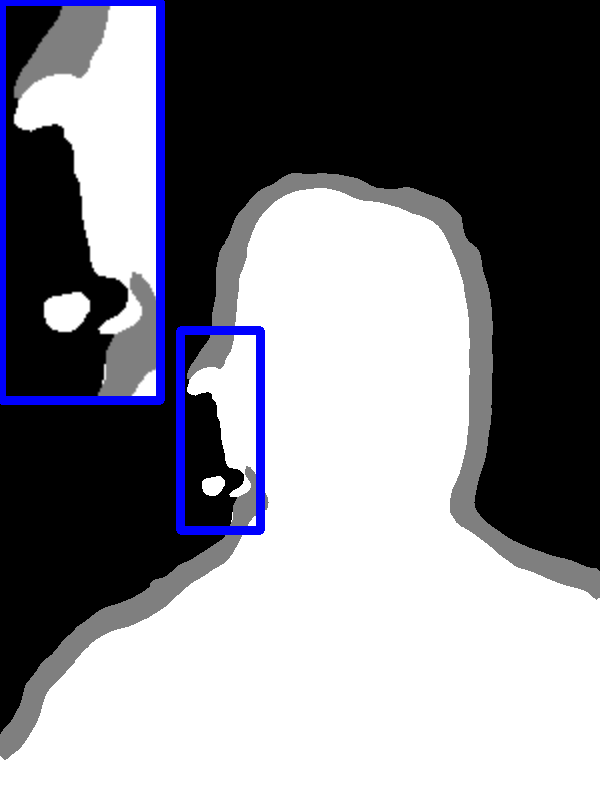}
    \includegraphics[width=\linewidth]{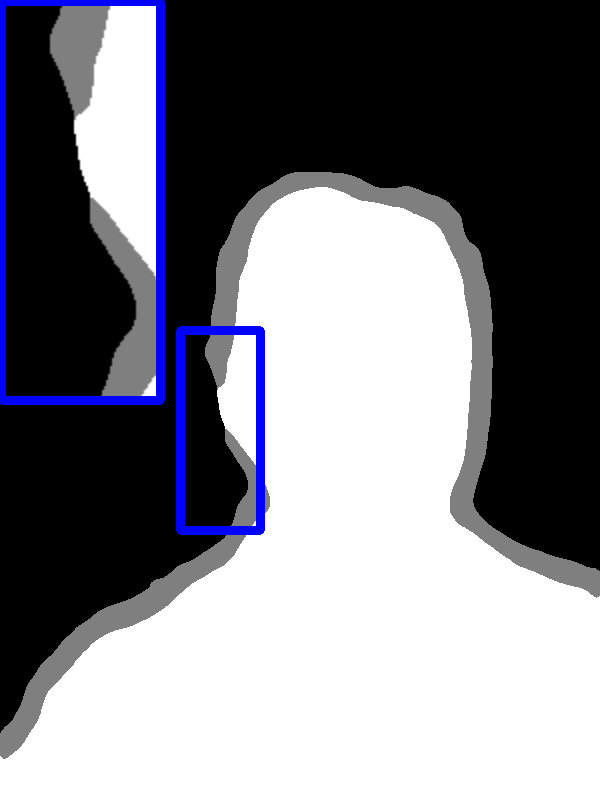}
    \includegraphics[width=\linewidth]{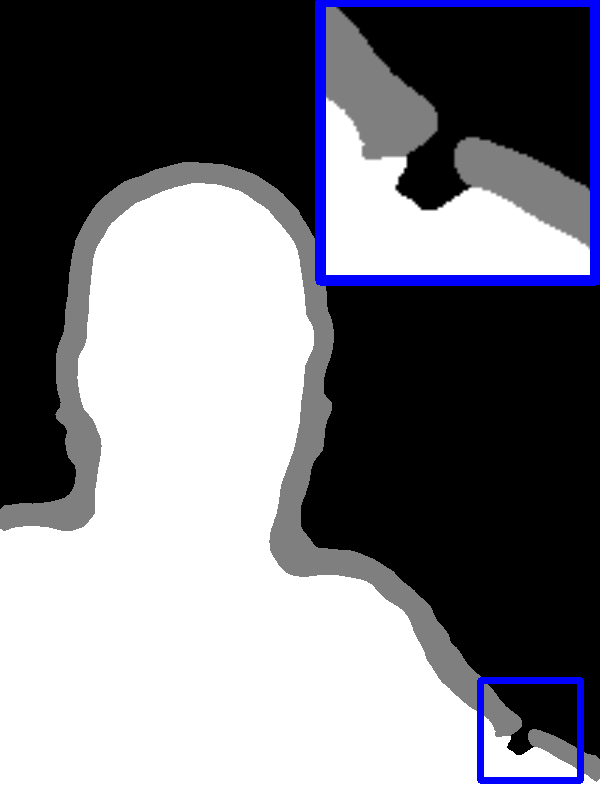}
    \includegraphics[width=\linewidth]{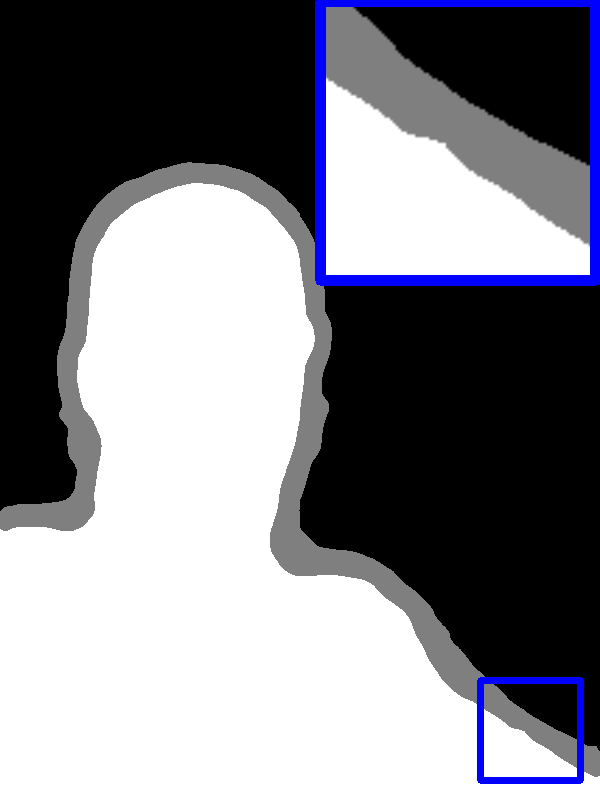}
    \includegraphics[width=\linewidth]{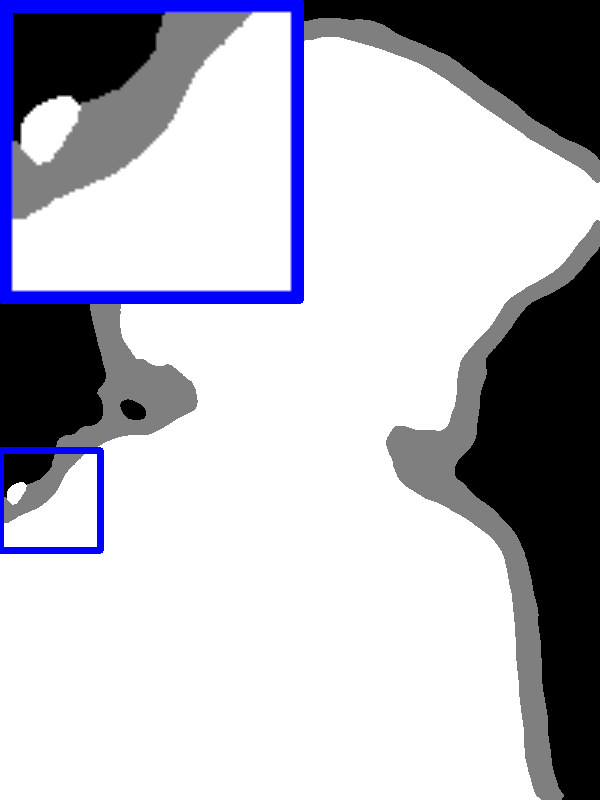}
    \includegraphics[width=\linewidth]{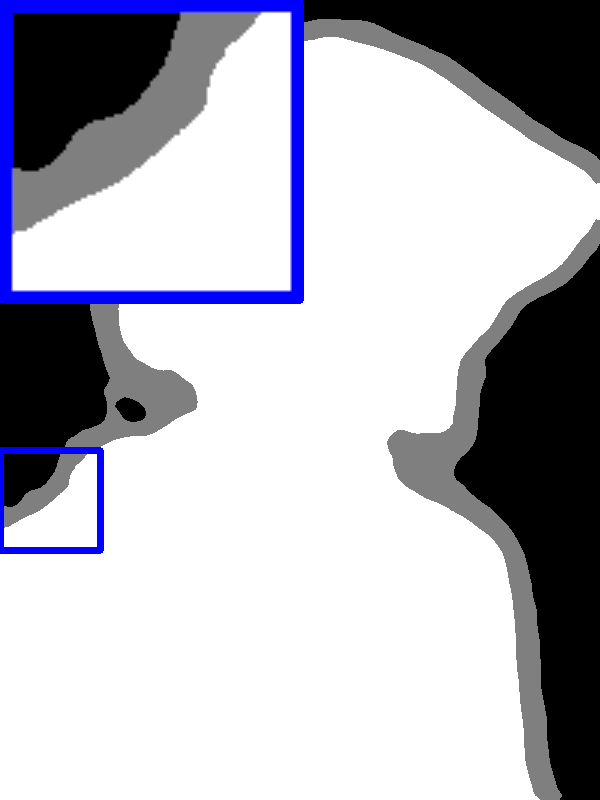}
    \includegraphics[width=\linewidth]{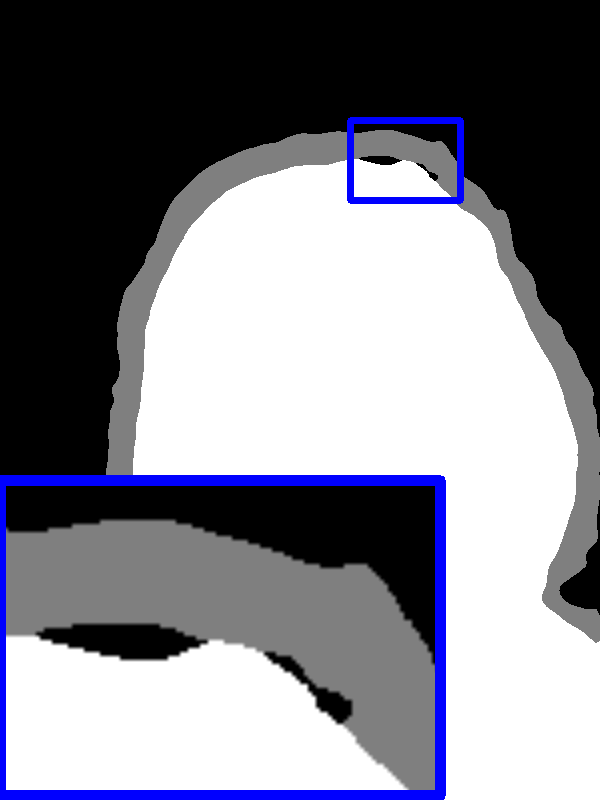}
    \includegraphics[width=\linewidth]{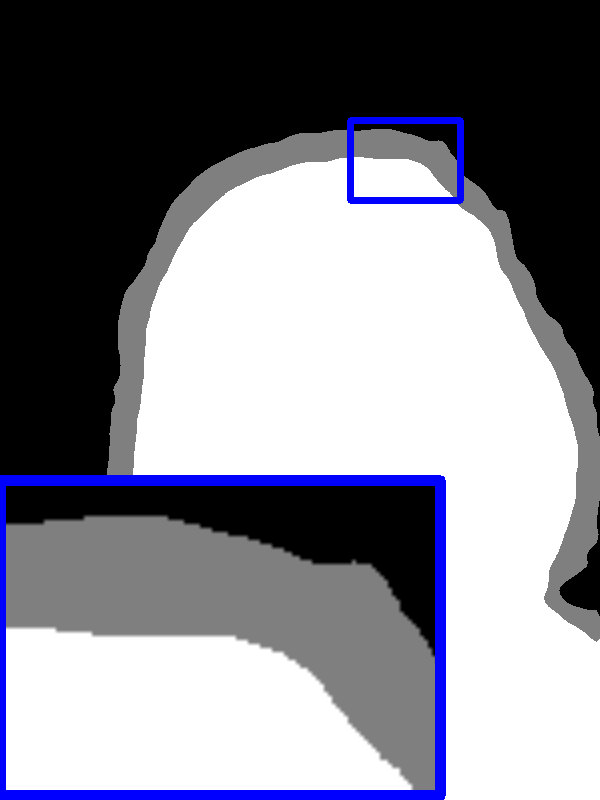}
    \end{minipage}
  }
  \hspace{-3.5mm}
  \subfloat[\scriptsize{Matte}]{\label{fig:analysis_c}
    \begin{minipage}{0.09\linewidth}
    \includegraphics[width=\linewidth]{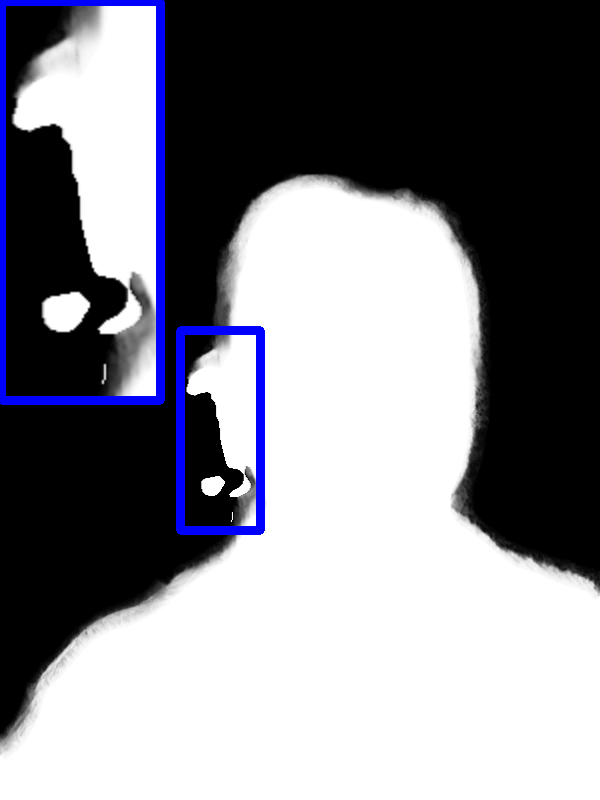}
    \includegraphics[width=\linewidth]{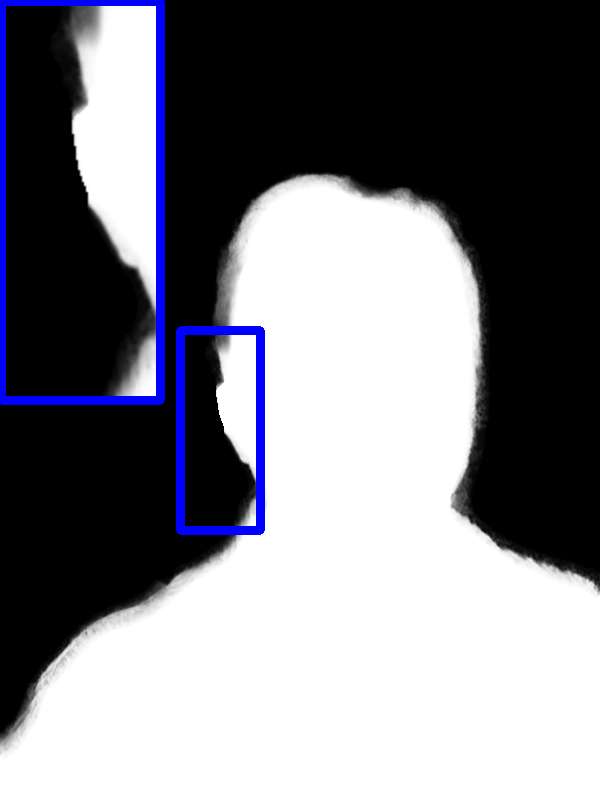}
    \includegraphics[width=\linewidth]{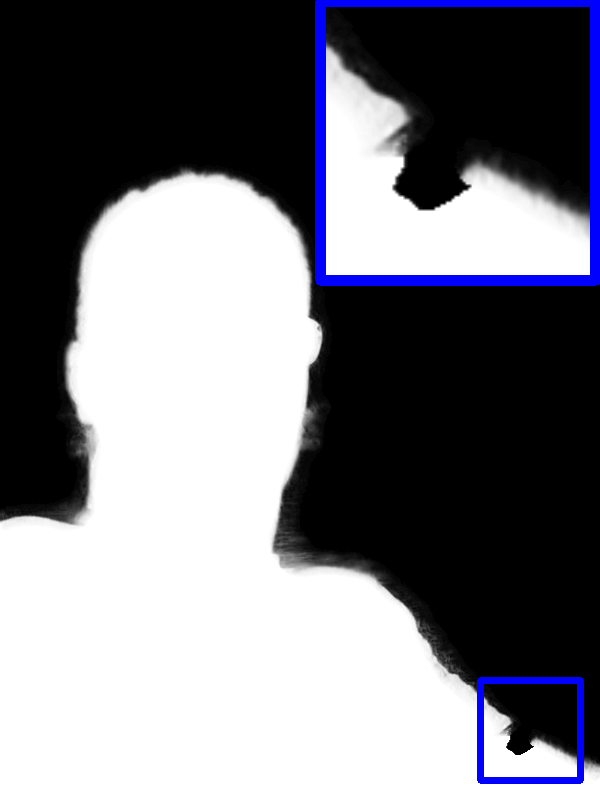}
    \includegraphics[width=\linewidth]{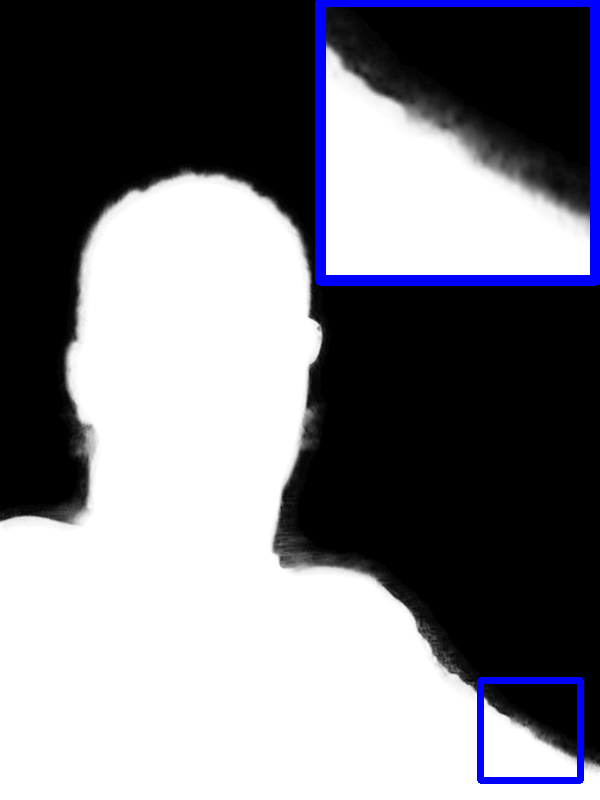}
    \includegraphics[width=\linewidth]{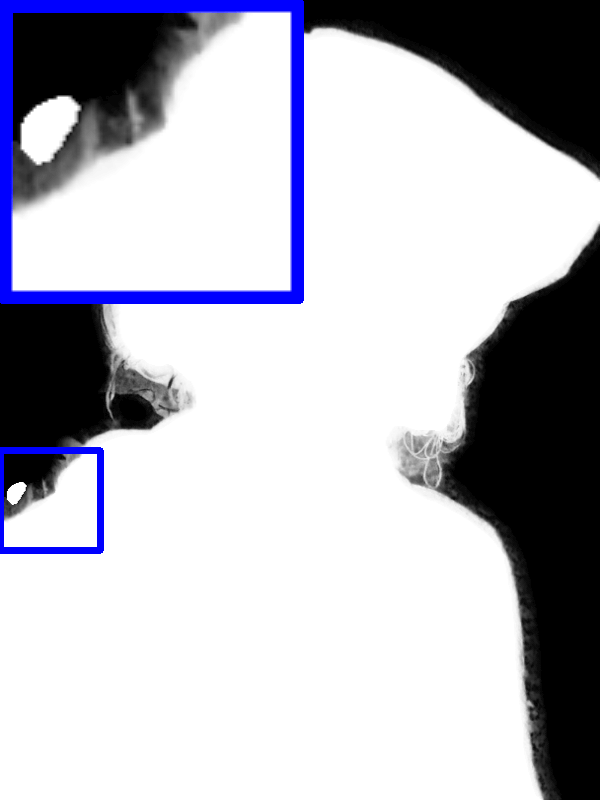}
    \includegraphics[width=\linewidth]{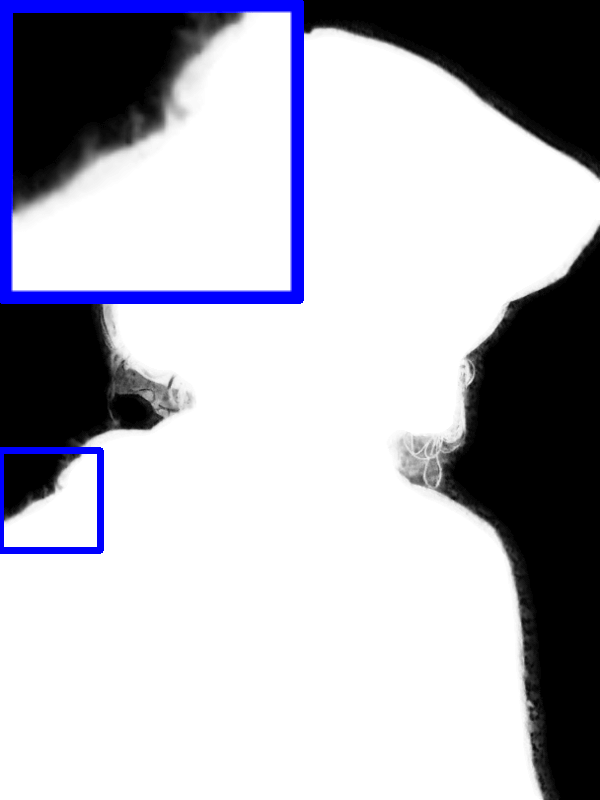}
    \includegraphics[width=\linewidth]{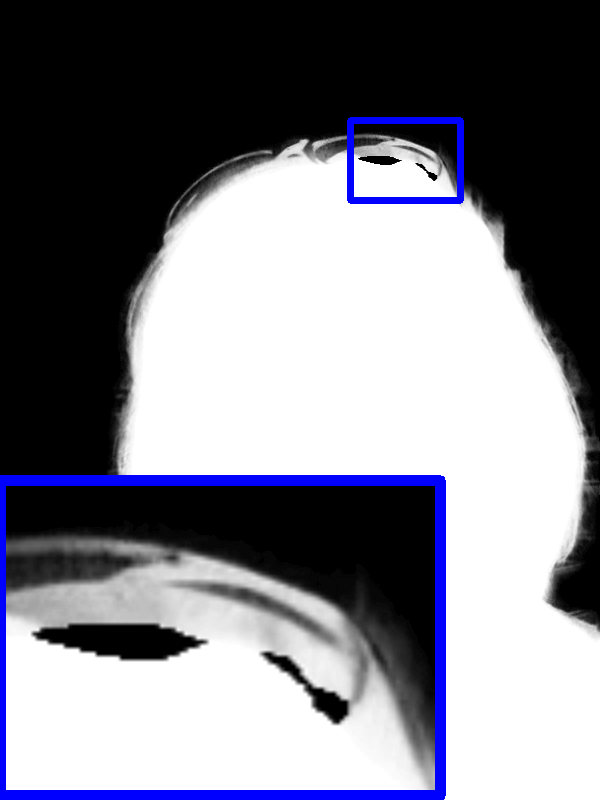}
    \includegraphics[width=\linewidth]{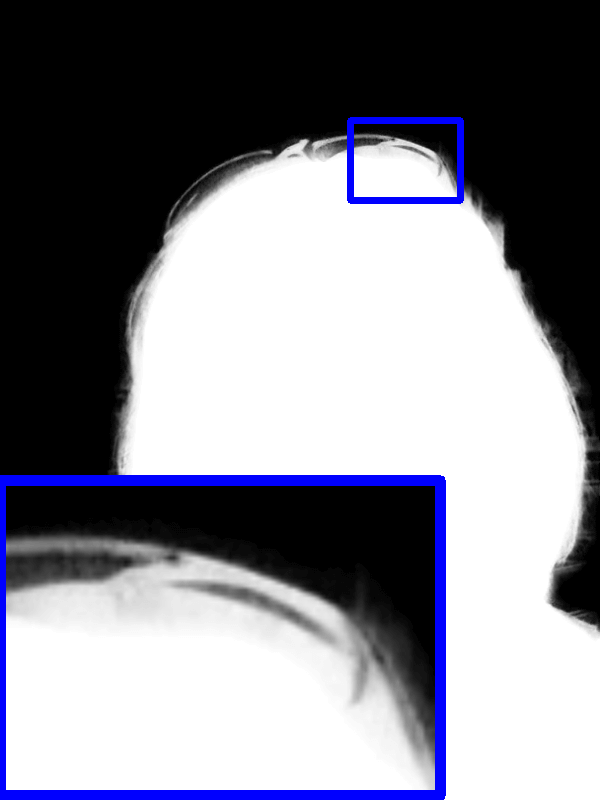}
    \end{minipage}
    }
  \hspace{-3.5mm}
  \subfloat[\scriptsize{Pert.\& GT}]{\label{fig:analysis_d}
    \begin{minipage}{0.09\linewidth}
    \includegraphics[width=\linewidth]{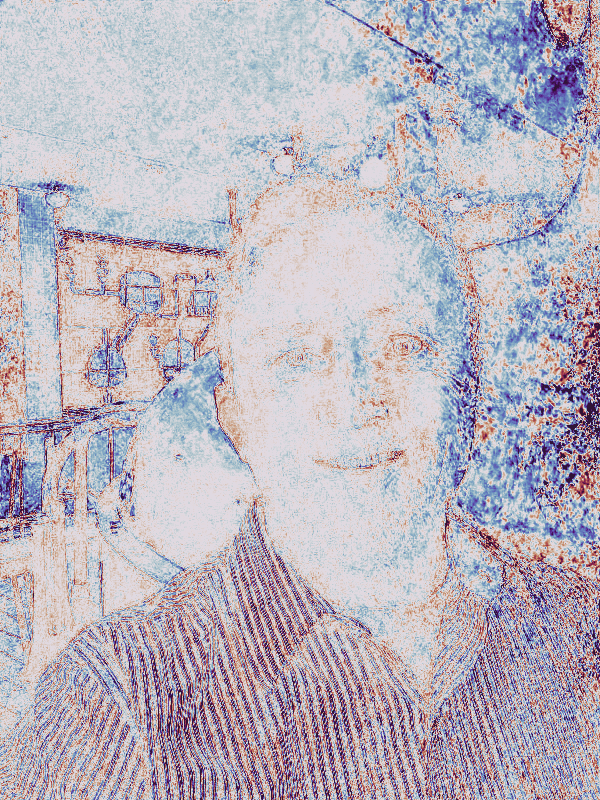}
    \includegraphics[width=\linewidth]{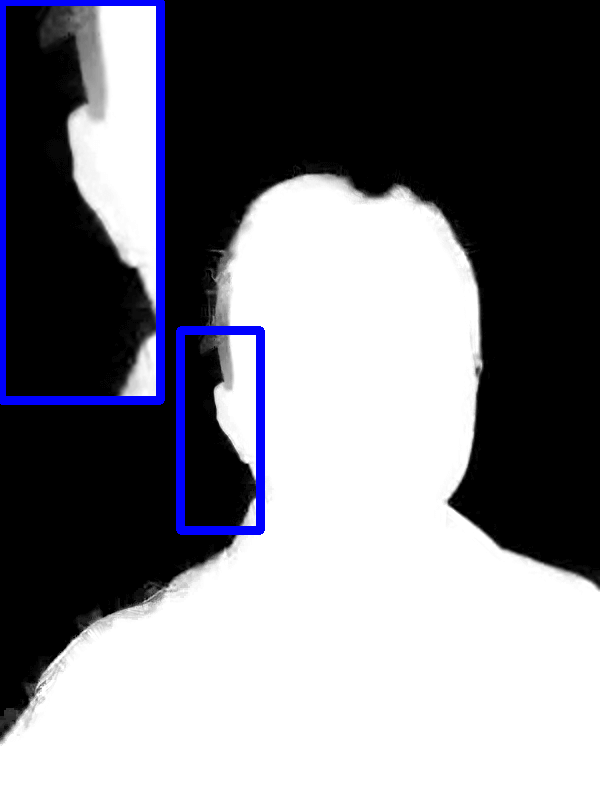}
    \includegraphics[width=\linewidth]{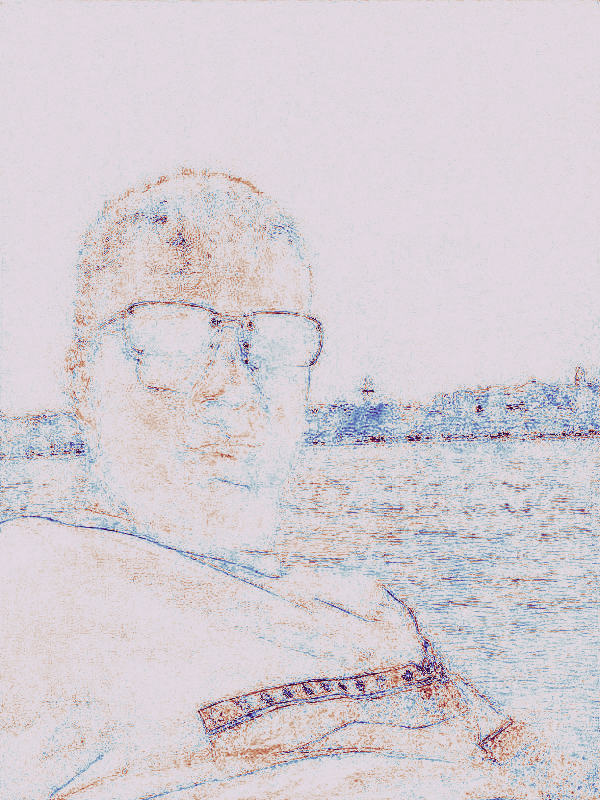}
    \includegraphics[width=\linewidth]{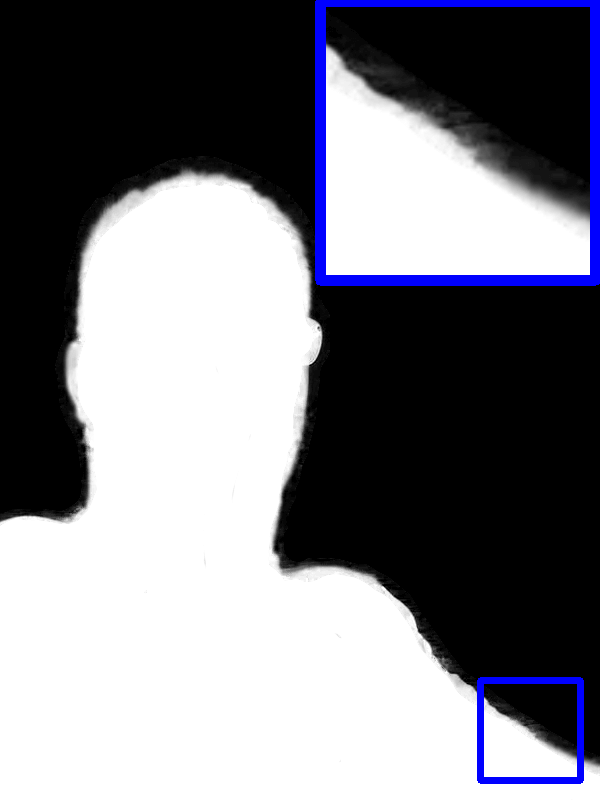}
    \includegraphics[width=\linewidth]{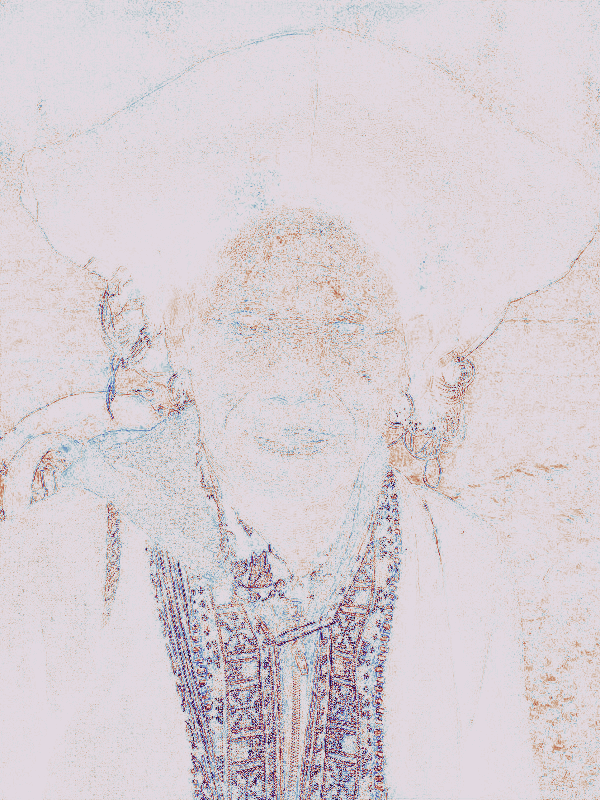}
    \includegraphics[width=\linewidth]{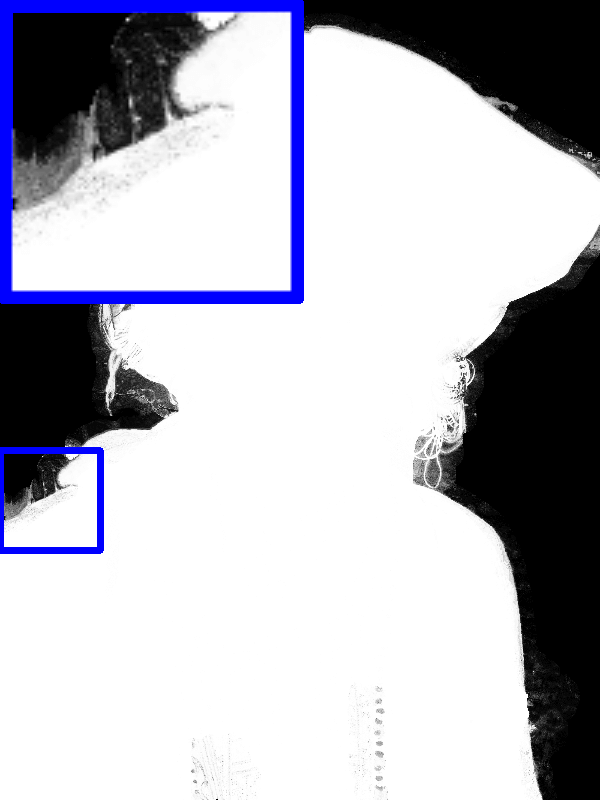}
    \includegraphics[width=\linewidth]{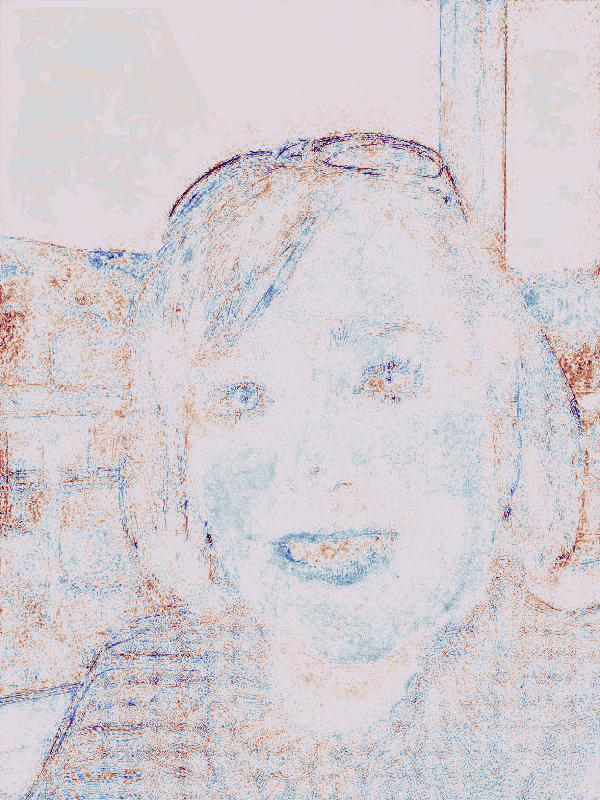}
    \includegraphics[width=\linewidth]{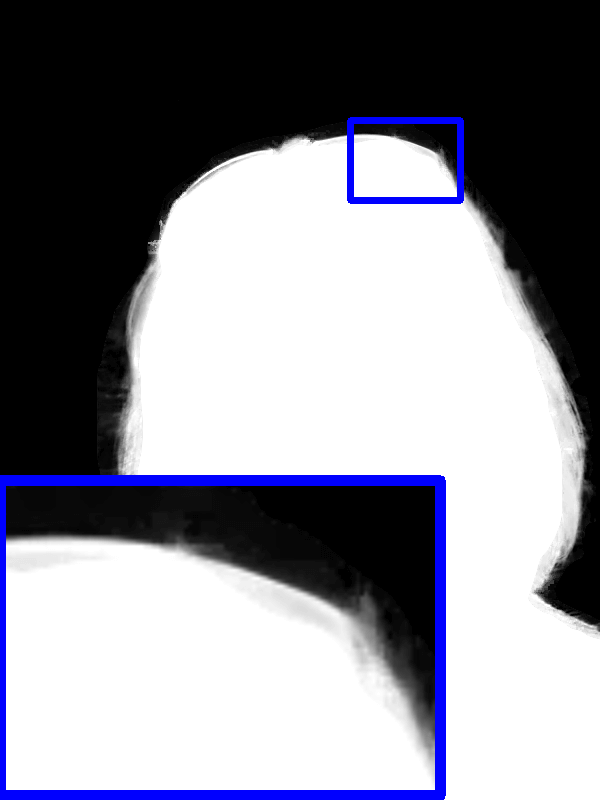}
    \end{minipage}
  }
  \hspace{-3.5mm}
  \subfloat[\scriptsize{Composition}]{\label{fig:analysis_e}
    \begin{minipage}{0.09\linewidth}
    \includegraphics[width=\linewidth]{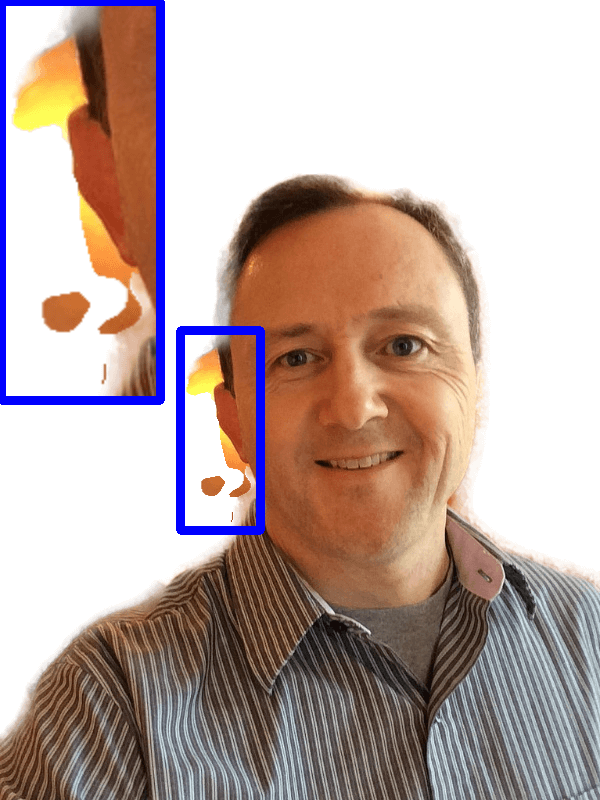}
    \includegraphics[width=\linewidth]{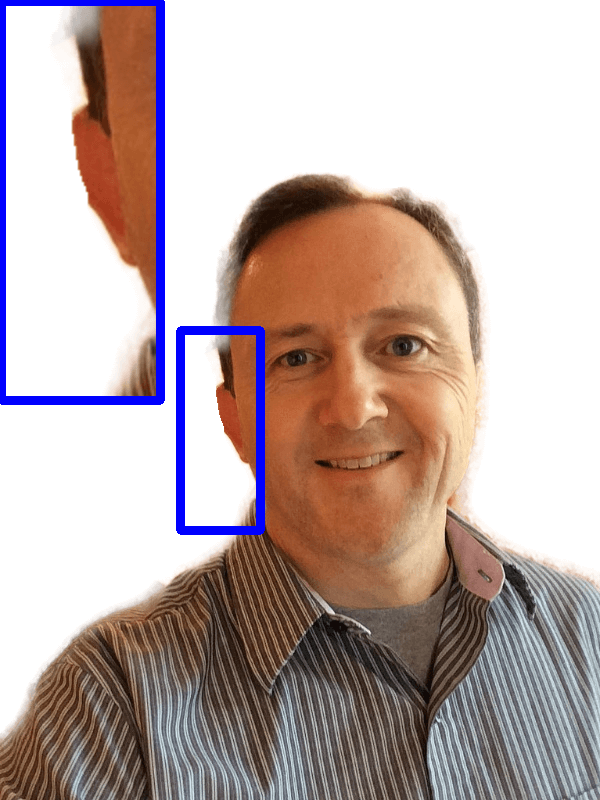}
    \includegraphics[width=\linewidth]{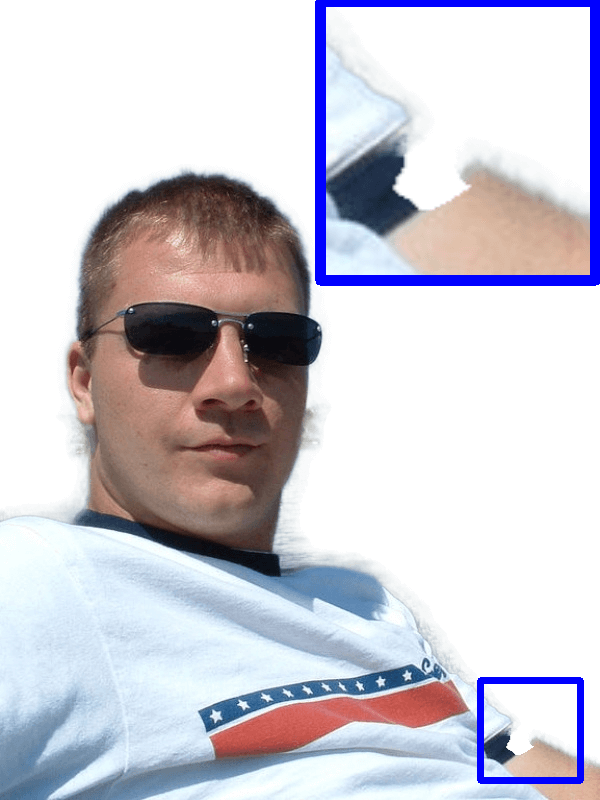}
    \includegraphics[width=\linewidth]{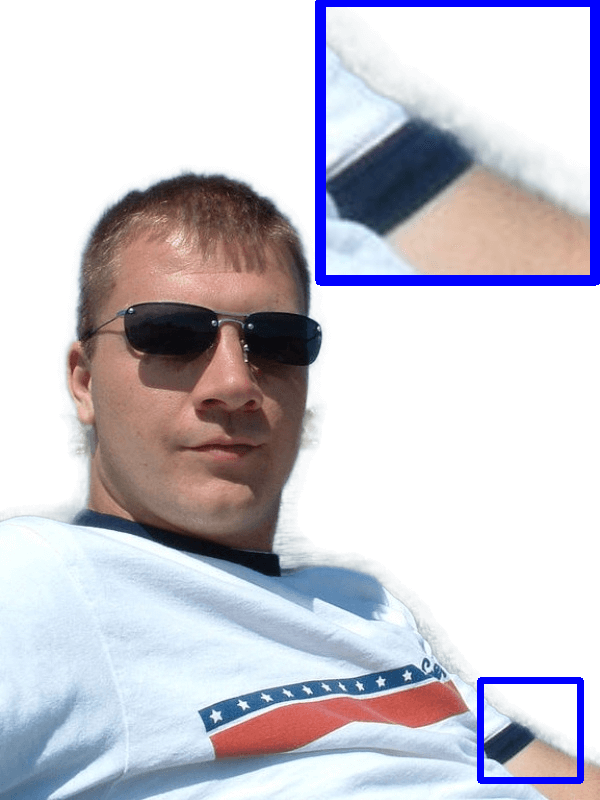}
    \includegraphics[width=\linewidth]{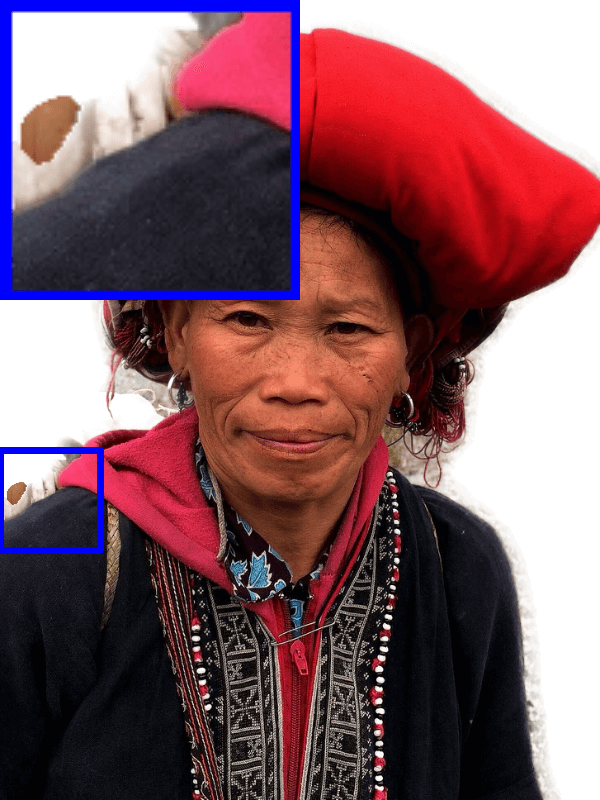}
    \includegraphics[width=\linewidth]{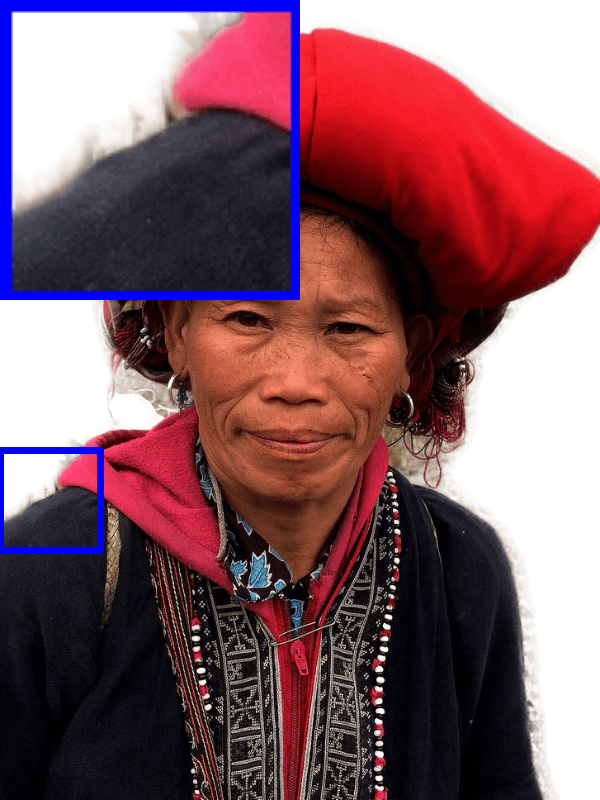}
    \includegraphics[width=\linewidth]{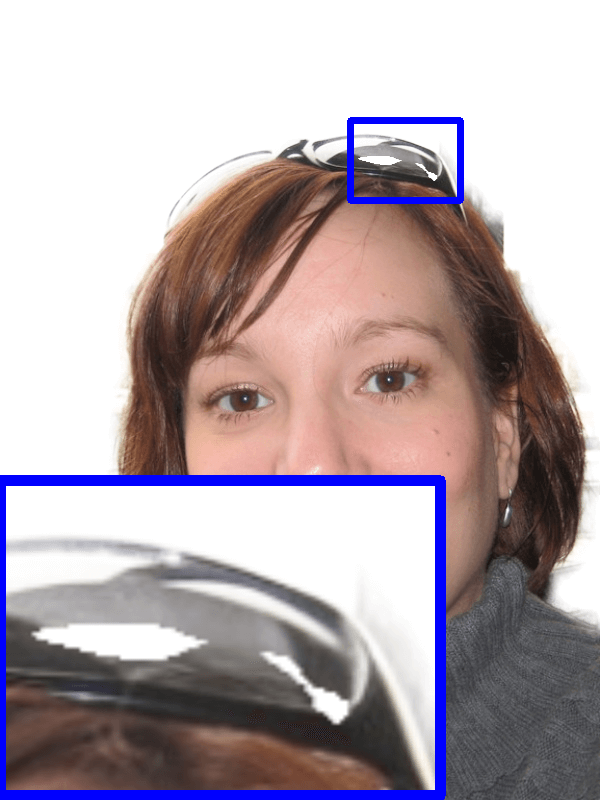}
    \includegraphics[width=\linewidth]{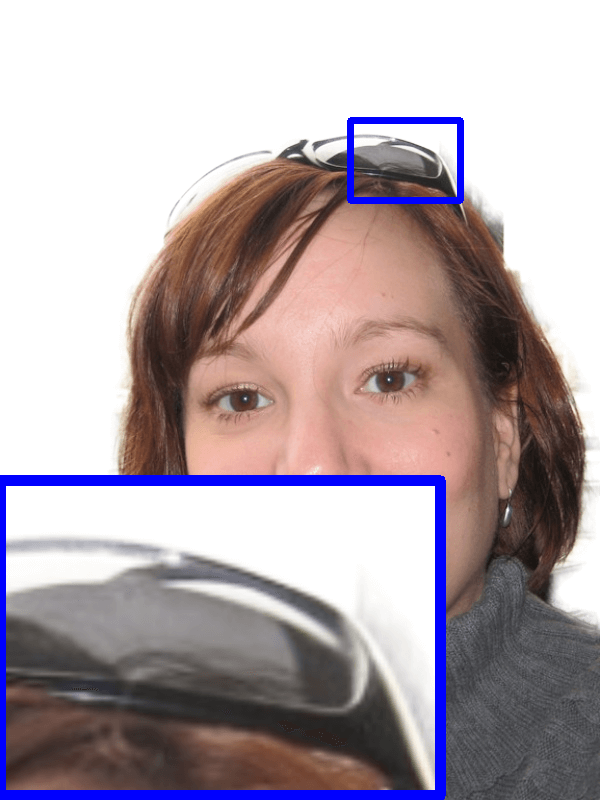}
    \end{minipage}
    }
  \hspace{1mm}
  \subfloat[\scriptsize{Portrait}]{\label{fig:analysis_f}
    \begin{minipage}{0.09\linewidth}
      \includegraphics[width=\linewidth]{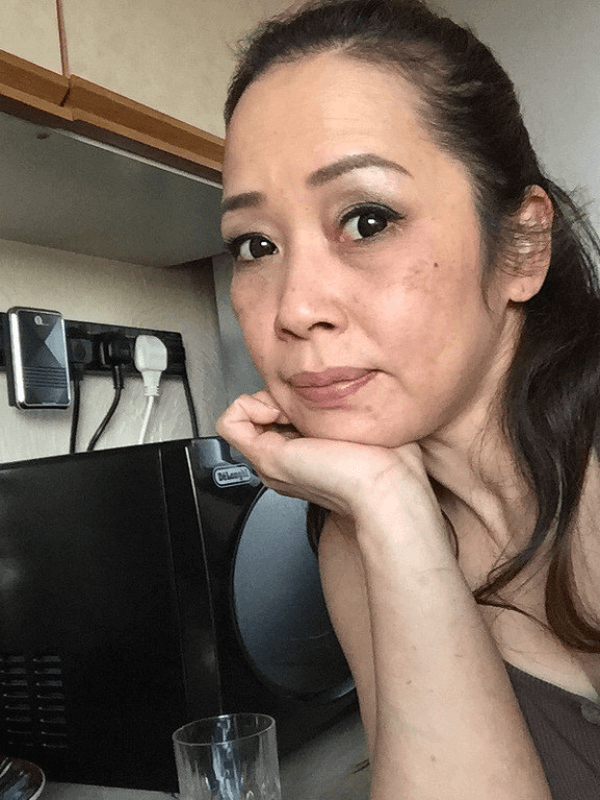}
      \includegraphics[width=\linewidth]{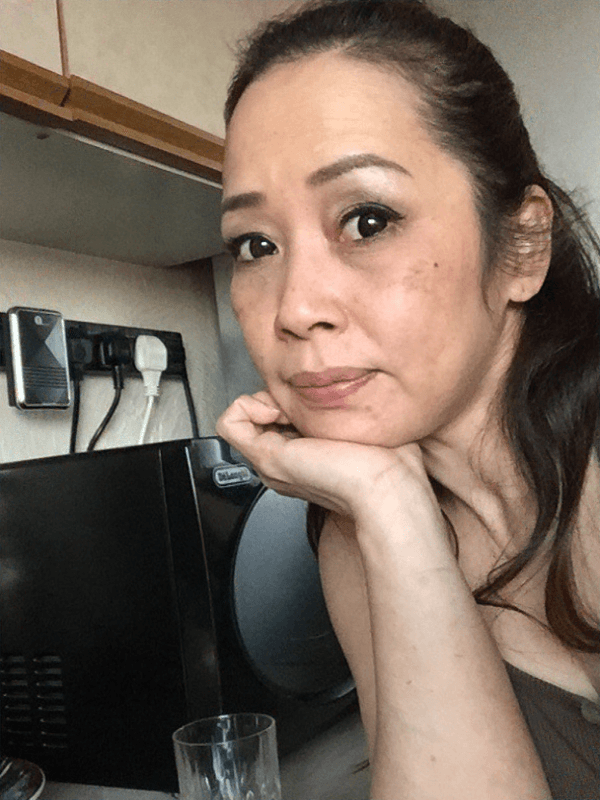}
      \includegraphics[width=\linewidth]{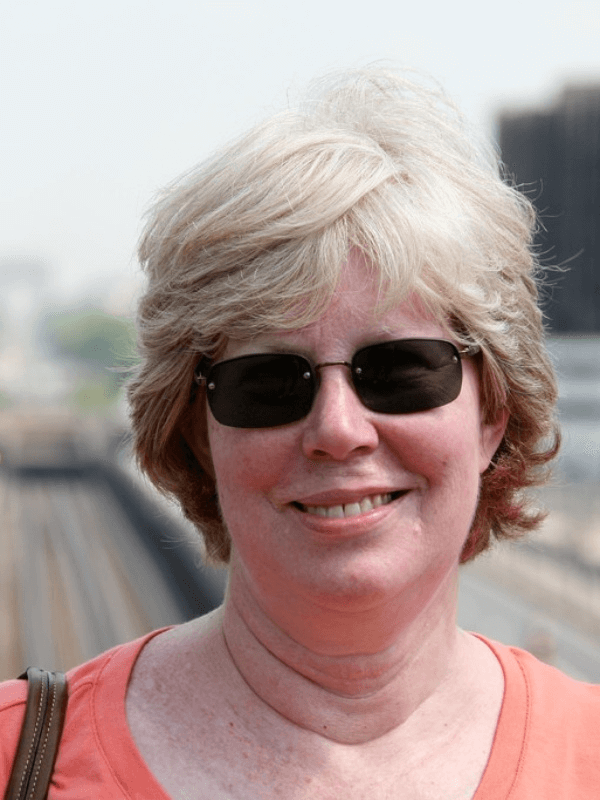}
      \includegraphics[width=\linewidth]{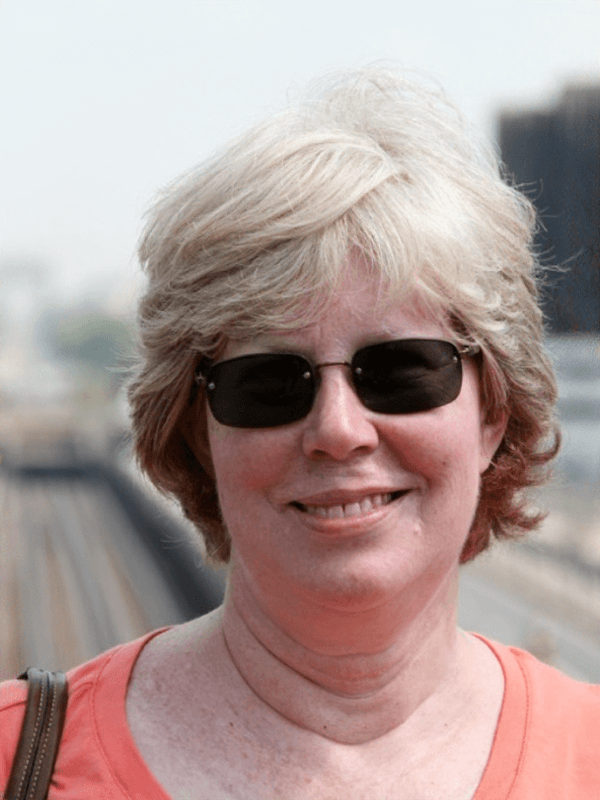}
      \includegraphics[width=\linewidth]{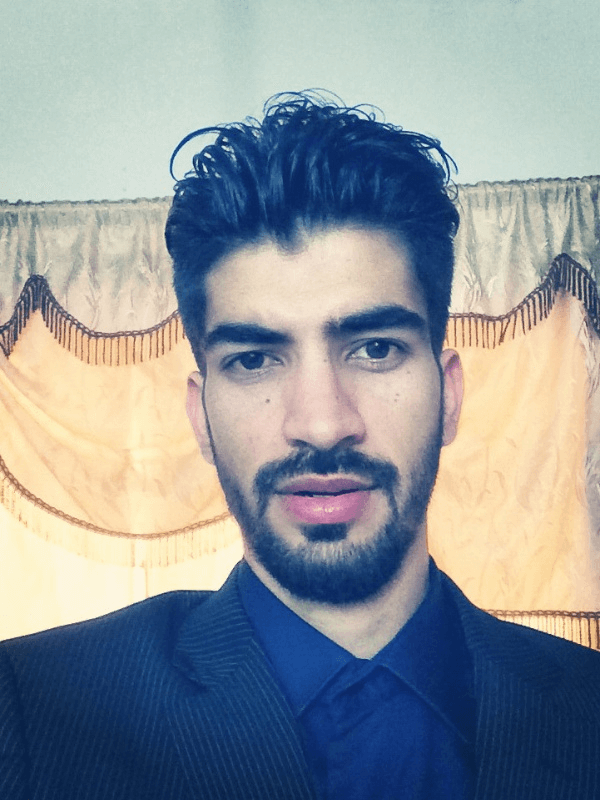}
      \includegraphics[width=\linewidth]{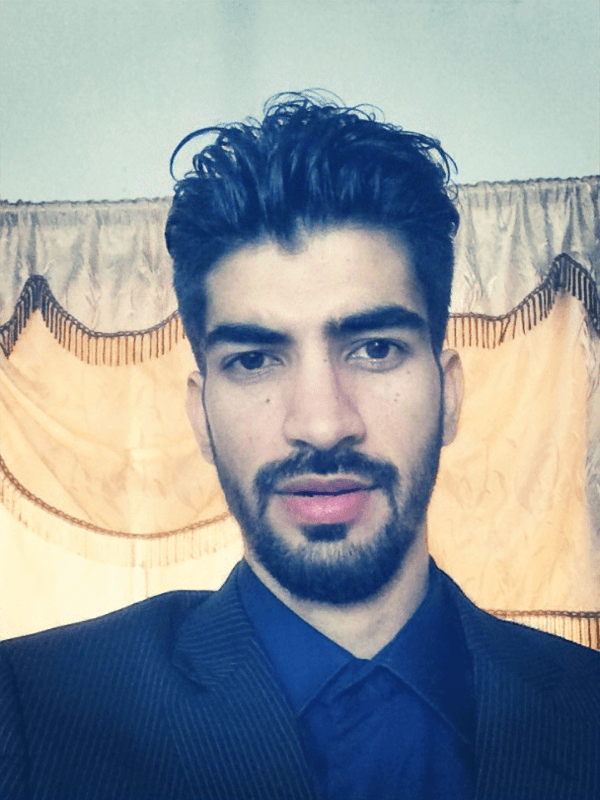}
      \includegraphics[width=\linewidth]{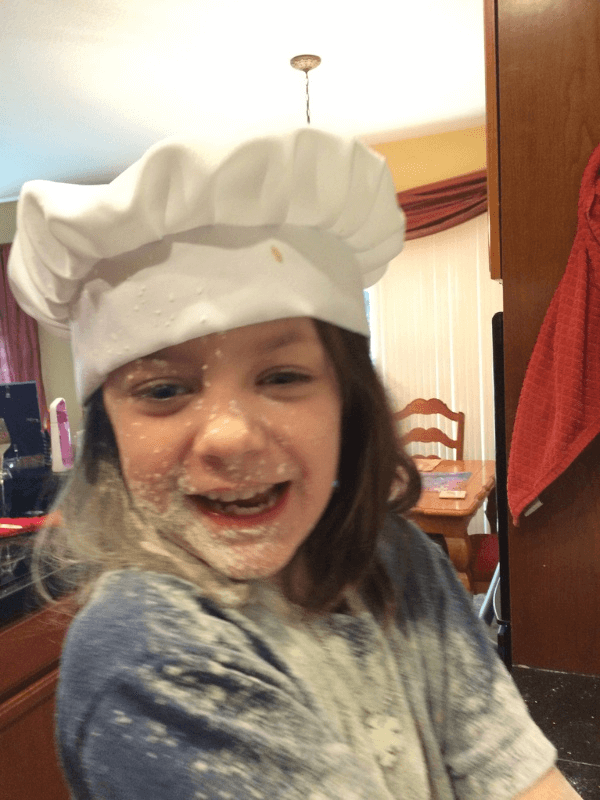}
      \includegraphics[width=\linewidth]{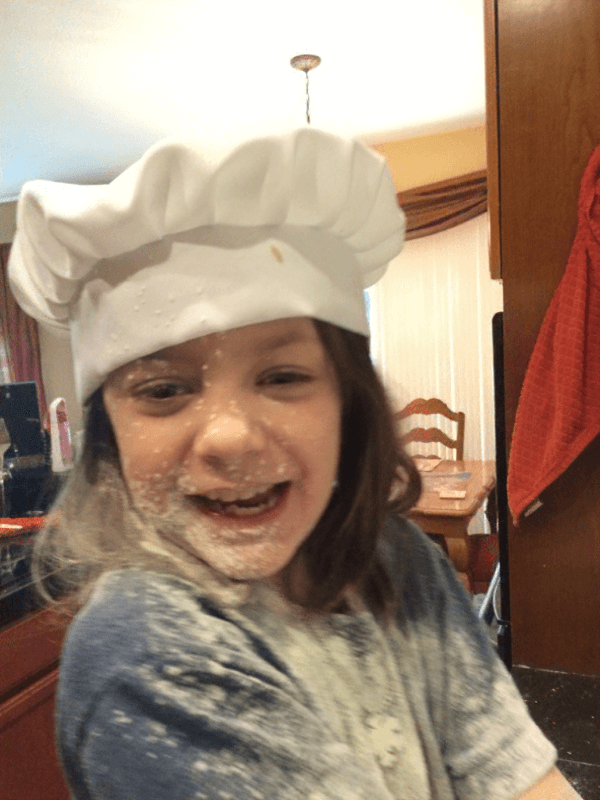}
    \end{minipage}
    }
  \hspace{-3.5mm}
  \subfloat[\scriptsize{Trimap}]{\label{fig:analysis_g}
    \begin{minipage}{0.09\linewidth}
      \includegraphics[width=\linewidth]{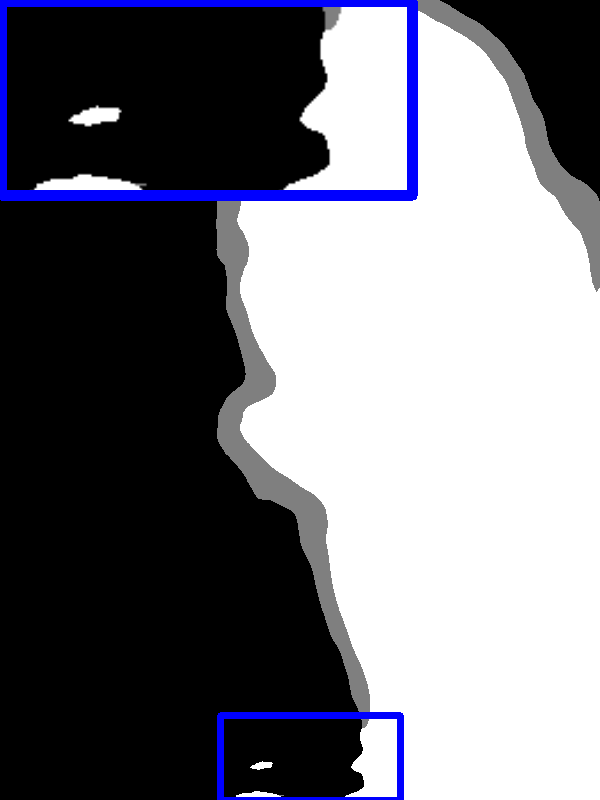}
      \includegraphics[width=\linewidth]{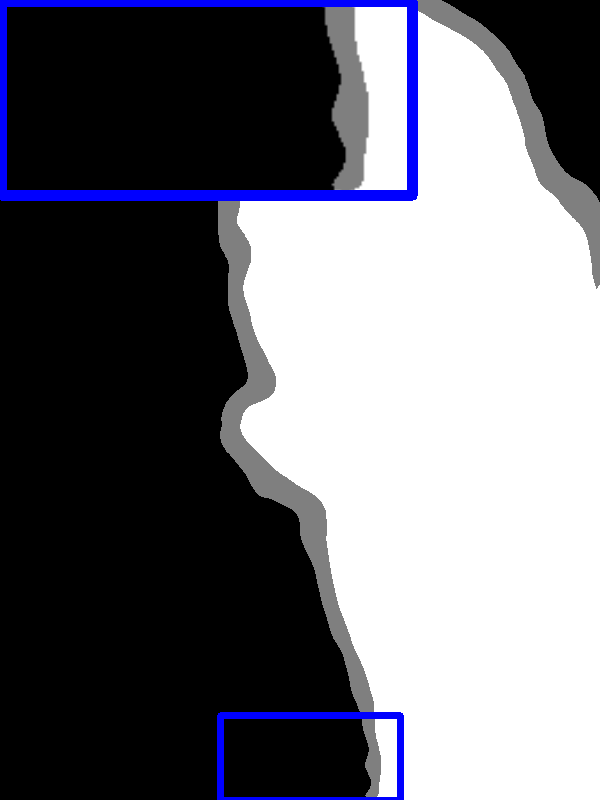}
      \includegraphics[width=\linewidth]{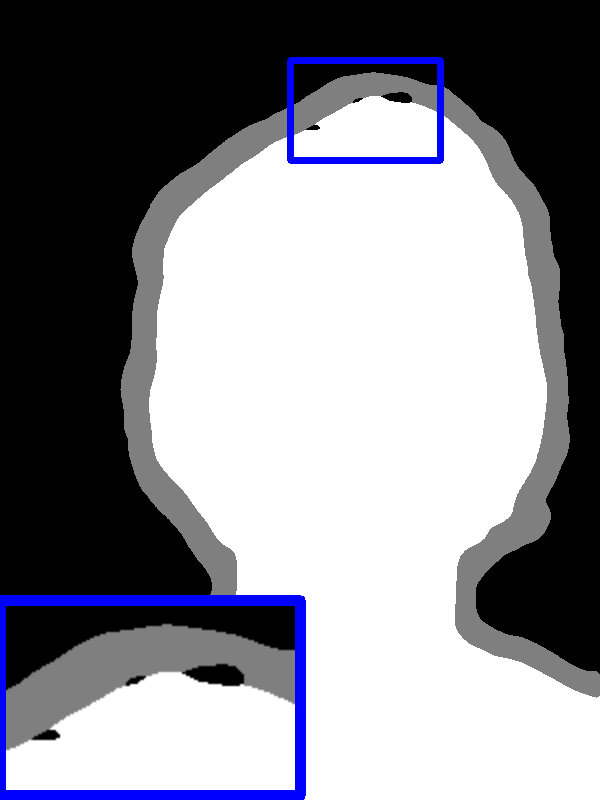}
      \includegraphics[width=\linewidth]{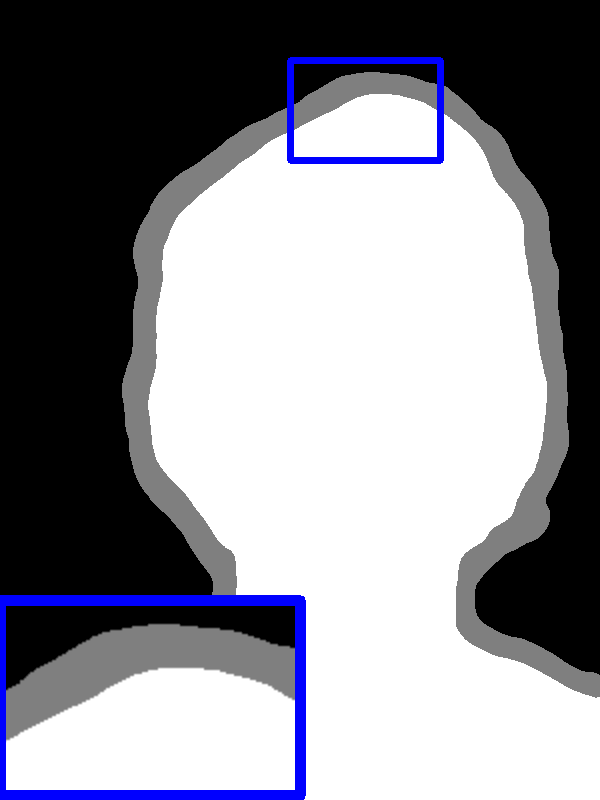}
      \includegraphics[width=\linewidth]{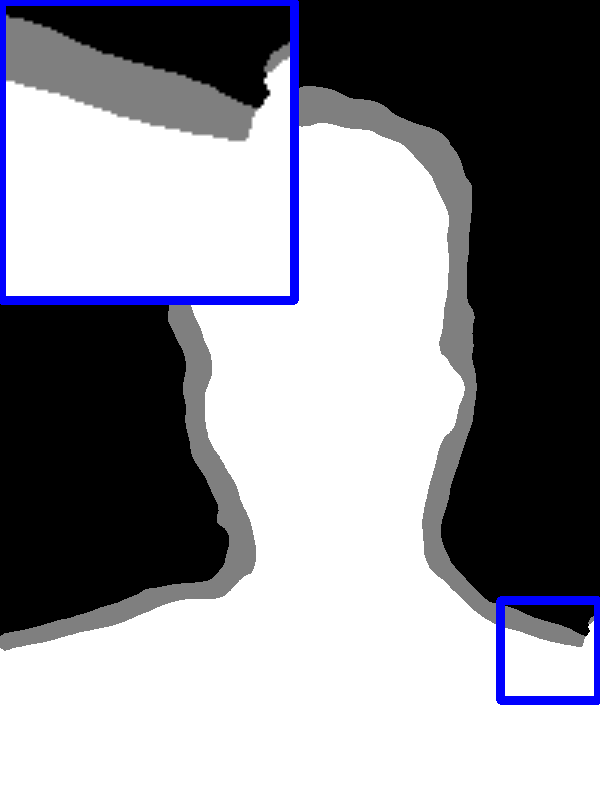}
      \includegraphics[width=\linewidth]{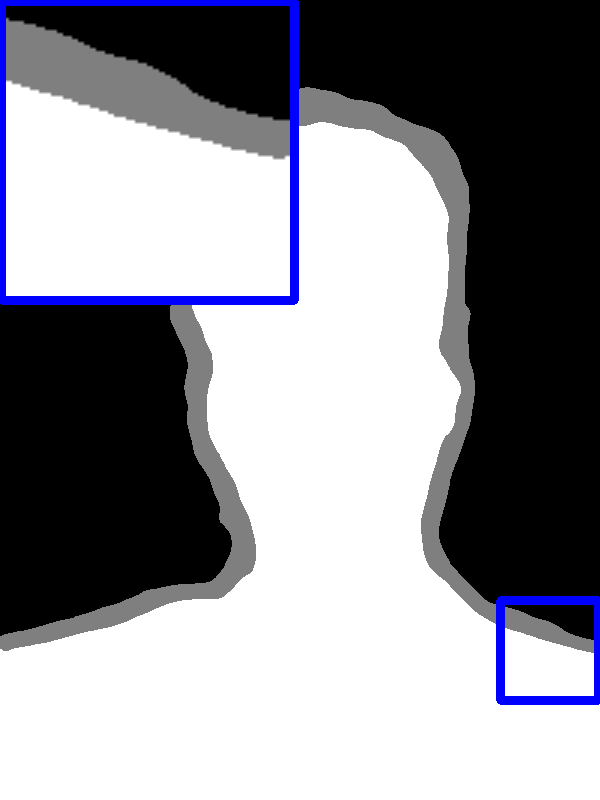}
      \includegraphics[width=\linewidth]{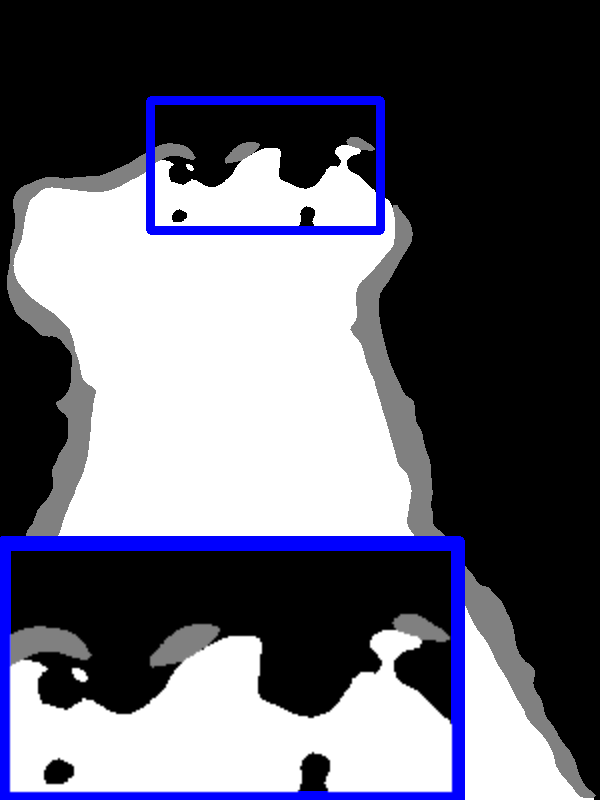}
      \includegraphics[width=\linewidth]{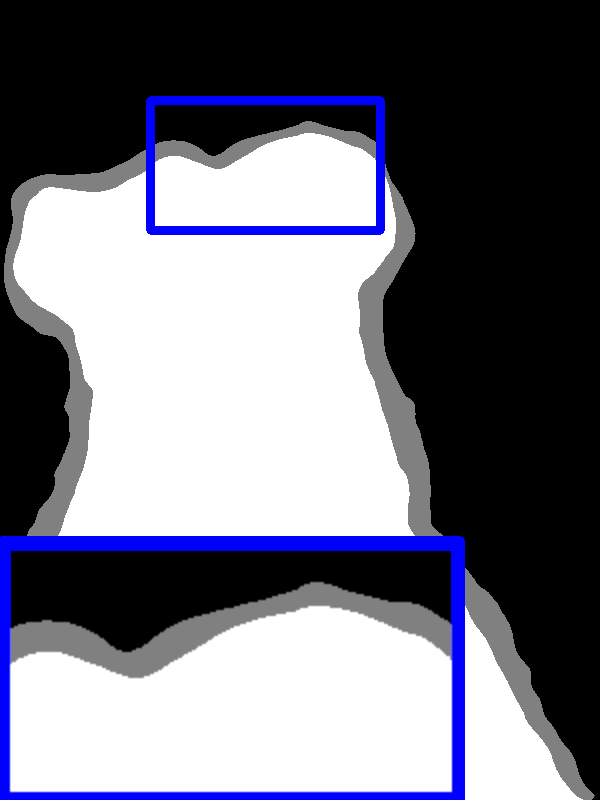}
    \end{minipage}
  }
  \hspace{-3.5mm}
  \subfloat[\scriptsize{Matte}]{\label{fig:analysis_h}
    \begin{minipage}{0.09\linewidth}
      \includegraphics[width=\linewidth]{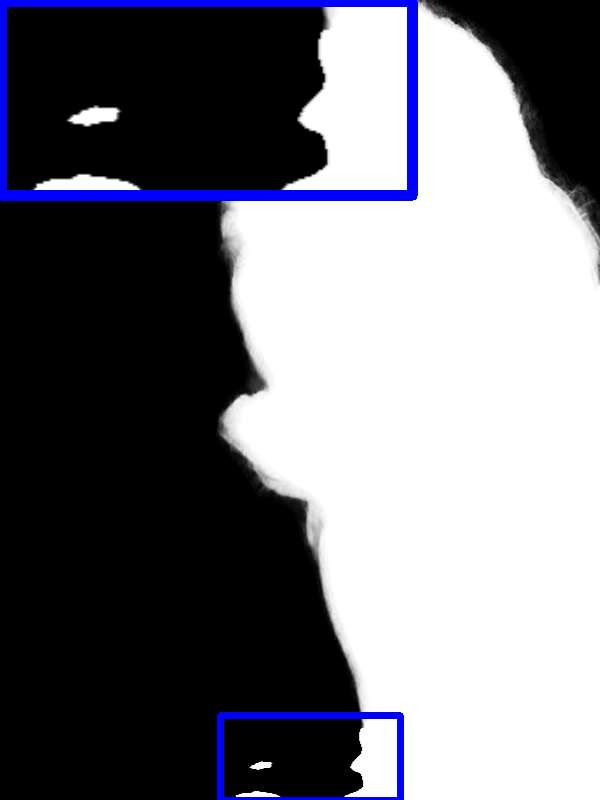}
      \includegraphics[width=\linewidth]{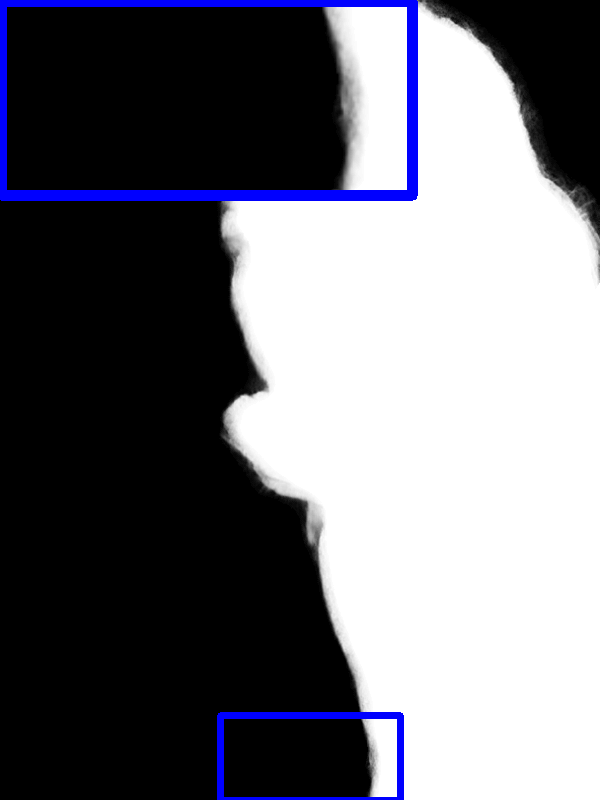}
      \includegraphics[width=\linewidth]{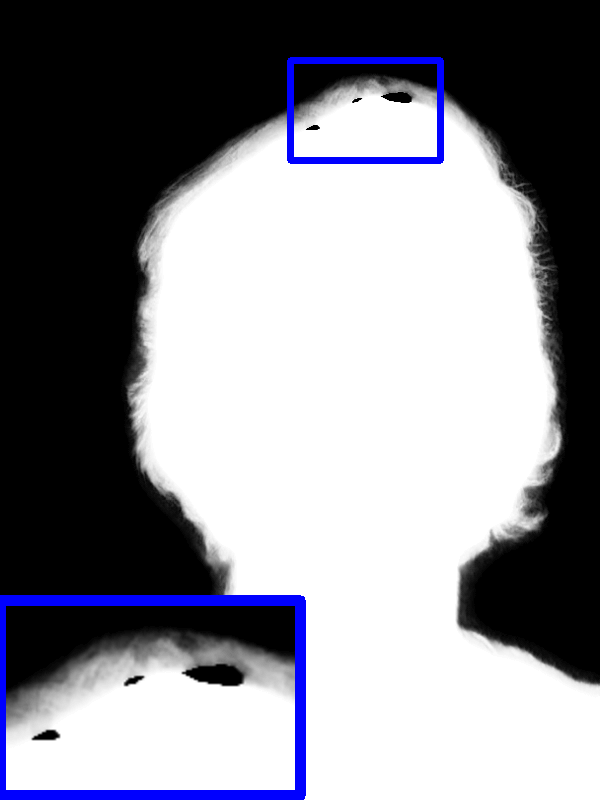}
      \includegraphics[width=\linewidth]{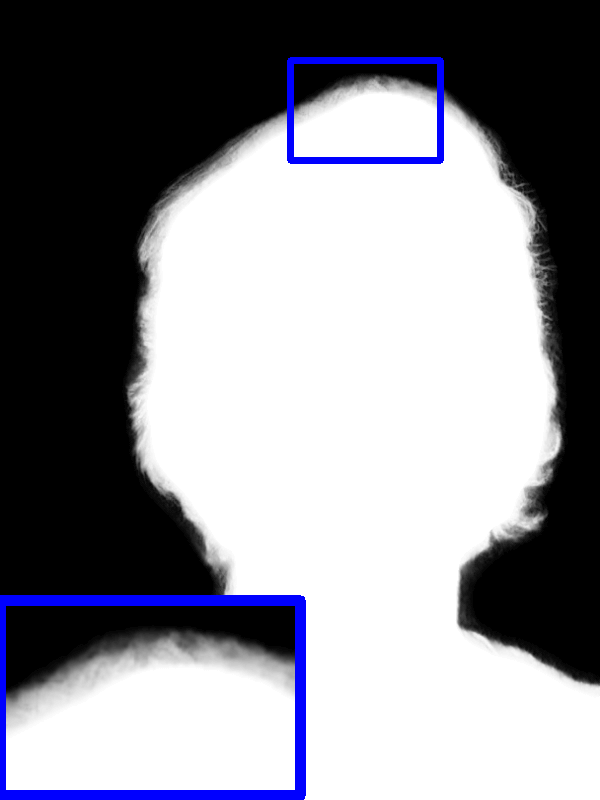}
      \includegraphics[width=\linewidth]{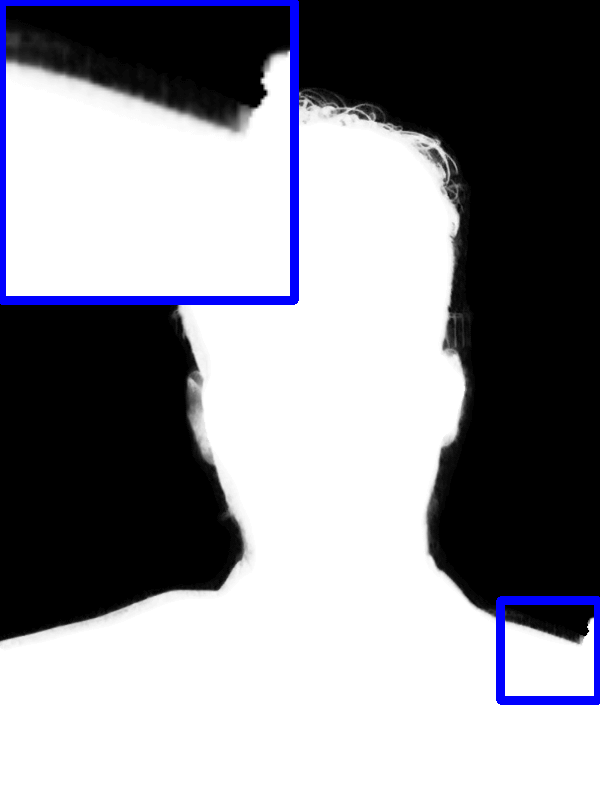}
      \includegraphics[width=\linewidth]{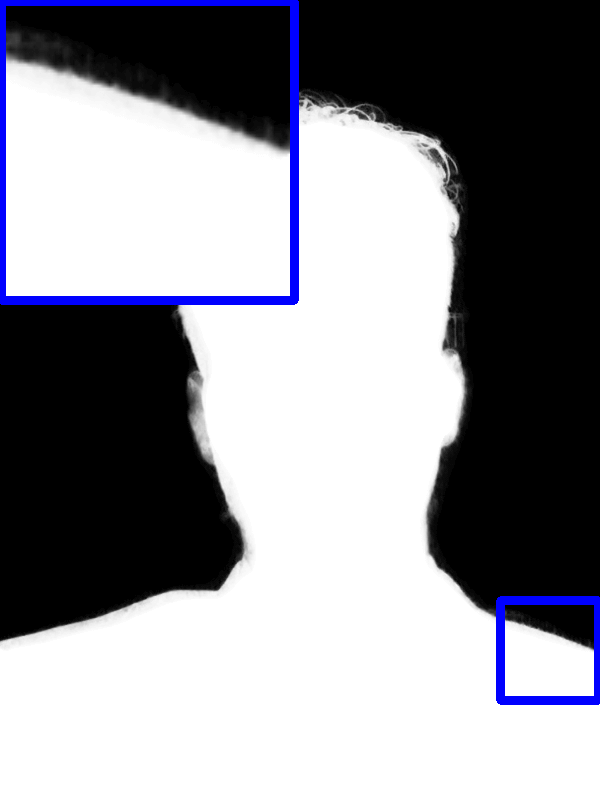}
      \includegraphics[width=\linewidth]{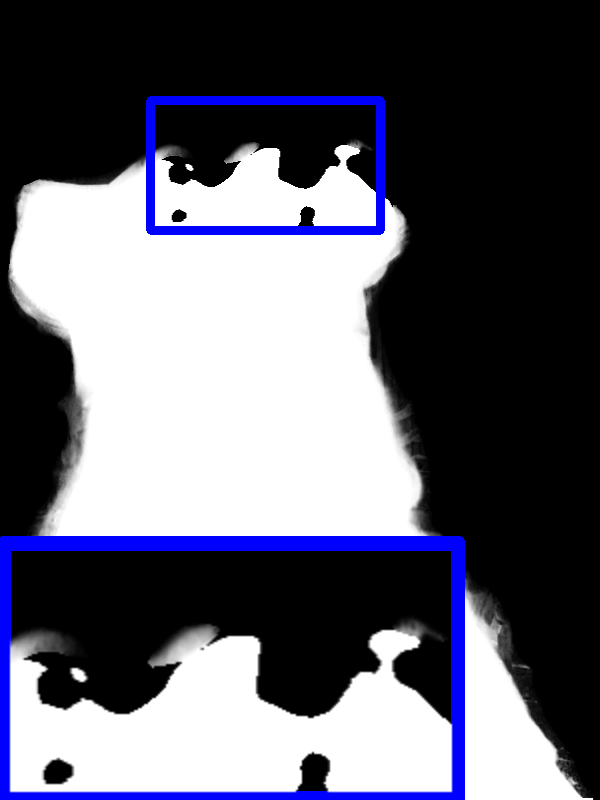}
      \includegraphics[width=\linewidth]{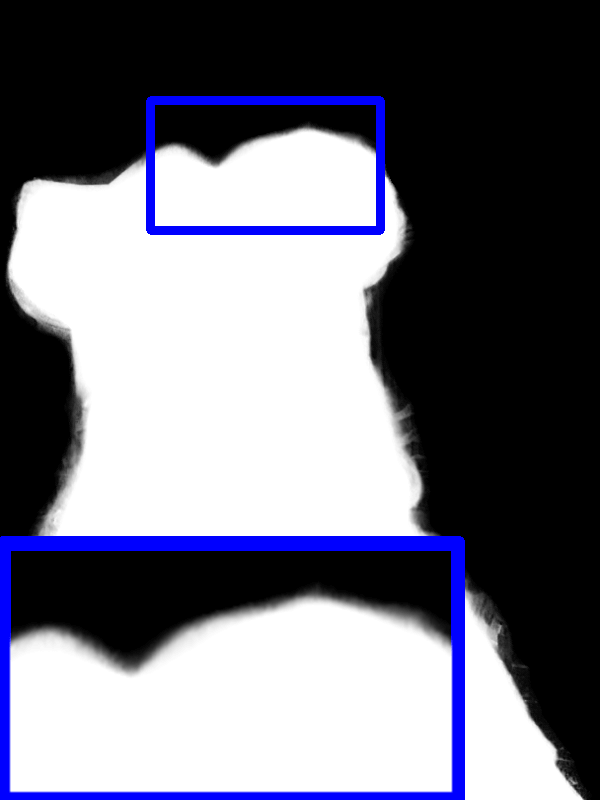}
    \end{minipage}
    }
  \hspace{-3.5mm}
  \subfloat[\scriptsize{Pert.\& GT}]{\label{fig:analysis_i}
    \begin{minipage}{0.09\linewidth}
      \includegraphics[width=\linewidth]{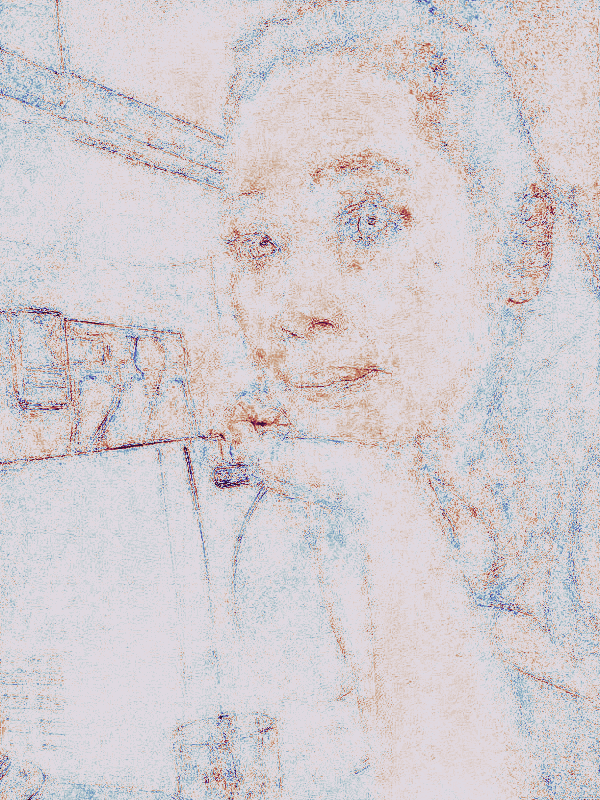}
      \includegraphics[width=\linewidth]{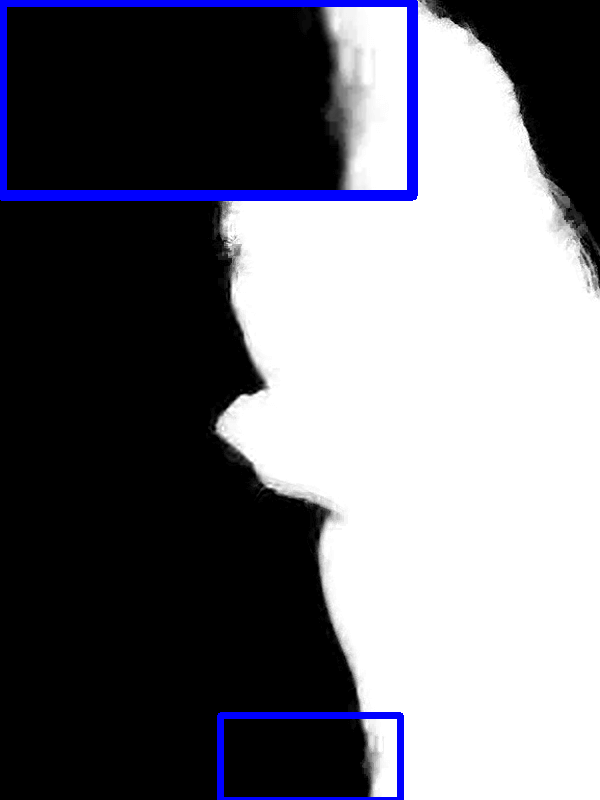}
      \includegraphics[width=\linewidth]{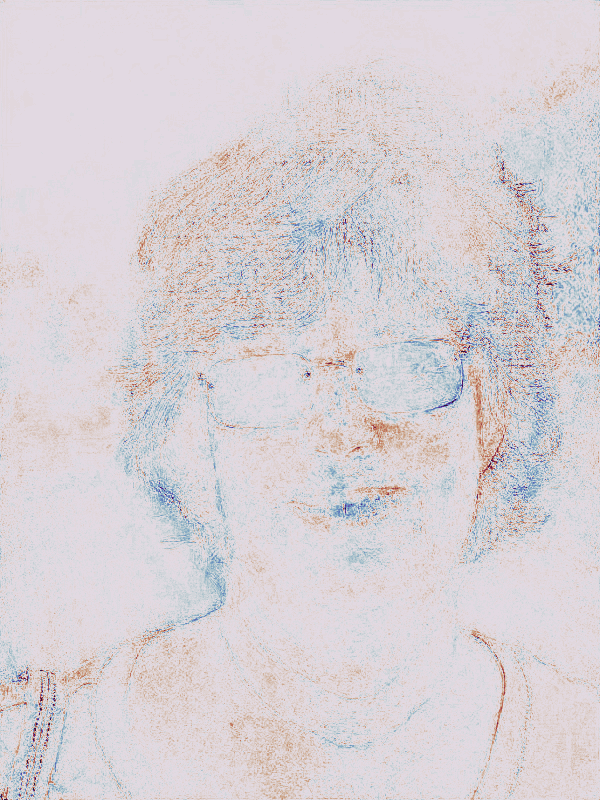}
      \includegraphics[width=\linewidth]{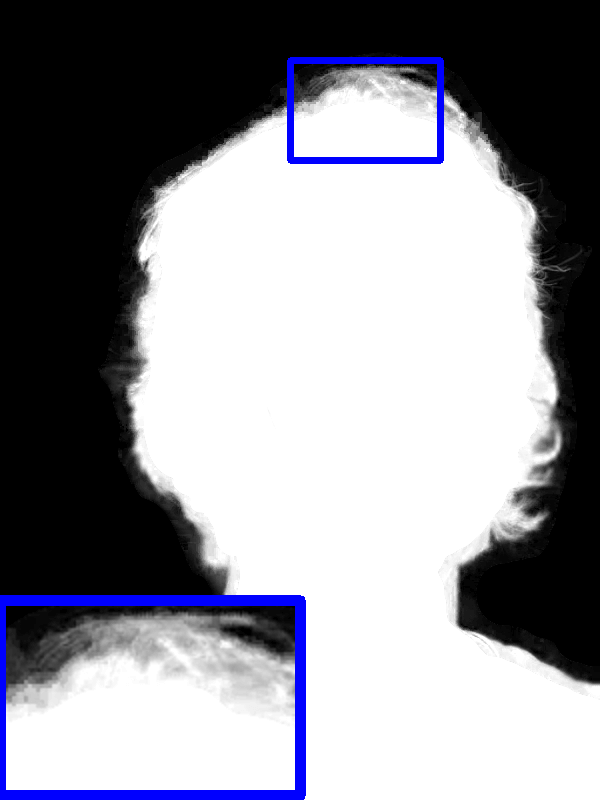}
      \includegraphics[width=\linewidth]{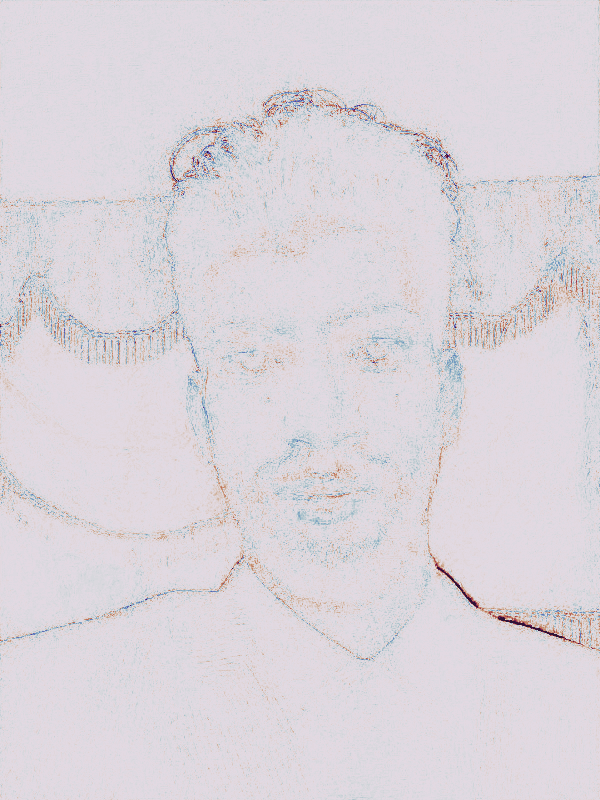}
      \includegraphics[width=\linewidth]{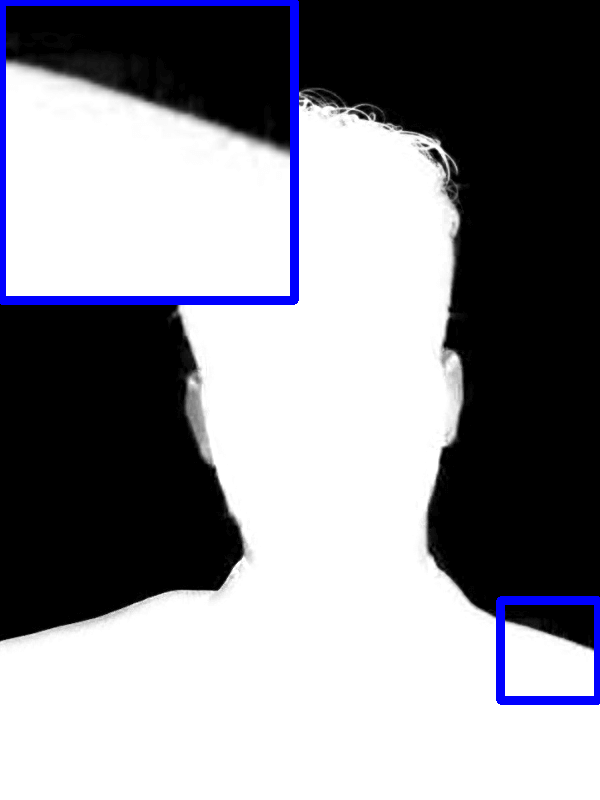}
      \includegraphics[width=\linewidth]{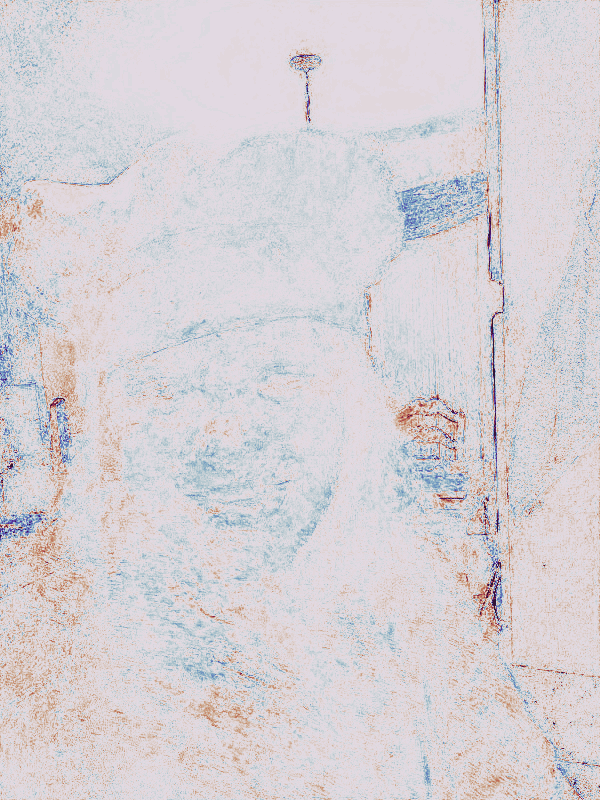}
      \includegraphics[width=\linewidth]{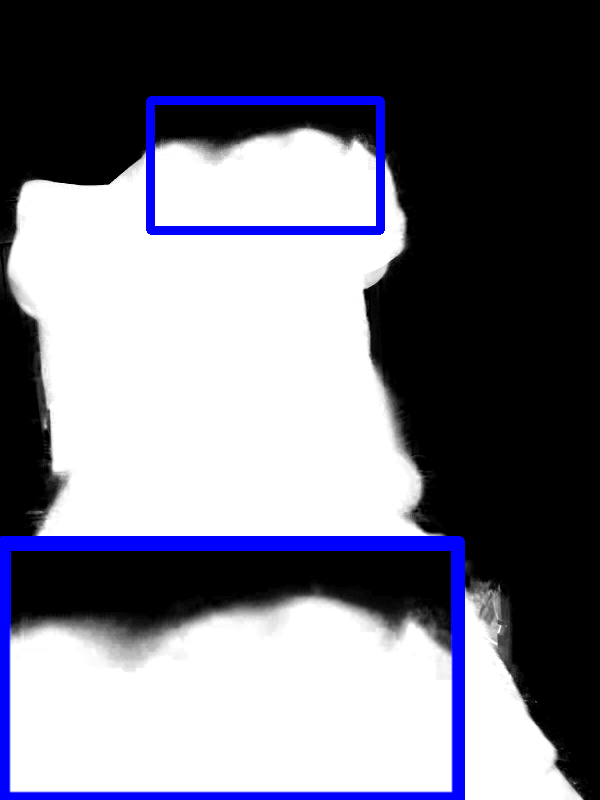}
    \end{minipage}
  }
  \hspace{-3.5mm}
  \subfloat[\scriptsize{Composition}]{\label{fig:analysis_j}
    \begin{minipage}{0.09\linewidth}
      \includegraphics[width=\linewidth]{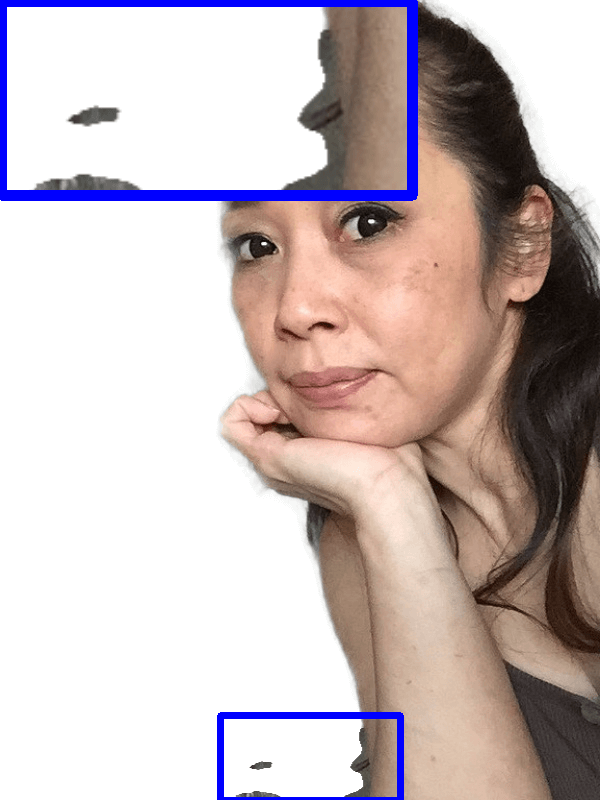}
      \includegraphics[width=\linewidth]{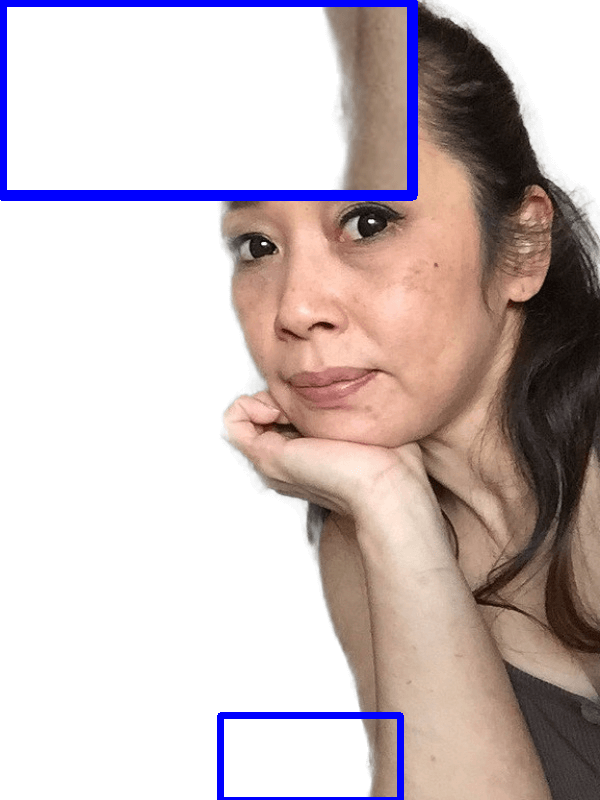}
      \includegraphics[width=\linewidth]{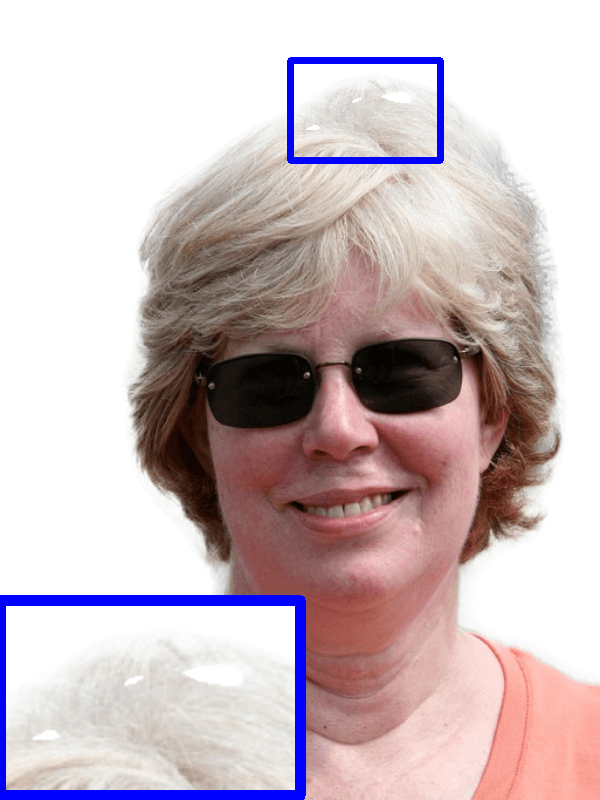}
      \includegraphics[width=\linewidth]{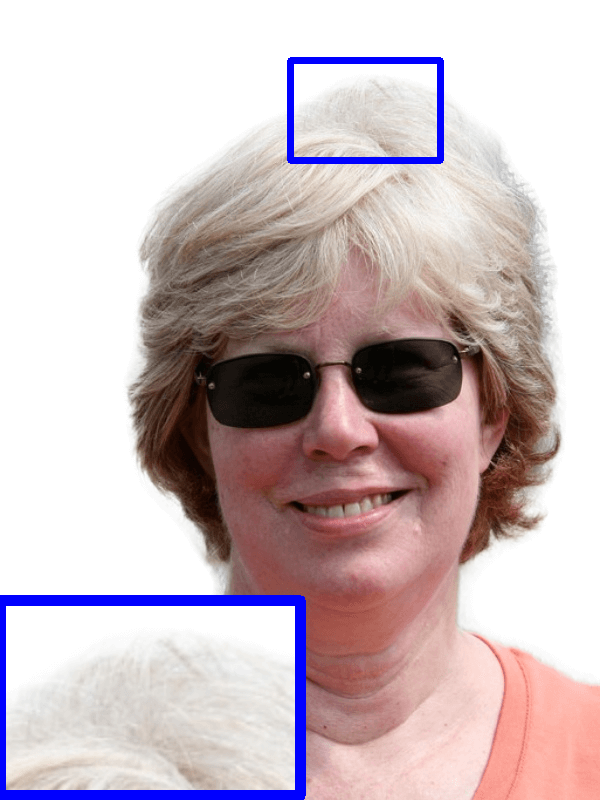}
      \includegraphics[width=\linewidth]{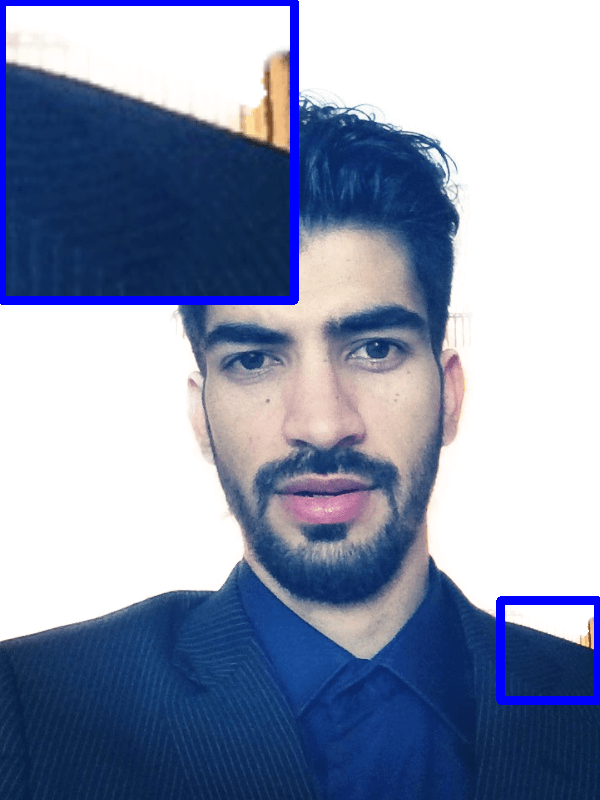}
      \includegraphics[width=\linewidth]{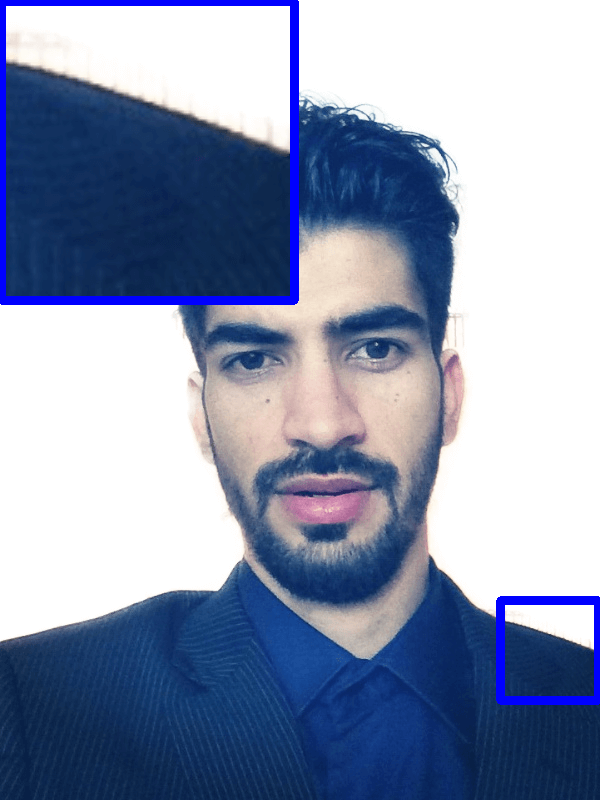}
      \includegraphics[width=\linewidth]{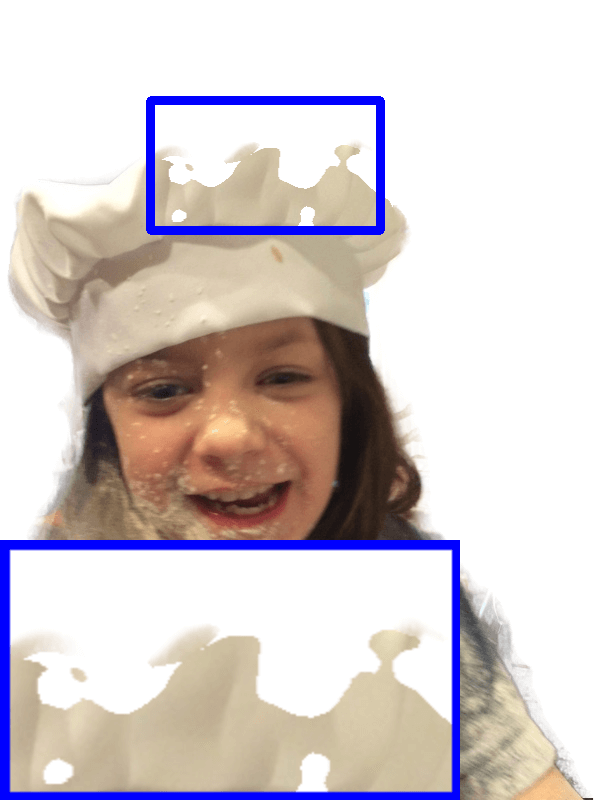}
      \includegraphics[width=\linewidth]{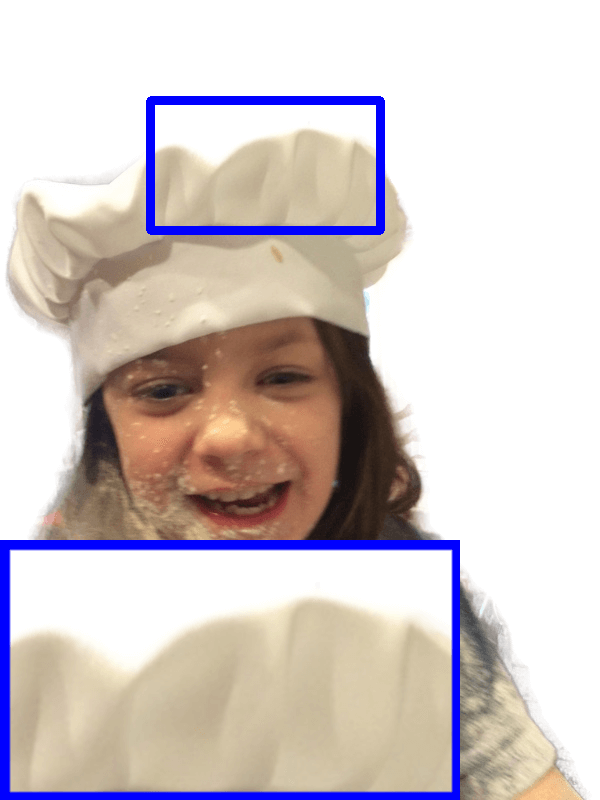}
    \end{minipage}
    }
  \hspace{-3.5mm}
  \caption{Qualitative evaluation on portrait enhancement. The odd row images are original portraits with their corresponding predicted trimap, alpha matte, perturbations (rescaled and remapped), and the composite image (with a white background). The even row images show the enhanced results and ground truth (d).}
  \label{fig:result1}
\end{figure*}

\subsection{Evaluation on Portrait Enhancement} 
In this section, we evaluate the effectiveness of the proposed portrait enhancement method. The performances of matting models on the original portraits are set as baselines.
Note that we froze all parameters of the matting model and only optimize the latent space to find the optimal high-dimensional representation that most compatible with these matting models. We show both quantitative and qualitative results in the following subsections.

\subsubsection{Quantitative Results}

Quantitative evaluation on the proposed portrait enhancement approach is shown in Tab.~\ref{table:1}. A clear performance boosting can be found for all the methods on both two datasets. Particularly, perceptual-based metrics Grad and Conn reflect the enhancement property of our method. Both these two metrics focus on evaluating continuity of the generated alpha mattes, which is the main source affects the human perception of the new composite images. As can be seen, the proposed enhancement improves existing methods by a large margin for all trimap-based matting models, \eg, an average Grad improvement of 11.32\% on portrait dataset and 20.732\% on adobe matting dataset has been obtained. This can be specifically attributed to our entropy minimization strategy and compositional loss, as discontinuous regions are mostly with a low-degree of confidence and can be easily detected in the composite images. As a result, the proposed enhancement mostly corrects discontinuity (as shown in Sec.~\ref{sec:result}) and more favorable to human perception. In terms of error measurement, both the MSE and SAD achieves at least 4\% improvement for all the six methods on the portrait dataset. In particular, although the room for MSE improvement is small, the best two methods DIM and IndexNet increase by more than 10\%. SAD metric shows a larger performance gain in number (0.933 improvement for IndexNet). These statistical improvements demonstrate that the proposed enhancement approach removes prediction outliers very well. \rev{Given the predicted trimap in the trimap-free method MODNet~\cite{ke2020modnet}, our method can also improve its performance on all metrics.} Overall, the proposed enhancement shows great improvements on different metrics (up to 10\% for most metrics), and also demonstrates the generality for state-of-the-art methods (most methods are improved by a large margin).

\begin{table*}[h]
  \renewcommand\arraystretch{1.2}
         \caption{Quantitative evaluation on portrait augmentation with respect to \rev{five} matting models on two datasets. {``FT'' represents the results of fine-tuned model.} The lower the better for all metrics. Top 2 improved methods (in percentage) for each metric are marked in \textcolor[rgb]{0.00,0.07,1.00}{blue} and \textcolor[rgb]{0.00,0.59,0.00}{green} respectively.}
          \begin{center}
          \setlength{\tabcolsep}{0.125cm}{
          \begin{tabular}{c|c|c|c|c|c|c|c|c|c|c|c|c}
              \hline
              \multirow{2}{*}{\diagbox{Models}{Metrics}}     &\multicolumn{3}{c|}{Grad$\downarrow$}  &\multicolumn{3}{c|}{Conn$\downarrow$}  &\multicolumn{3}{c|}{MSE$\downarrow$}  &\multicolumn{3}{c}{SAD$\downarrow$} \\
              \cline{2-13}
                                                                &Ori.         &FT    & $\Delta$ / $|$\%$\Delta$$|$
                                                                &Ori.         &FT    & $\Delta$ / $|$\%$\Delta$$|$
                                                                &Ori.         &FT    & $\Delta$ / $|$\%$\Delta$$|$
                                                                &Ori.         &FT    & $\Delta$ / $|$\%$\Delta$$|$ \\
         \hline

        \multicolumn{13}{c}{Portrait Matting Dataset~\cite{shen2016deep}} \\
        \hline
        DIM~\cite{xu2017deep}                                     &8.699        &8.212        &\textcolor[rgb]{0.00,0.07,1.00}{-0.487/5.60\%}
                                                                  &12.291       &11.721       &\textcolor[rgb]{0.00,0.07,1.00}{-0.570/4.64\%}
                                                                  &0.017        &0.016        &-0.001/5.88\%
                                                                  &13.269       &12.973       &-0.296/2.23\%  \\

        \rowcolor{gray!20}IndexNet~\cite{lu2019indices}           &7.405        &7.029        &\textcolor[rgb]{0.00,0.59,0.00}{-0.376/5.08\%}
                                                                  &11.345       &10.933       &\textcolor[rgb]{0.00,0.59,0.00}{-0.412/3.63\%}
                                                                  &0.016        &0.015        &\textcolor[rgb]{0.00,0.59,0.00}{0.001/6.25\%}
                                                                  &11.826       &11.472       &\textcolor[rgb]{0.00,0.07,1.00}{-0.354/2.99\%} \\

        GCA~\cite{li2020natural}                                  &9.170        &8.913        &-0.257/2.80\%
                                                                  &11.382       &11.131       &-0.251/2.21\%
                                                                  &0.017        &0.016        &{-0.001/5.88\%}
                                                                  &11.751       &11.428       &\textcolor[rgb]{0.00,0.59,0.00}{-0.323/2.75\%} \\

        \rowcolor{gray!20}AdaMatting~\cite{cai2019disentangled}   &21.882       &21.037       &-0.845/3.86\%
                                                                  &18.291       &17.776       &-0.515/2.82\%
                                                                  &0.029        &0.027        &\textcolor[rgb]{0.00,0.07,1.00}{0.002/6.90\%}
                                                                  &18.225       &17.854       &-0.371/2.04\%   \\

        \rev{MODNet~\cite{ke2020modnet}}                              &\rev{22.436}   &\rev{22.033}   &\rev{-0.403/1.79\%}
                                                                      &\rev{17.523}   &\rev{17.374}   &\rev{-0.149/0.85\%}
                                                                      &\rev{0.028}    &\rev{0.027}    &\rev{-0.001/3.57\%}
                                                                      &\rev{19.359}   &\rev{18.871}   &\rev{-0.488/2.52\%}\\
                                                           
        \hline
        \multicolumn{13}{c}{Adobe Image Matting Dataset~\cite{xu2017deep}} \\
        \hline
        DIM~\cite{xu2017deep}                                     &21.718       &21.163       &-0.555/2.56\%
                                                                  &23.364       &22.775       &\textcolor[rgb]{0.00,0.07,1.00}{-0.589/2.52\%}
                                                                  &0.038        &0.037        &\textcolor[rgb]{0.00,0.59,0.00}{-0.001/2.63\%}
                                                                  &23.417       &23.056       &-0.361/1.54\%   \\

        \rowcolor{gray!20}IndexNet~\cite{lu2019indices}           &21.268       &20.224       &\textcolor[rgb]{0.00,0.59,0.00}{-1.044/4.91\%}
                                                                  &22.542       &22.111       &{-0.431/1.91\%}
                                                                  &0.038        &0.037        &\textcolor[rgb]{0.00,0.59,0.00}{-0.001/2.63\%}
                                                                  &22.701       &22.123       &{-0.578/2.54\%} \\

        GCA~\cite{li2020natural}                                  &22.489       &21.117       &\textcolor[rgb]{0.00,0.07,1.00}{-1.372/6.10\%}
                                                                  &22.106       &21.652       &\textcolor[rgb]{0.00,0.59,0.00}{-0.454/2.05\%}
                                                                  &0.038        &0.037        &\textcolor[rgb]{0.00,0.59,0.00}{-0.001/2.63\%}
                                                                  &22.205       &21.142       &\textcolor[rgb]{0.00,0.07,1.00}{-0.163/4.79\%}   \\

        \rowcolor{gray!20}AdaMatting~\cite{cai2019disentangled}   &24.934       &23.775       &-1.159/4.65\%
                                                                  &23.274       &22.876       &{-0.398/1.71\%}
                                                                  &0.041        &0.039        &\textcolor[rgb]{0.00,0.07,1.00}{-0.002/5.88\%}
                                                                  &25.613       &24.487       &\textcolor[rgb]{0.00,0.59,0.00}{-1.126/4.04\%}   \\

        \rev{MODNet~\cite{ke2020modnet}}                          &\rev{25.042}   &\rev{24.836}   &\rev{-0.206/0.82\%}
                                                                  &\rev{24.433}   &\rev{24.023}   &\rev{-0.530/1.68\%}
                                                                  &\rev{0.045}    &\rev{0.044}    &\rev{-0.001/2.22\%}
                                                                  &\rev{27.335}   &\rev{27.033}   &\rev{-0.302/1.10\%}\\

        \hline
        \end{tabular}
        }
        \end{center}
      \label{table:g}
\end{table*}

\begin{figure*}[h]
   \centering
   {
     \begin{minipage}{0.099\linewidth}
     \includegraphics[width=\linewidth]{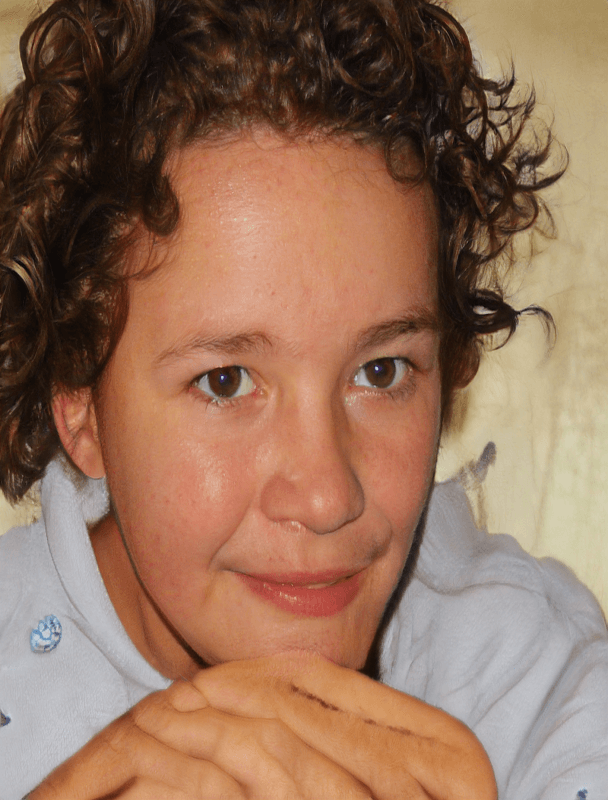}
     \includegraphics[width=\linewidth]{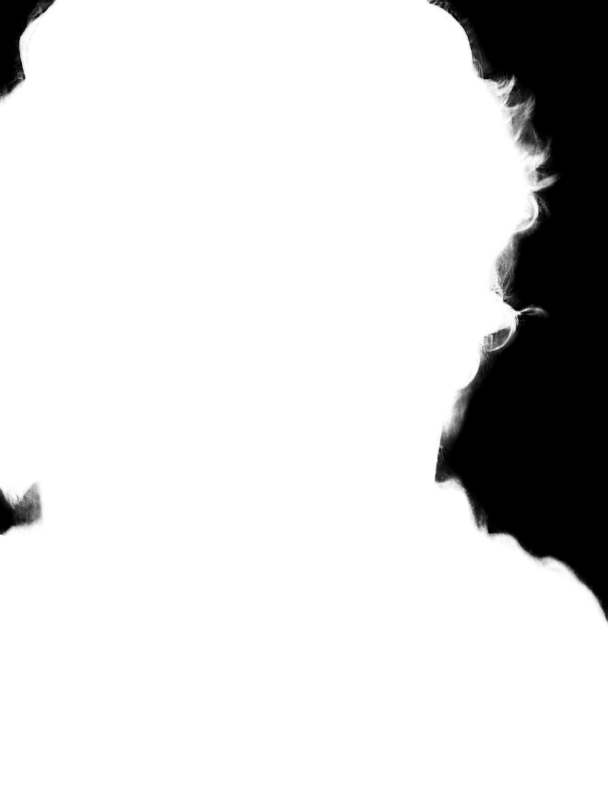}
     \end{minipage}
     }
   \hspace{-5mm}
   {
     \begin{minipage}{0.099\linewidth}
     \includegraphics[width=\linewidth]{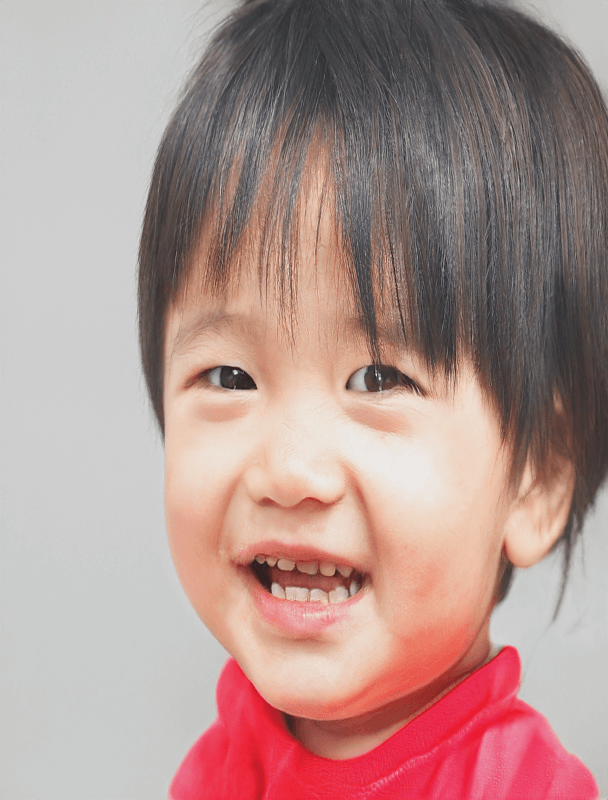}
     \includegraphics[width=\linewidth]{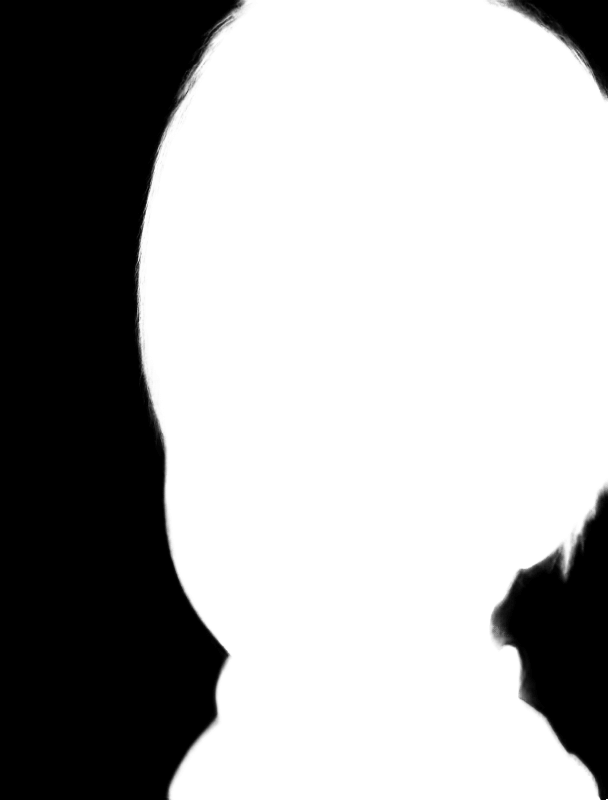}
     \end{minipage}
     }
   \hspace{-5mm}
   {
     \begin{minipage}{0.099\linewidth}
     \includegraphics[width=\linewidth]{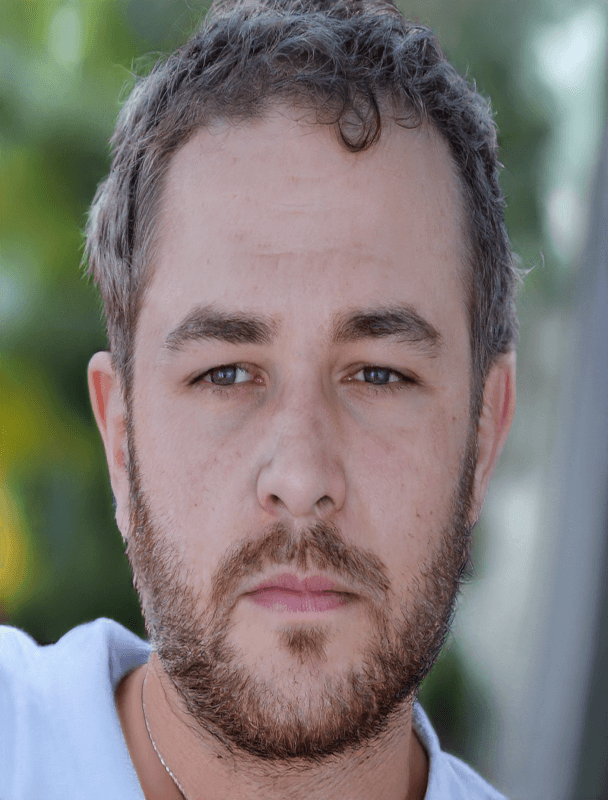}
     \includegraphics[width=\linewidth]{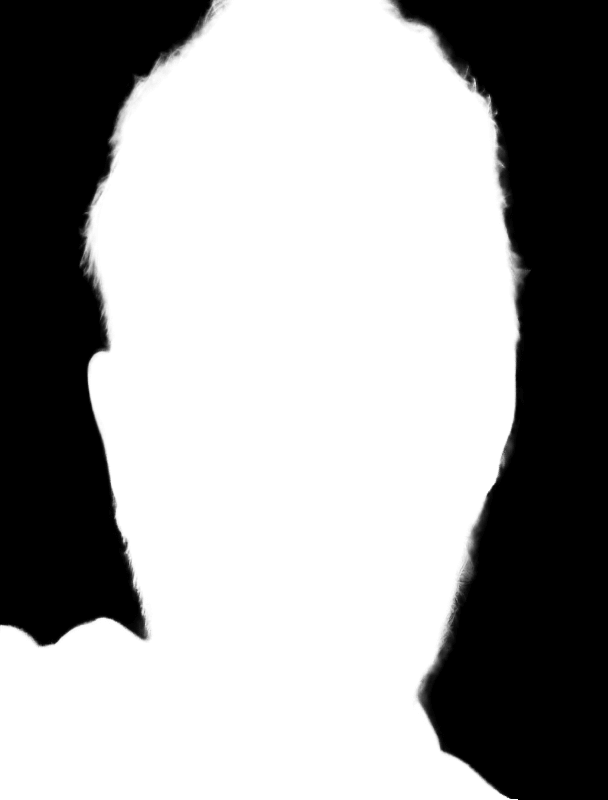}
     \end{minipage}
     }
   \hspace{-5mm}
   {
     \begin{minipage}{0.099\linewidth}
     \includegraphics[width=\linewidth]{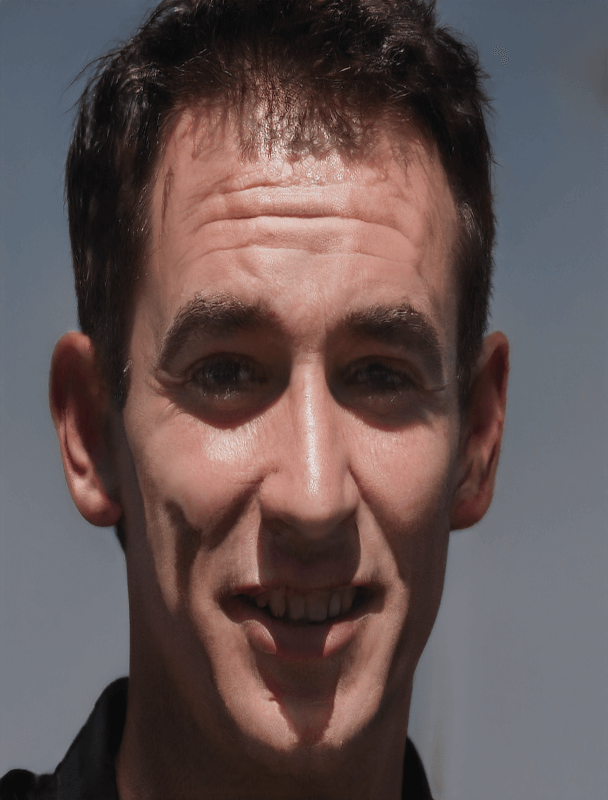}
     \includegraphics[width=\linewidth]{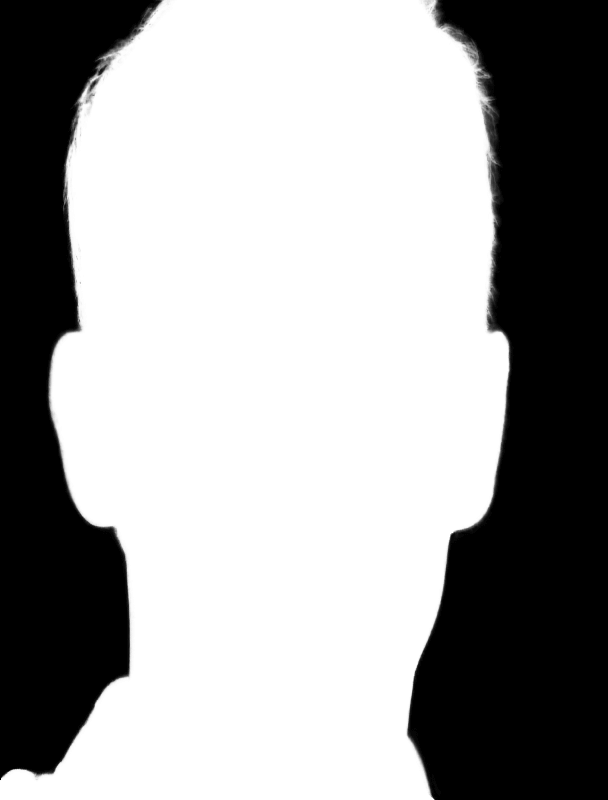}
     \end{minipage}
     }
   \hspace{-5mm}
   {
     \begin{minipage}{0.099\linewidth}
     \includegraphics[width=\linewidth]{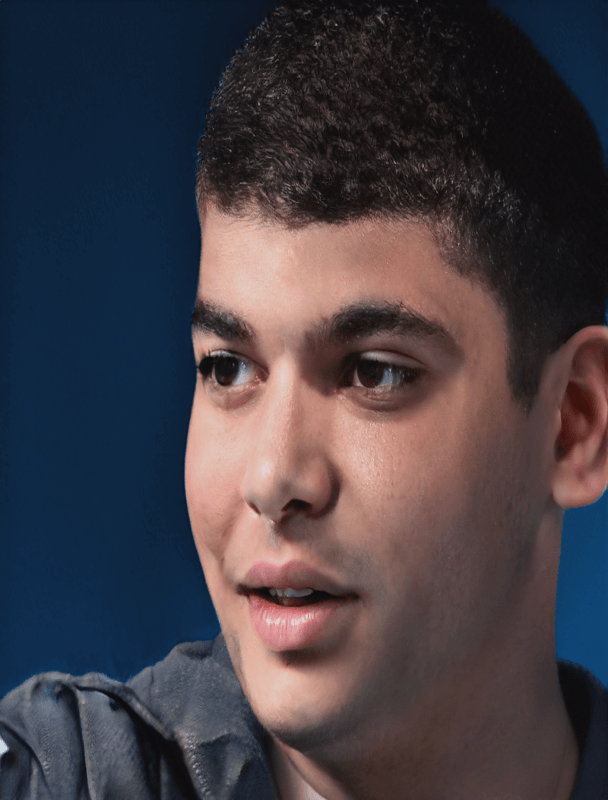}
     \includegraphics[width=\linewidth]{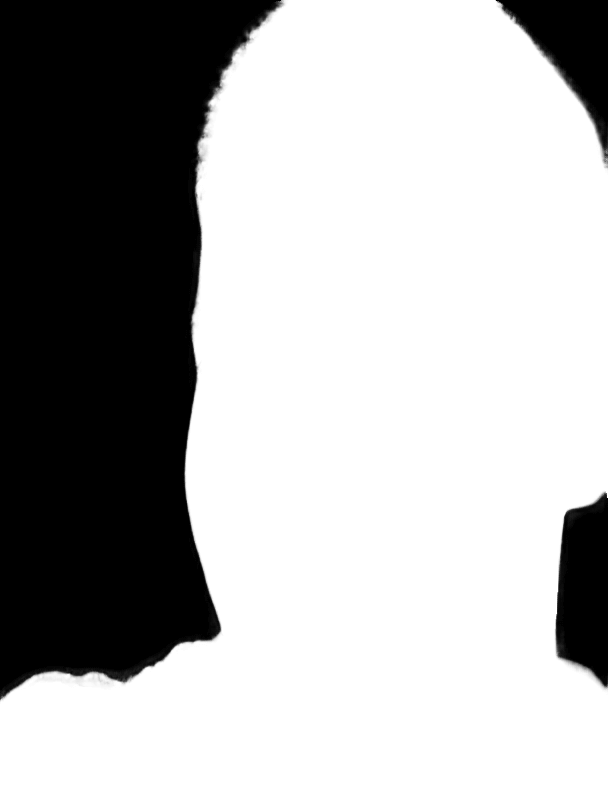}
     \end{minipage}
     }
   \hspace{-5mm}
   {
     \begin{minipage}{0.099\linewidth}
     \includegraphics[width=\linewidth]{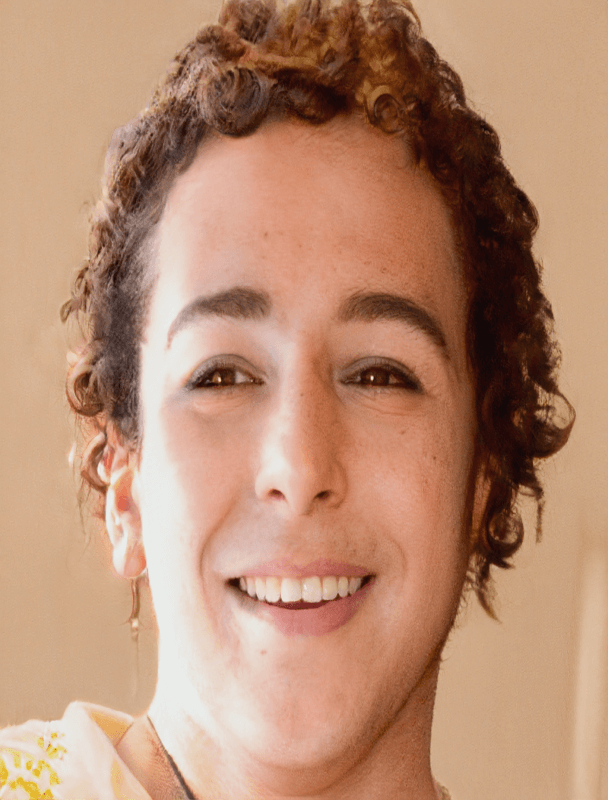}
     \includegraphics[width=\linewidth]{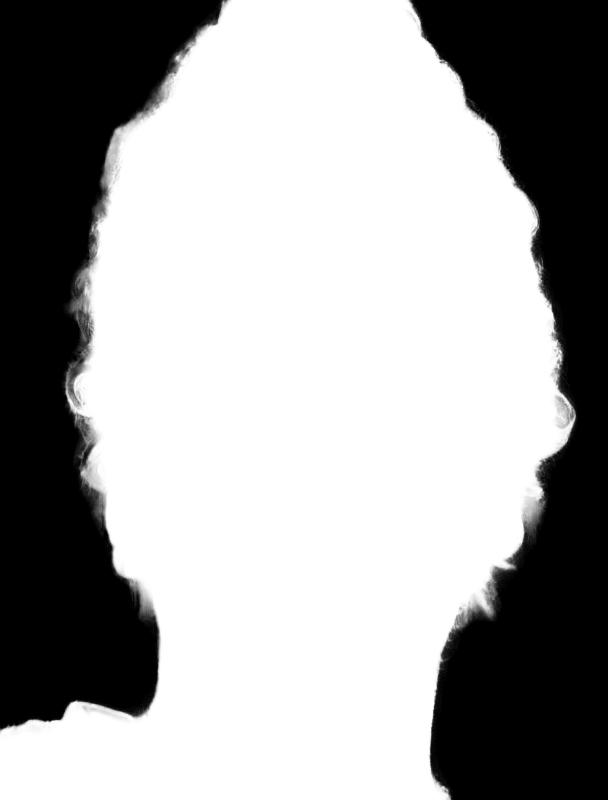}
     \end{minipage}
     }
   \hspace{-5mm}
   {
     \begin{minipage}{0.099\linewidth}
     \includegraphics[width=\linewidth]{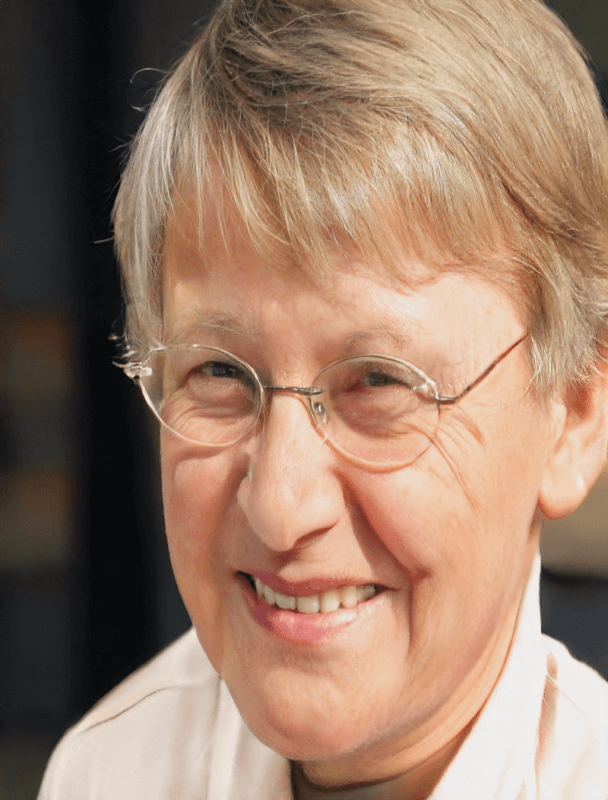}
     \includegraphics[width=\linewidth]{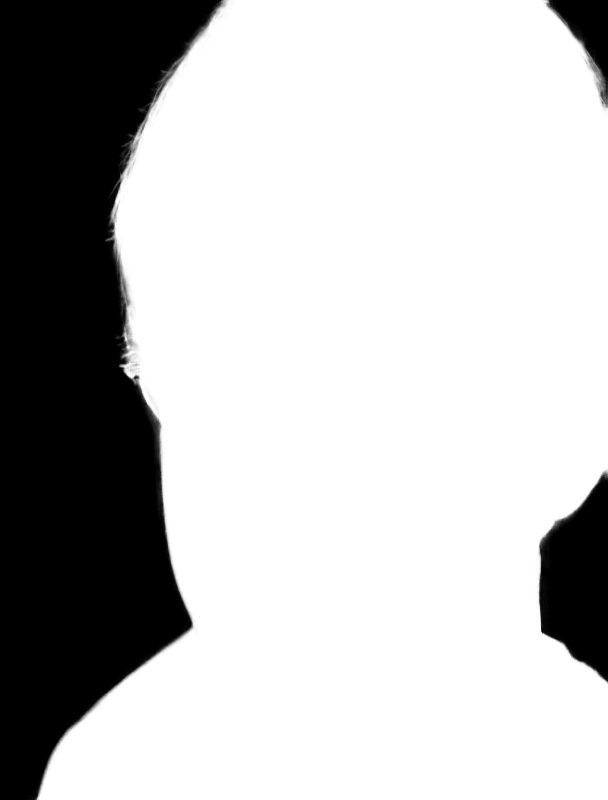}
     \end{minipage}
     }
   \hspace{-5mm}
   {
     \begin{minipage}{0.099\linewidth}
     \includegraphics[width=\linewidth]{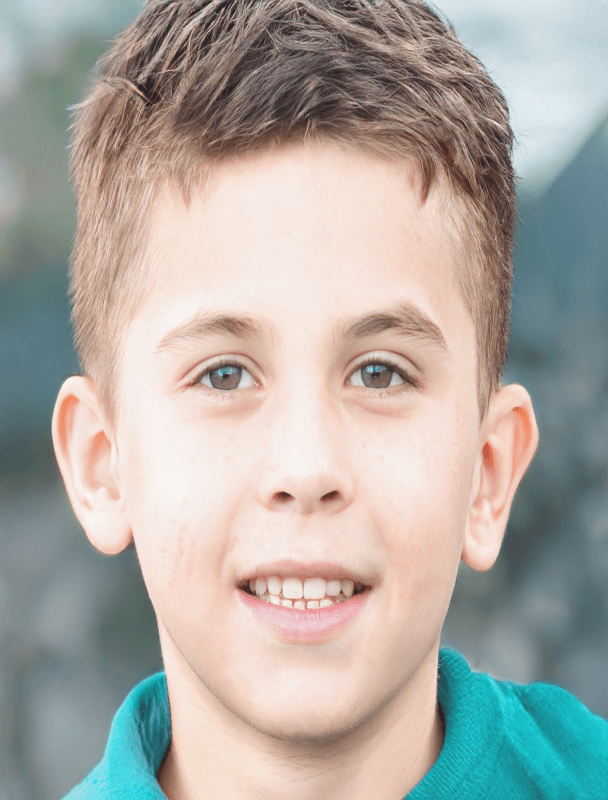}
     \includegraphics[width=\linewidth]{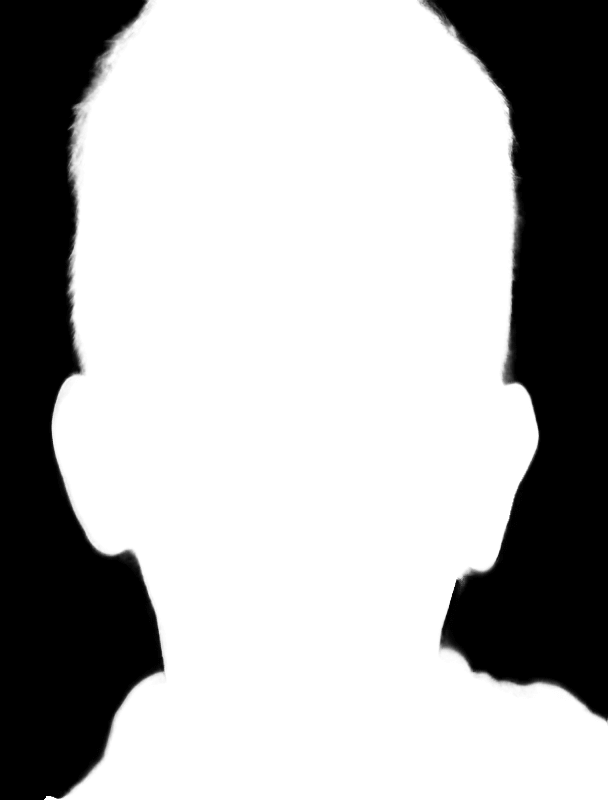}
     \end{minipage}
   }
   \hspace{-5mm}
   {
     \begin{minipage}{0.099\linewidth}
     \includegraphics[width=\linewidth]{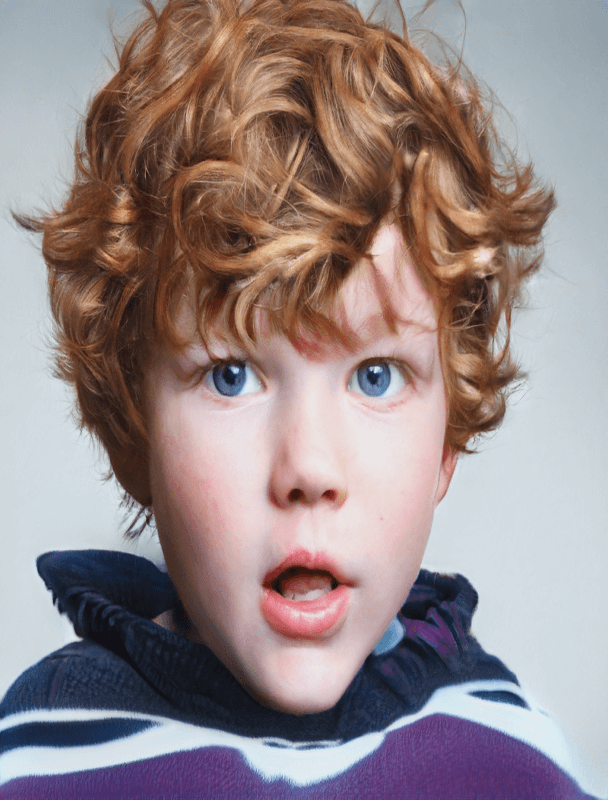}
     \includegraphics[width=\linewidth]{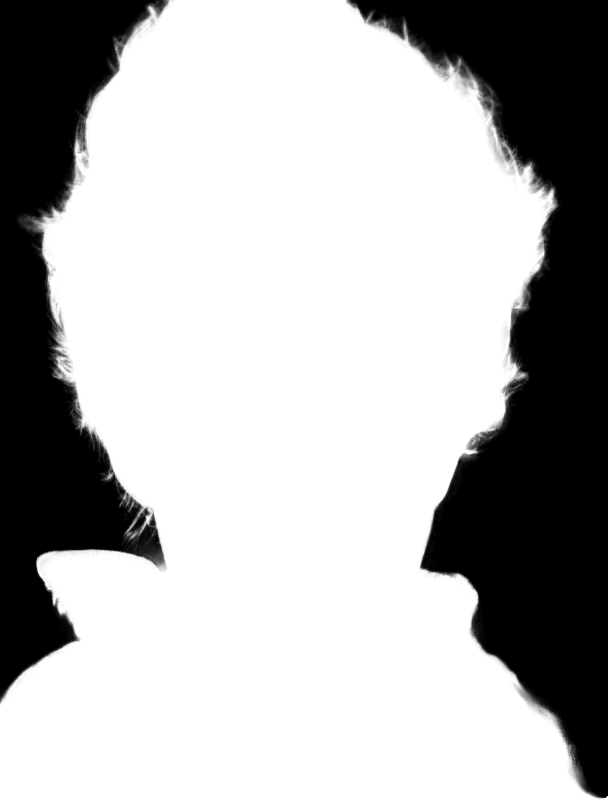}
     \end{minipage}
   }
   \hspace{-5mm}
   {
     \begin{minipage}{0.099\linewidth}
     \includegraphics[width=\linewidth]{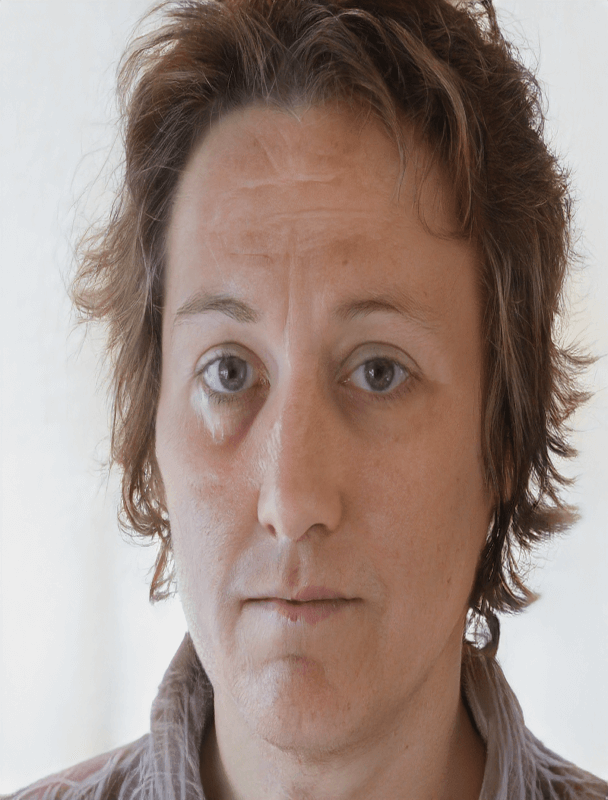}
     \includegraphics[width=\linewidth]{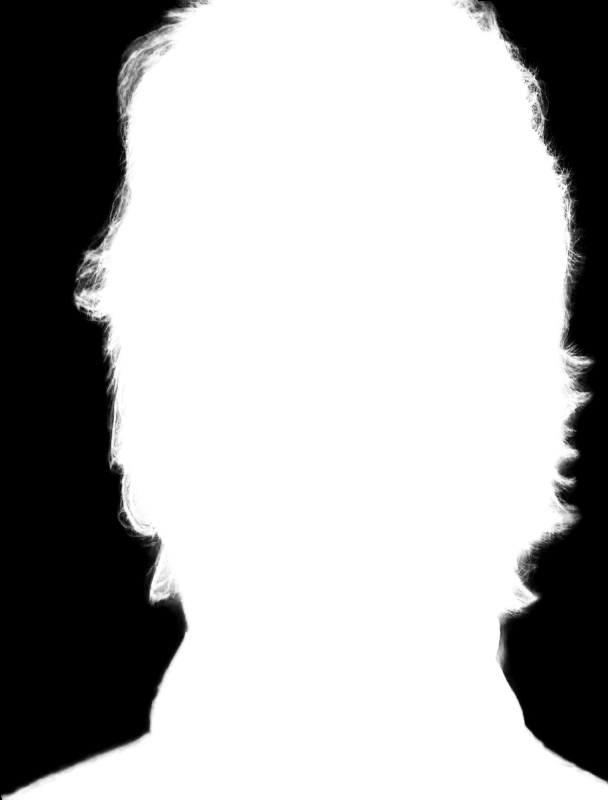}
     \end{minipage}
   }\vspace{2mm}
\caption{Examples of the generated portraits and their corresponding alpha mattes. The synthesized portraits are diverse and no discontinuous region appears in the pseudo GT data.}
\label{fig:generation_exp}
\end{figure*} 

\subsubsection{Qualitative Results}\label{sec:result}

Fig.~\ref{fig:result1} shows the qualitative results of portrait enhancement. As the proposed entropy minimization loss is directly applied to trimap prediction, we can see from the second column of each group, the enhanced trimaps fixed discontinuous regions, leading to better alpha mattes and composition results (see blue boxes). On the other hand, as can be seen in the perturbation images, the enhanced portraits add subtle details in almost all the areas. This demonstrates our observation that minor changes can yield a significantly different alpha matte. An interesting observation is that noise patterns are mostly attached to high-frequency components of the images, \eg, edges. This automatically learned operation surprisingly agrees with human perception, as humans are not particularly sensitive to high-frequency content in images. We further observe that relatively intense noise patterns may add to background regions, as shown in the top-left example in Fig.~\ref{fig:result1}. This interesting observation shows that the network is ``taught'' to avoid adding obvious noise to the foreground. We believe this attributes to the proposed compositional adversarial loss, as it ensures the high quality of composite images. These meaningful patterns are the reason of correcting those discontinuous holes in the alpha mattes. 


\subsection{Evaluation on Portrait Generation}

\begin{table*}[h]
   \caption{Comparisons with various variants using IndexNet as the reference matting model, under the application of portrait enhancement. ``Baseline'' represents the performance on the original portraits and ``($w^*,n^*$)'' represents the performance of our complete model. The lower the better for all metrics and the best results are marked in \textbf{bold}.}
       \begin{center}
       \setlength{\tabcolsep}{0.275cm}{
       \begin{tabular}{c|c|c|c|c|c|c|c|c}
           \hline
           \diagbox{Metrics}{Variants}        &Baseline    &($w^*,n$)           &($w,n^*$)      &w/o~$\mathcal{L}_{em}$    &w/o~$\mathcal{L}_{ca}$   &w/o~$\mathcal{L}_{pc}$  &w/o~$\mathcal{L}_{pp}$  &($w^*,n^*$)  \\

           \hline
           Grad$\downarrow$                               &7.404       &6.751           &6.688              &7.015                           &6.368               &6.613       &6.232      &\textbf{6.168}         \\
           \rowcolor{gray!20}Conn$\downarrow$             &11.345      &10.657          &11.013             &11.082                         &10.528              &10.688      &10.378     &\textbf{10.302}        \\
           MSE$\downarrow$                                &0.016       &0.015           &0.015              & 0.015                         &0.014               &0.014       &0.014        &\textbf{0.014}        \\
           \rowcolor{gray!20}SAD$\downarrow$              &11.824      &11.154          &11.477             & 11.568                        &10.994              &10.893      &10.956    &\textbf{10.799}       \\
           \hline

       \end{tabular}
       }
      \end{center}
      \label{table:2}
      \vspace{-3mm}
\end{table*}

\begin{table*}[t]
   \caption{\rev{Comparisons with various variants with different parameter settings and foreground estimation methods. The lower the better for all metrics and the best results are marked in \textbf{bold}.}}
     \vspace{-0.3cm}
       \begin{center}
       \setlength{\tabcolsep}{0.3cm}{
       \begin{tabular}{c|c|c|c|c|c|c}
           \hline
           \diagbox{Metrics}{Variants}   &\rev{Hyper-param. \#1} &\rev{Hyper-param. \#2} &\rev{FG w/~\cite{germer2021fast}}  &\rev{FG w/~\cite{forte2021approximate}} &\rev{FG w/ GT $\alpha$} &($w^*,n^*$)  \\

           \hline
           Grad$\downarrow$       &\rev{\textbf{6.165}}       &\rev{6.166}    &\rev{6.165}  &\rev{\textbf{6.164}} &\rev{\textbf{6.165}}     &{6.168}         \\
           \rowcolor{gray!20}Conn$\downarrow$        &\rev{10.293}      &\rev{10.312} &\rev{10.292}      &\rev{10.303}  &\rev{\textbf{10.296}}               &{10.302}        \\
           MSE$\downarrow$                &\rev{0.014}       &\rev{0.014} &\rev{0.014}       &\rev{0.015}  &\rev{0.014}       &\textbf{0.014}        \\
           \rowcolor{gray!20}SAD$\downarrow$         &\rev{10.899}      &\rev{10.896}  &\rev{10.895}      &\rev{10.901}    &\rev{10.893}    &\textbf{10.799}       \\
           \hline

       \end{tabular}
       }
      \end{center}
      \label{table:3}
      \vspace{-3mm}
\end{table*}


Here we evaluate the performance of our second application, portrait generation. We use four state-of-the-art methods to govern the optimization, leading to four different sets of synthesized data. \rev{There might be a problem that the generated portrait-matte pairs overfit to a reference matting model and gain less improvement. Hence we mix the synthesized samples that come from different matting models for fine-tuning a matting model. This strategy is to increase the diversity of the data, as different models have different emphasizes on easy-to-extract images. Besides, it also avoids the overfitting problem, as image pairs generated by different matting models are blind to the target one.} Each set contains 10,000 synthesized images. These four sets of data, combined with the original training data of~\cite{shen2016deep}, are used to fine-tune four matting models.

Tab.~\ref{table:g} shows the quantitative evaluation on portrait augmentation with two datasets. We can see that all the state-of-the-art models improve their performances to a certain extent. This implies that the synthesized data is of high-confidence GT, effective for training matting models. Compared with portrait enhancement, {fine-tuning} with synthesized data receives less improvement. This is because although the network is {fine-tuned} with enhanced portraits, applying enhancement in the testing phrase optimizes an image-specific latent code that leads to the best matting performance.

We also show some examples of the generated portraits and the corresponding alpha mattes in Fig.~\ref{fig:generation_exp}. As can be seen, the generated portraits are diverse enough, with high-quality alpha mattes without discontinuous regions.

\begin{figure*}[!h]
  \centering
  \captionsetup[subfloat]{labelformat=empty,justification=centering}
  \subfloat[\scriptsize{Portrait}]{\label{fig:hard_a}
    \rotatebox[origin=c]{270}{Original \hspace{7mm} Enhanced \hspace{7mm} Original \hspace{7mm} Enhanced}
    \begin{minipage}{0.09\linewidth}
    \includegraphics[width=\linewidth]{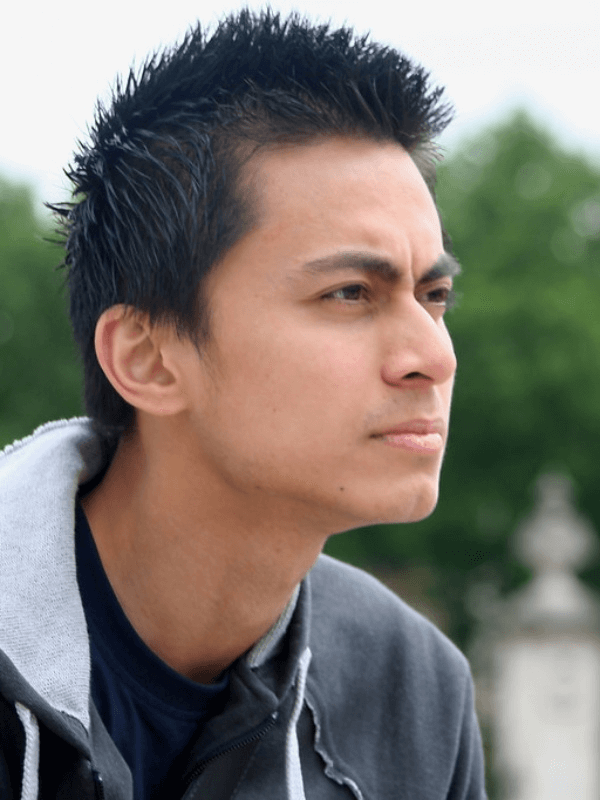}
    \includegraphics[width=\linewidth]{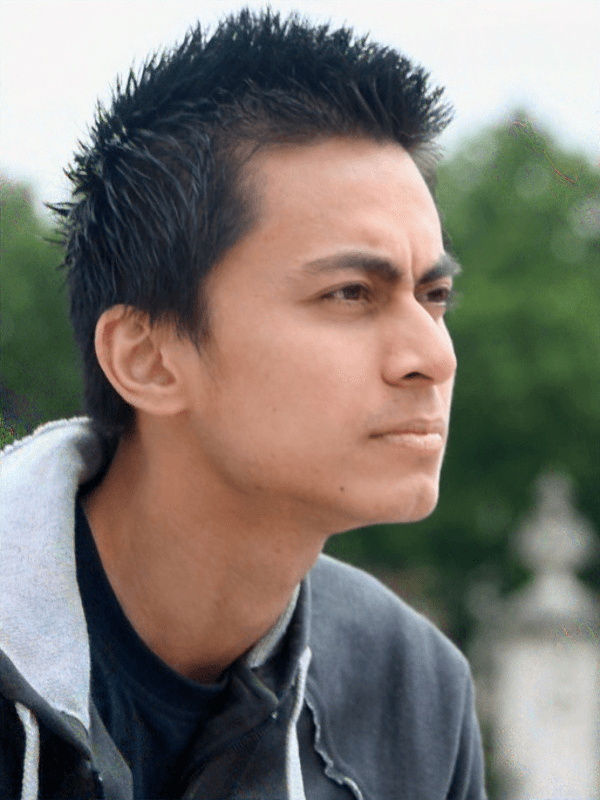}
    \includegraphics[width=\linewidth]{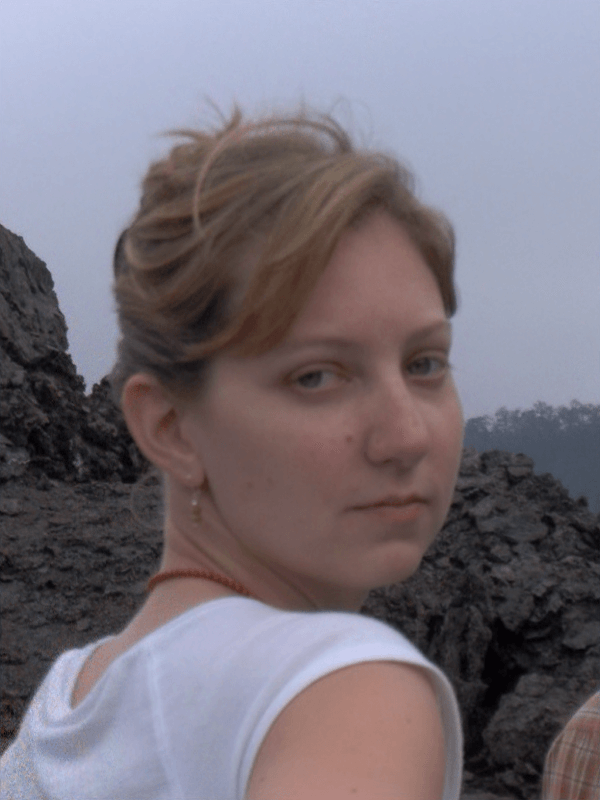}
    \includegraphics[width=\linewidth]{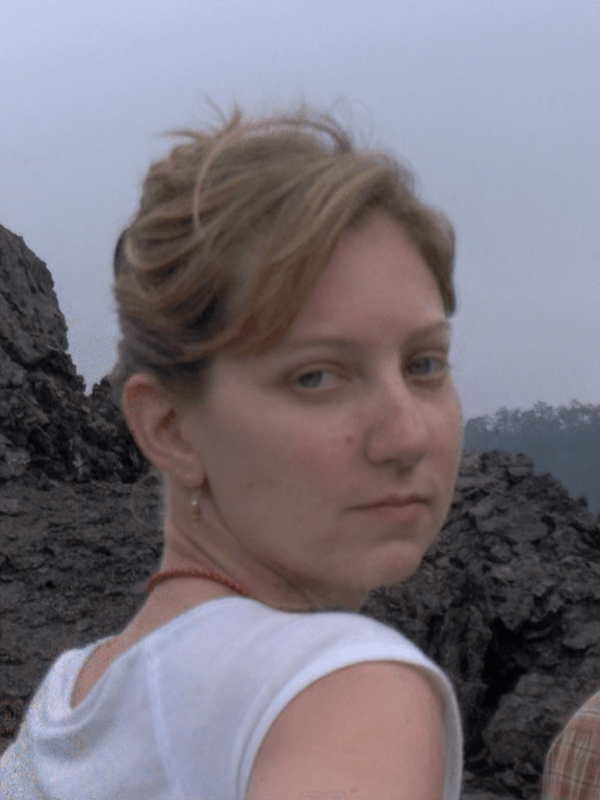}
    \end{minipage}
    }
  \hspace{-3.5mm}
  \subfloat[\scriptsize{Trimap}]{\label{fig:hard_b}
    \begin{minipage}{0.09\linewidth}
    \includegraphics[width=\linewidth]{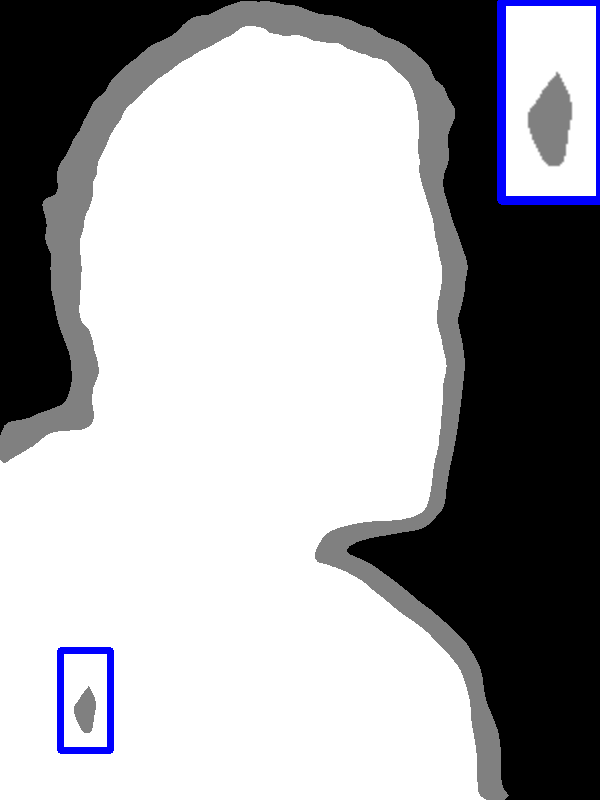}
    \includegraphics[width=\linewidth]{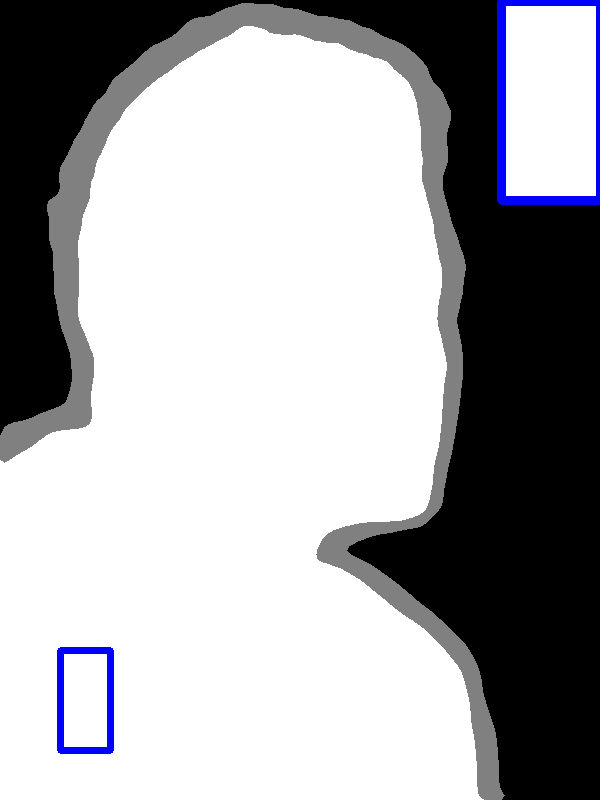}
    \includegraphics[width=\linewidth]{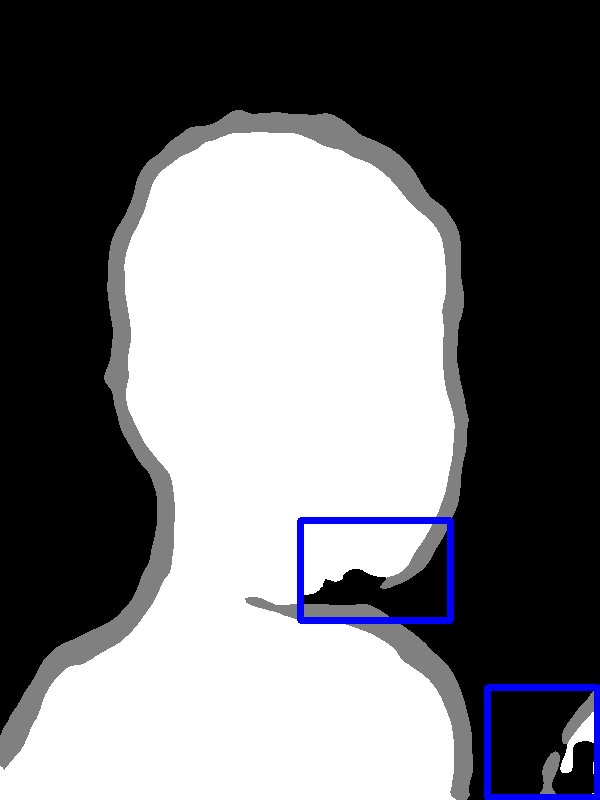}
    \includegraphics[width=\linewidth]{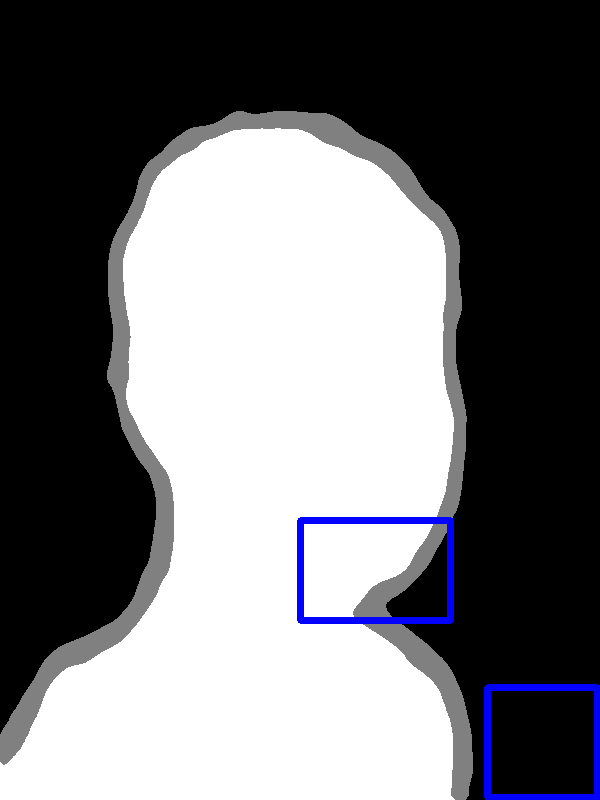}
    \end{minipage}
  }
  \hspace{-3.5mm}
  \subfloat[\scriptsize{Matte}]{\label{fig:hard_c}
    \begin{minipage}{0.09\linewidth}
    \includegraphics[width=\linewidth]{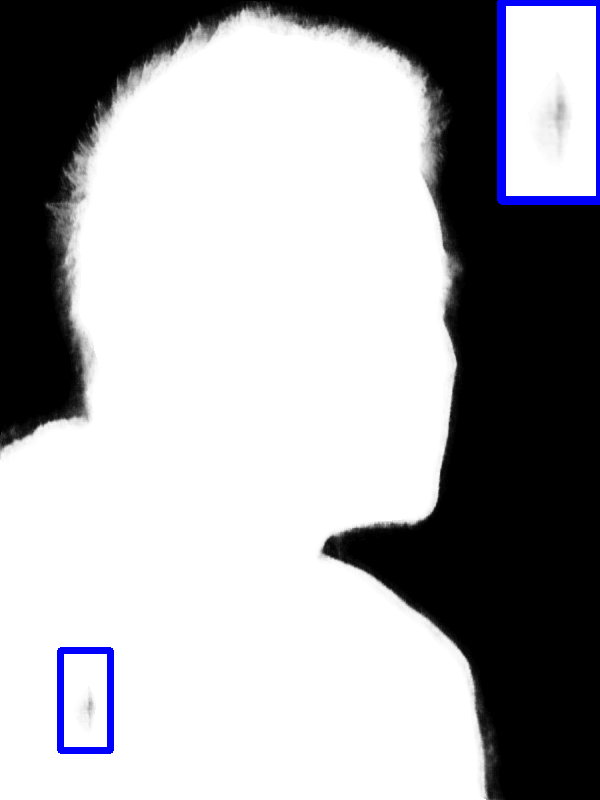}
    \includegraphics[width=\linewidth]{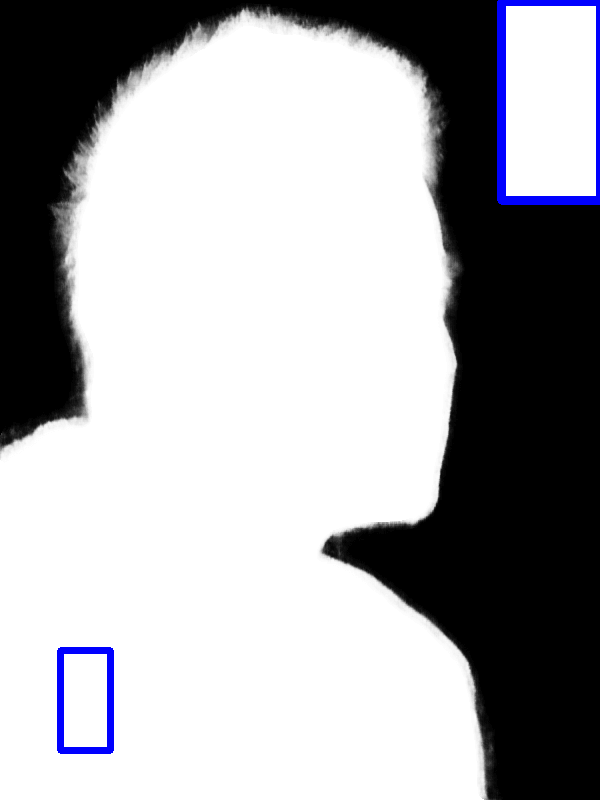}
    \includegraphics[width=\linewidth]{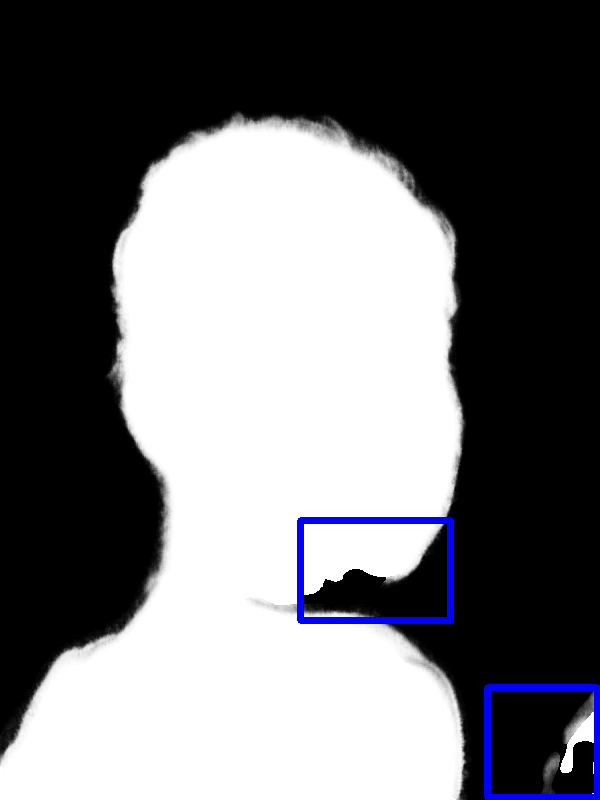}
    \includegraphics[width=\linewidth]{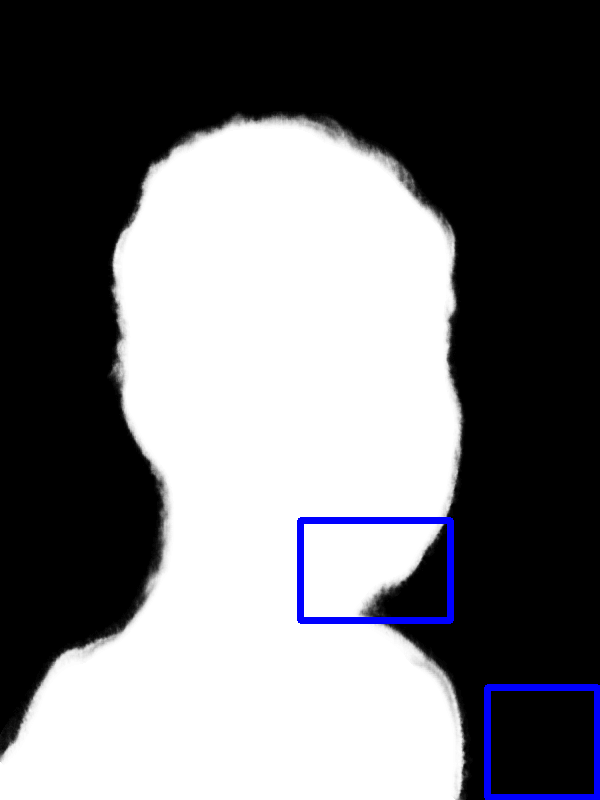}
    \end{minipage}
    }
  \hspace{-3.5mm}
  \subfloat[\scriptsize{Pert.\& GT}]{\label{fig:hard_d}
    \begin{minipage}{0.09\linewidth}
    \includegraphics[width=\linewidth]{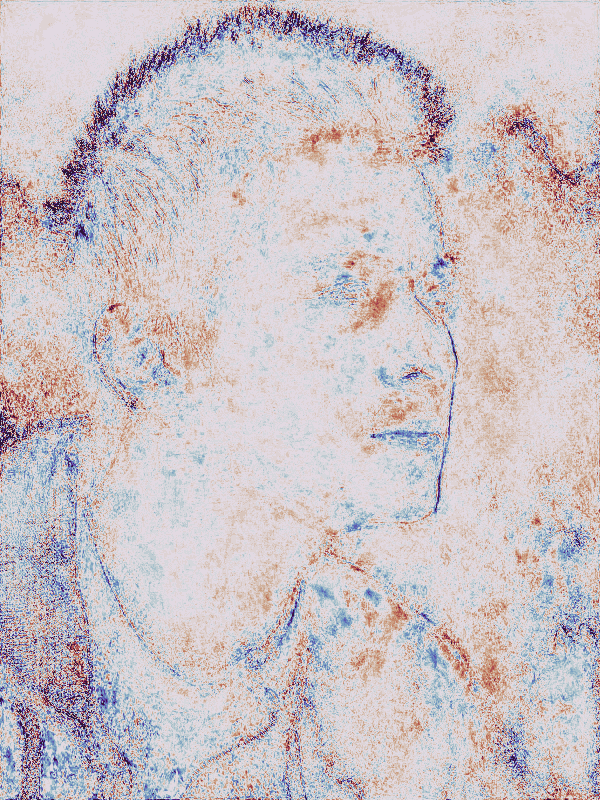}
    \includegraphics[width=\linewidth]{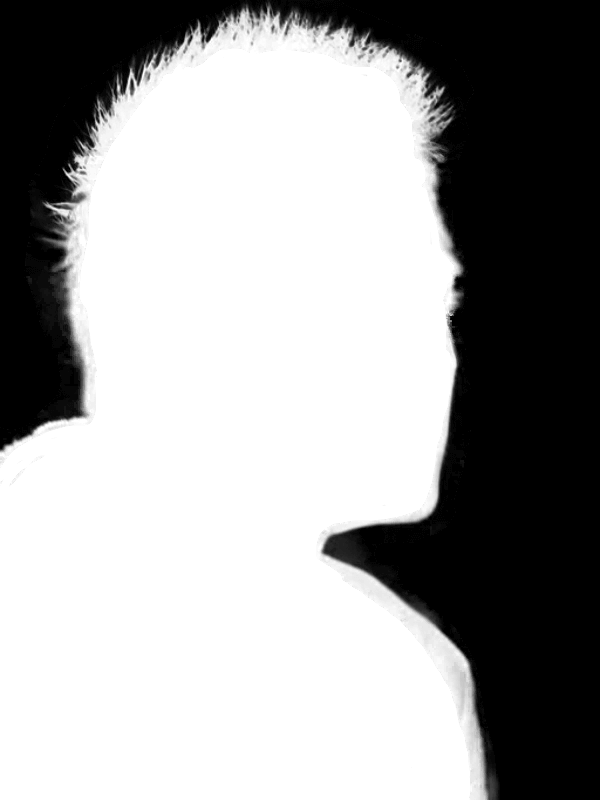}
    \includegraphics[width=\linewidth]{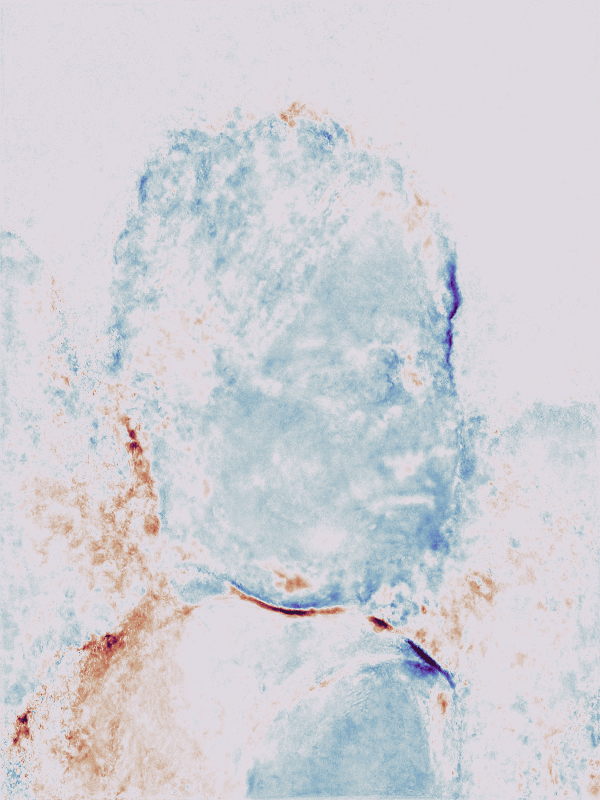}
    \includegraphics[width=\linewidth]{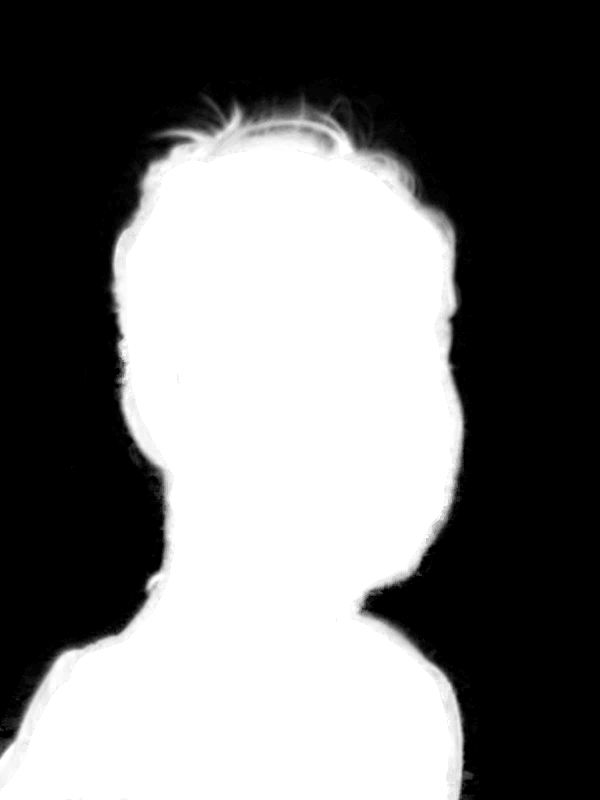}
    \end{minipage}
  }
  \hspace{-3.5mm}
  \subfloat[\scriptsize{Composition}]{\label{fig:hard_e}
    \begin{minipage}{0.09\linewidth}
    \includegraphics[width=\linewidth]{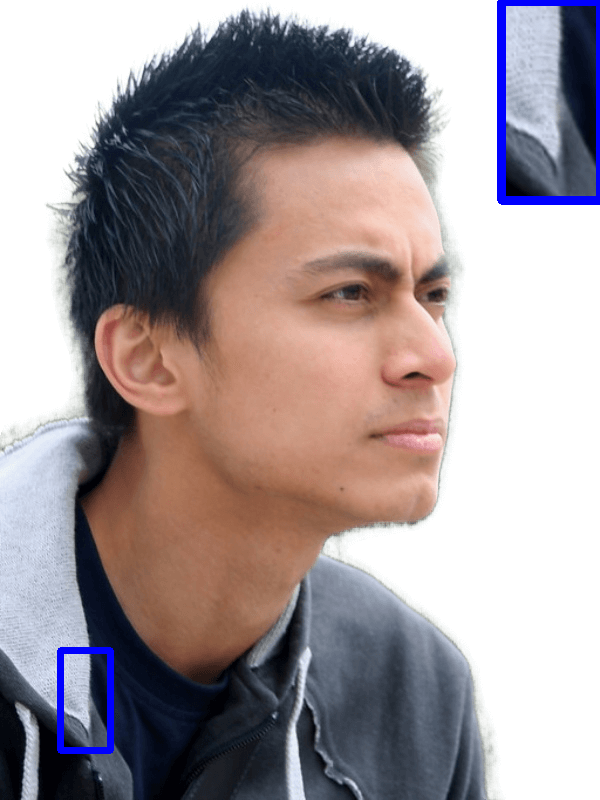}
    \includegraphics[width=\linewidth]{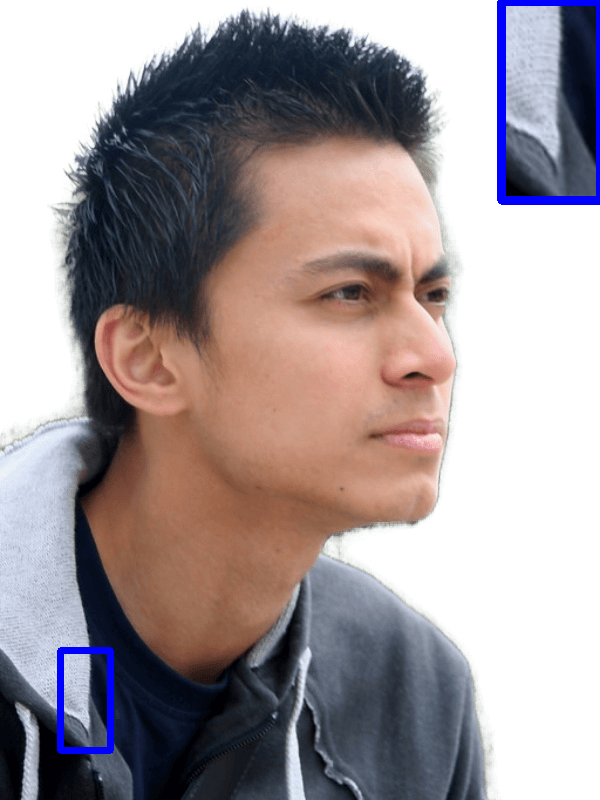}
    \includegraphics[width=\linewidth]{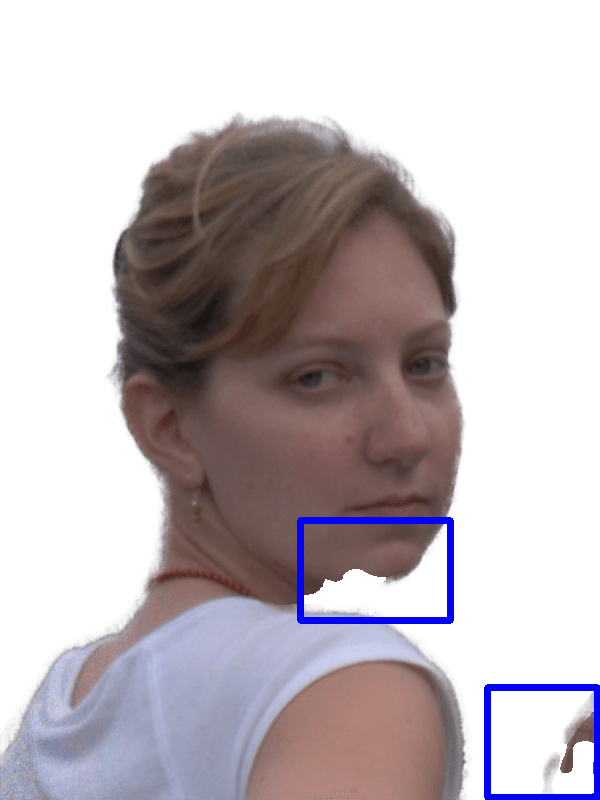}
    \includegraphics[width=\linewidth]{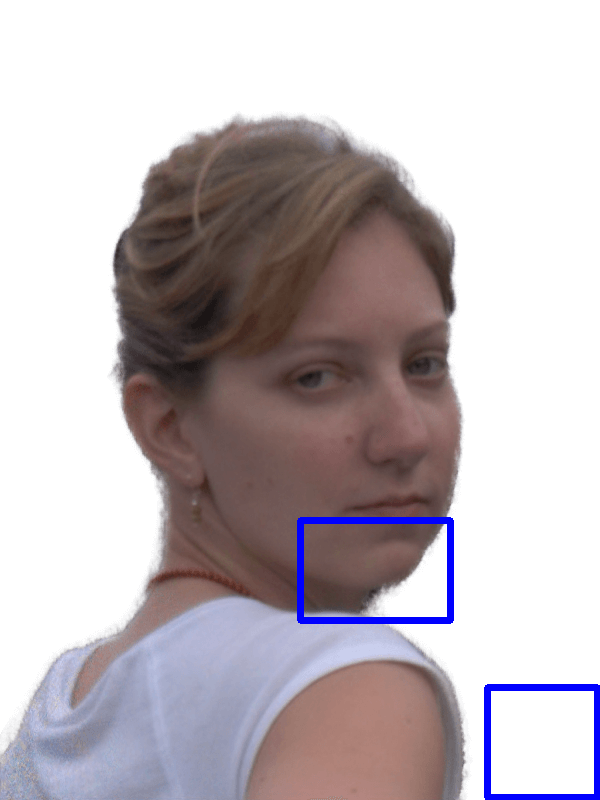}
    \end{minipage}
    }
  \hspace{1mm}
  \subfloat[\scriptsize{Portrait}]{\label{fig:hard_f}
    \begin{minipage}{0.09\linewidth}
      \includegraphics[width=\linewidth]{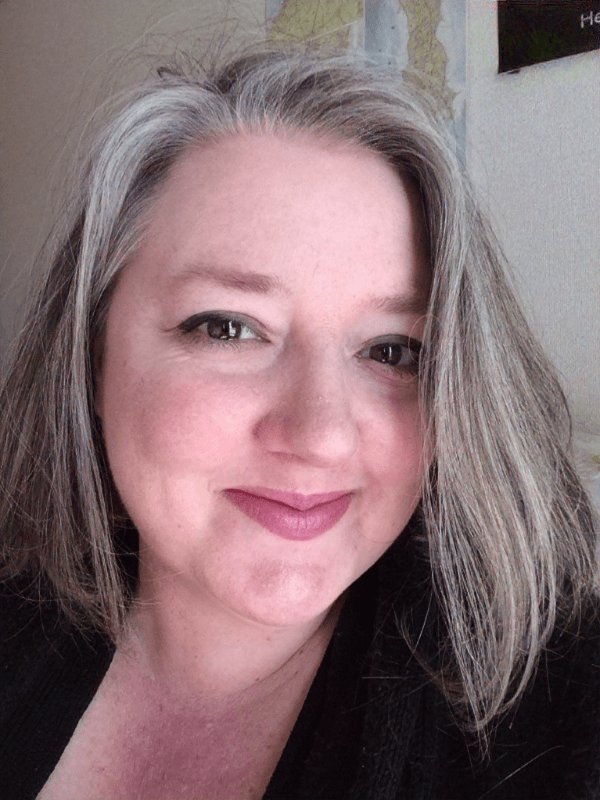}
      \includegraphics[width=\linewidth]{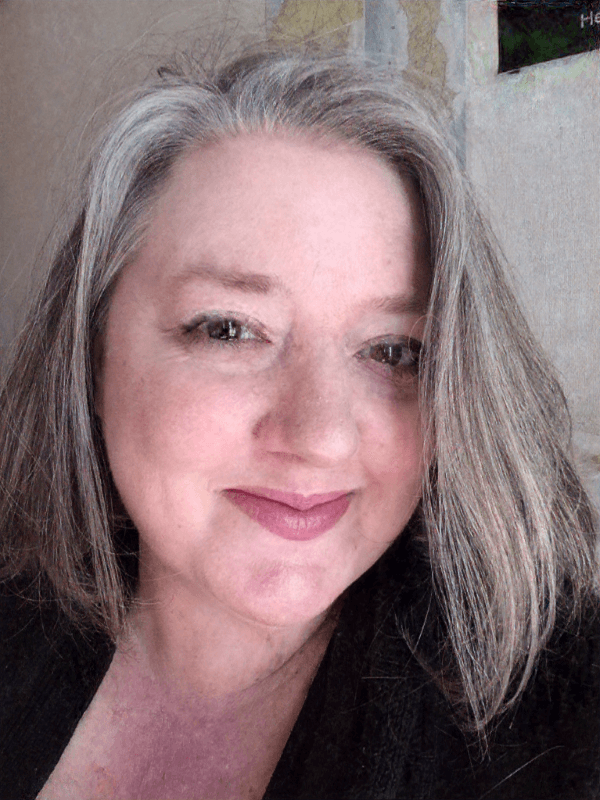}
      \includegraphics[width=\linewidth]{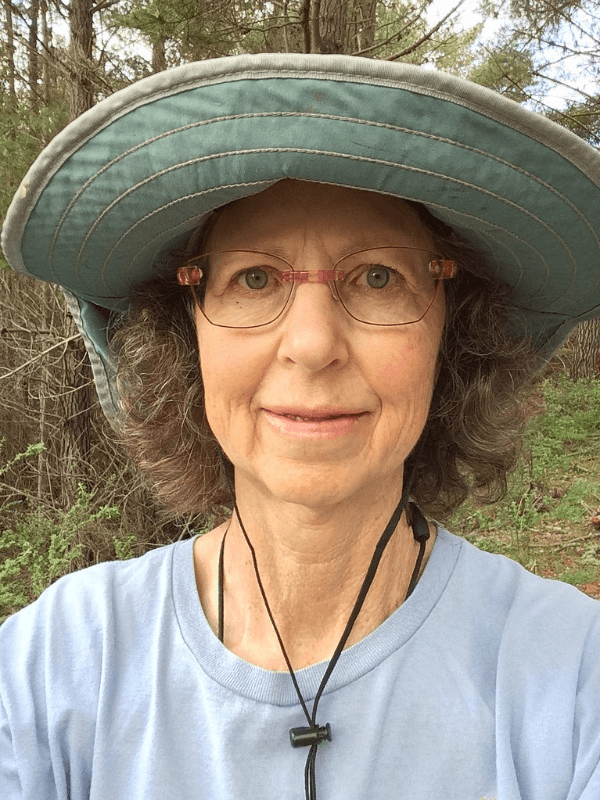}
      \includegraphics[width=\linewidth]{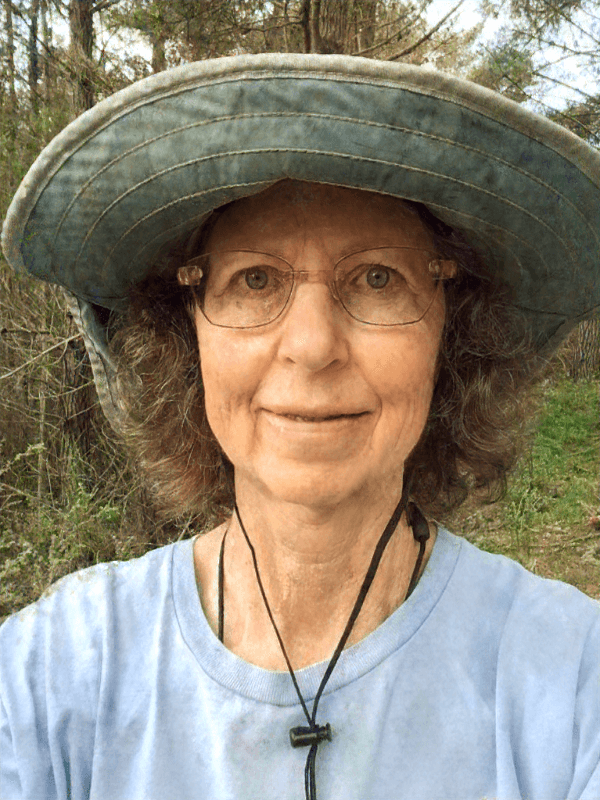}
    \end{minipage}
    }
  \hspace{-3.5mm}
  \subfloat[\scriptsize{Trimap}]{\label{fig:hard_g}
    \begin{minipage}{0.09\linewidth}
      \includegraphics[width=\linewidth]{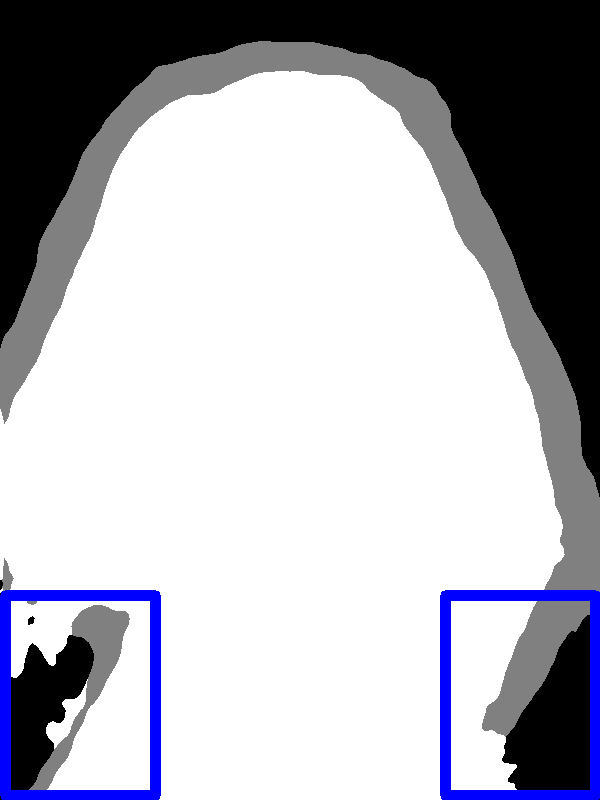}
      \includegraphics[width=\linewidth]{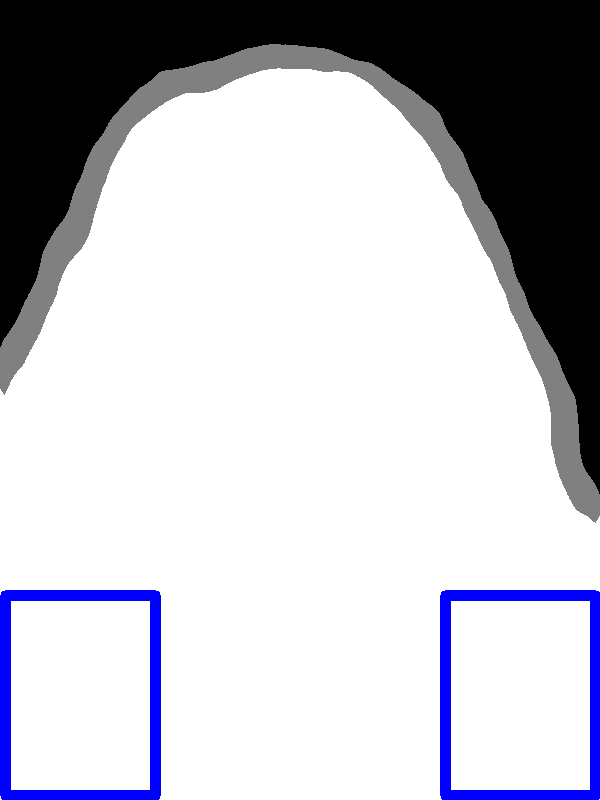}
      \includegraphics[width=\linewidth]{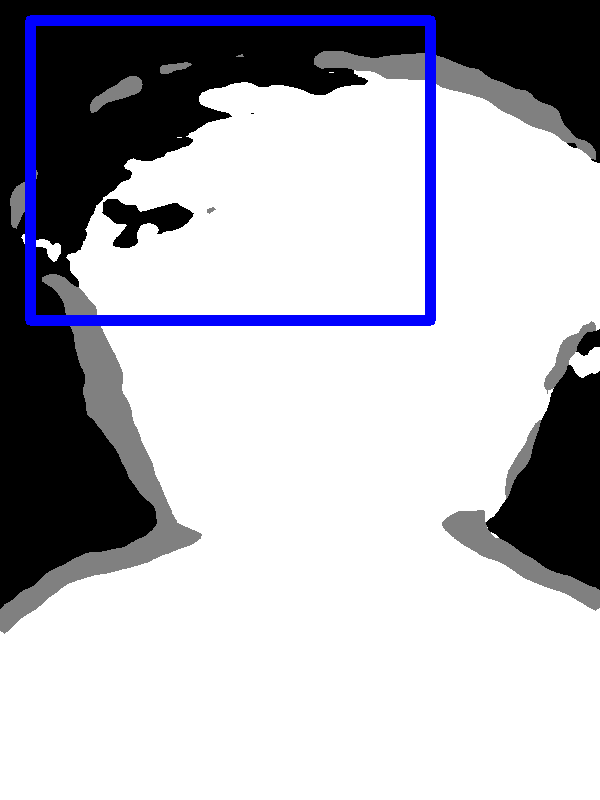}
      \includegraphics[width=\linewidth]{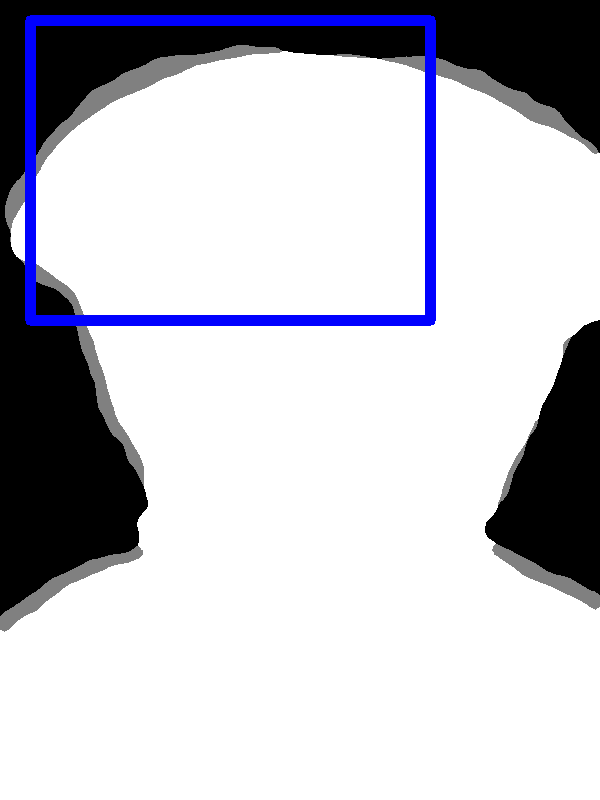}
    \end{minipage}
  }
  \hspace{-3.5mm}
  \subfloat[\scriptsize{Matte}]{\label{fig:hard_h}
    \begin{minipage}{0.09\linewidth}
      \includegraphics[width=\linewidth]{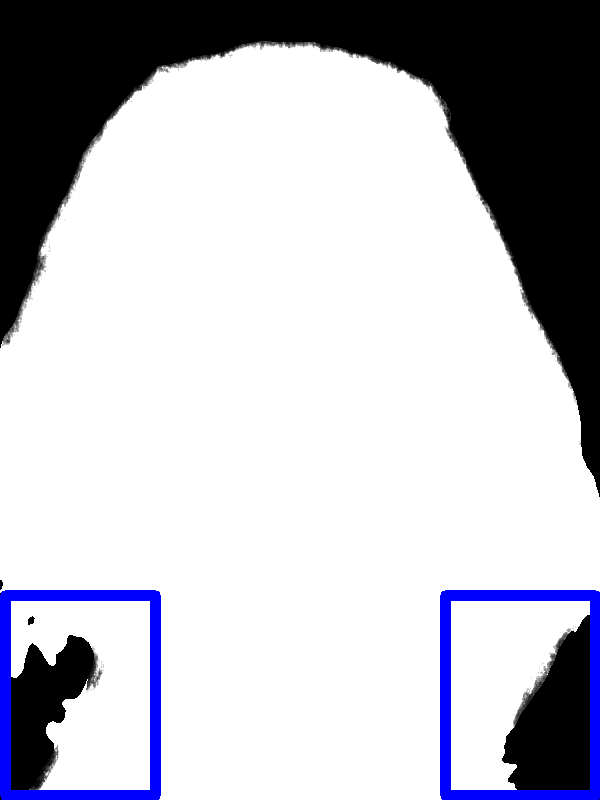}
      \includegraphics[width=\linewidth]{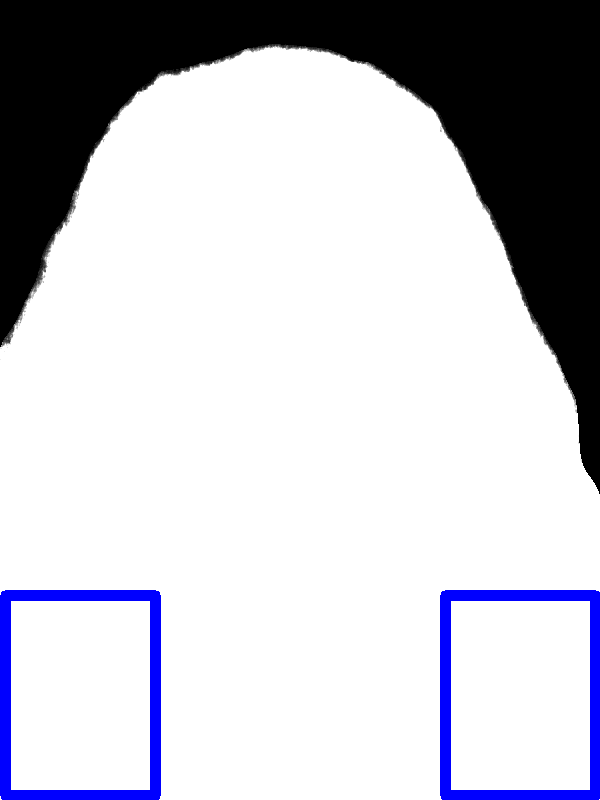}
      \includegraphics[width=\linewidth]{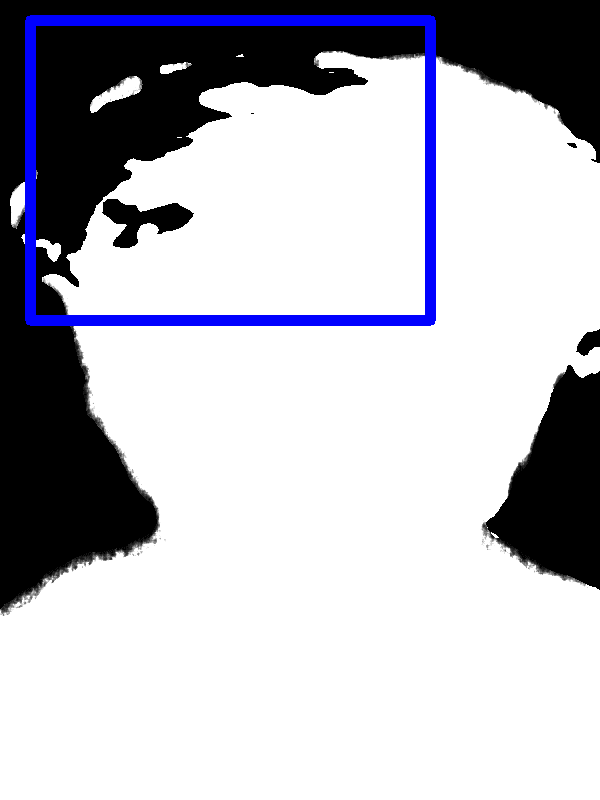}
      \includegraphics[width=\linewidth]{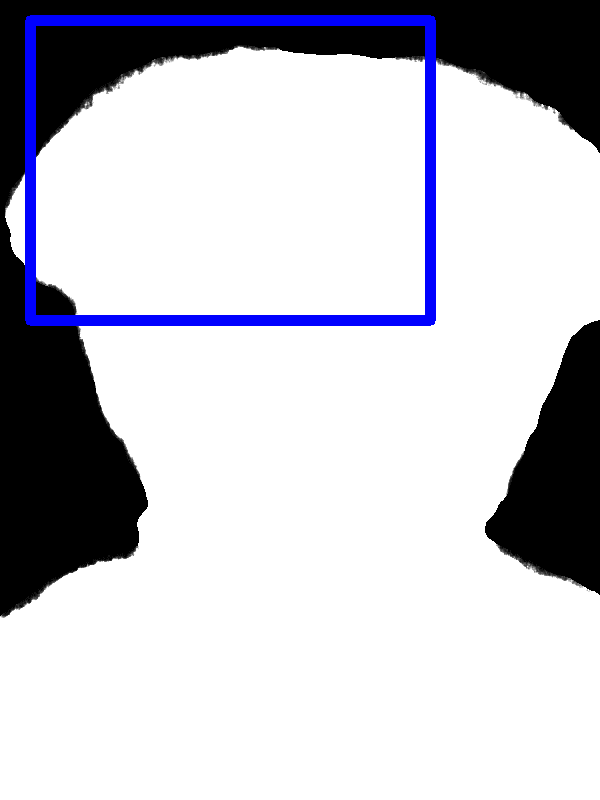}
    \end{minipage}
    }
  \hspace{-3.5mm}
  \subfloat[\scriptsize{Pert.\& GT}]{\label{fig:hard_i}
    \begin{minipage}{0.09\linewidth}
      \includegraphics[width=\linewidth]{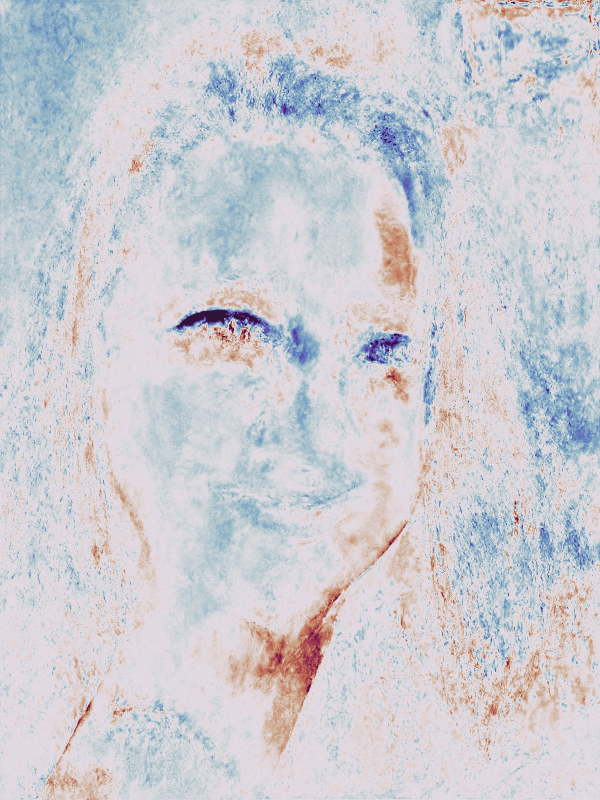}
      \includegraphics[width=\linewidth]{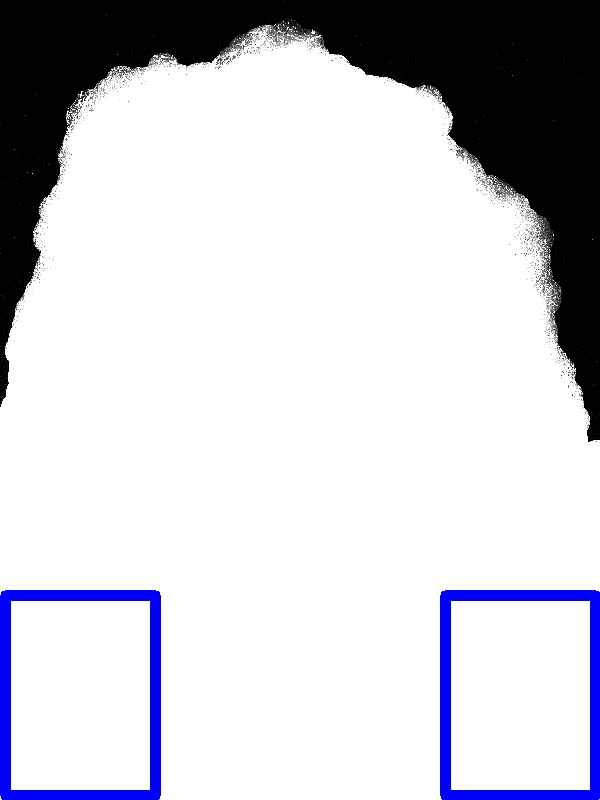}
      \includegraphics[width=\linewidth]{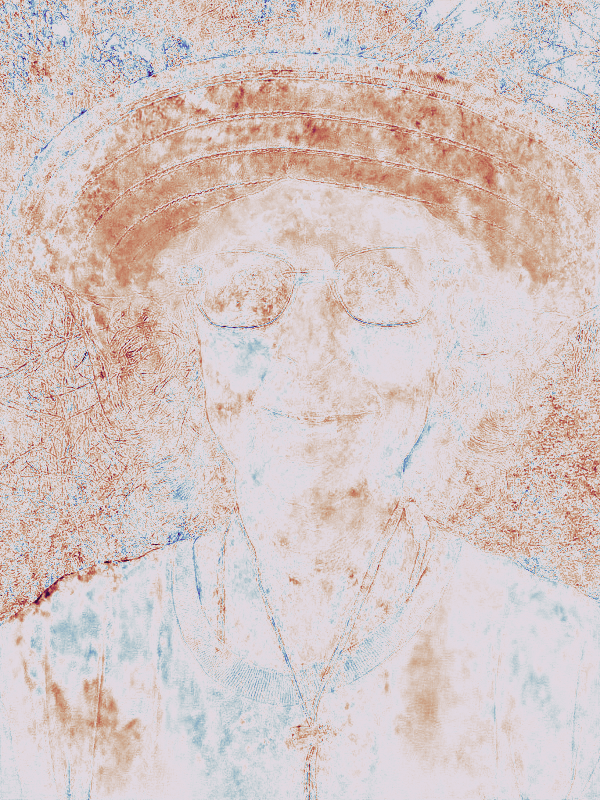}
      \includegraphics[width=\linewidth]{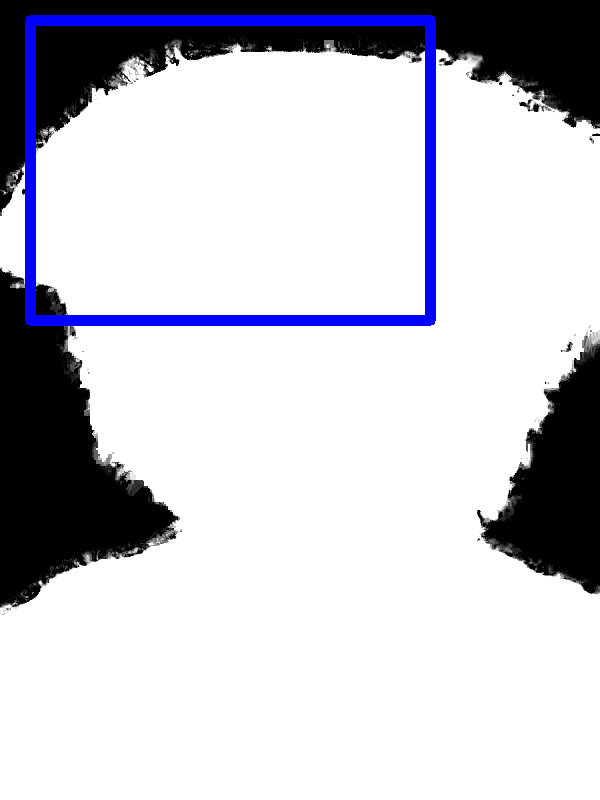}
    \end{minipage}
  }
  \hspace{-3.5mm}
  \subfloat[\scriptsize{Composition}]{\label{fig:hard_j}
    \begin{minipage}{0.09\linewidth}
      \includegraphics[width=\linewidth]{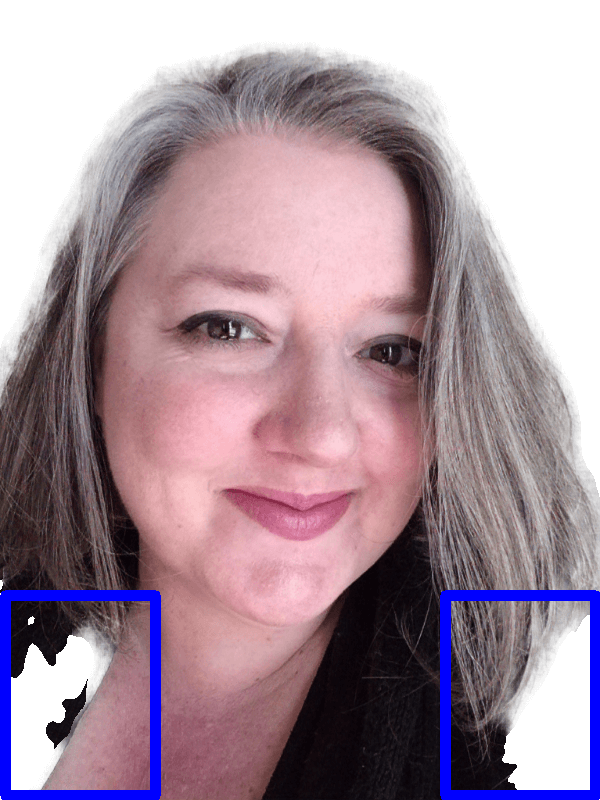}
      \includegraphics[width=\linewidth]{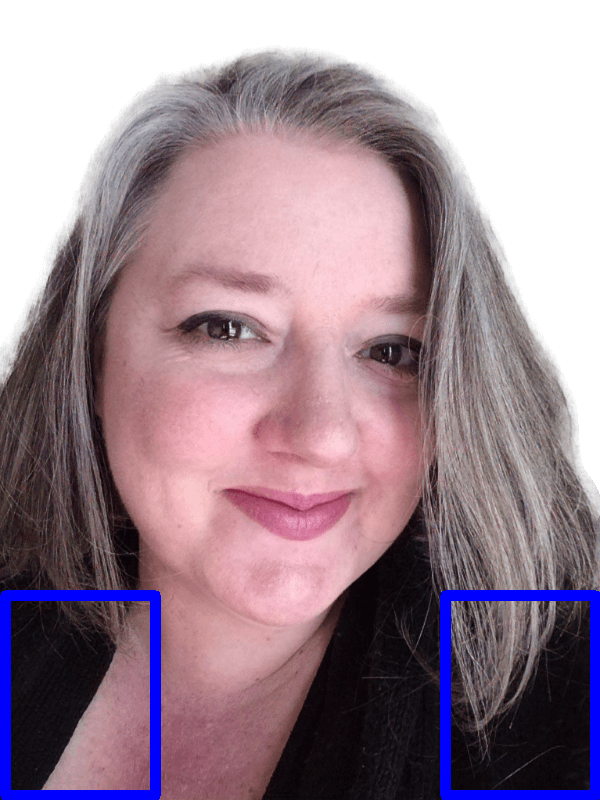}
      \includegraphics[width=\linewidth]{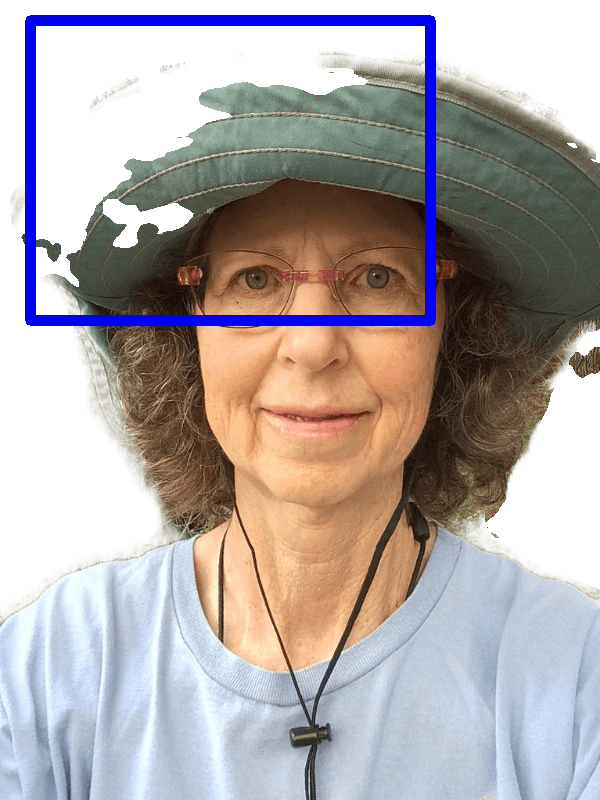}
      \includegraphics[width=\linewidth]{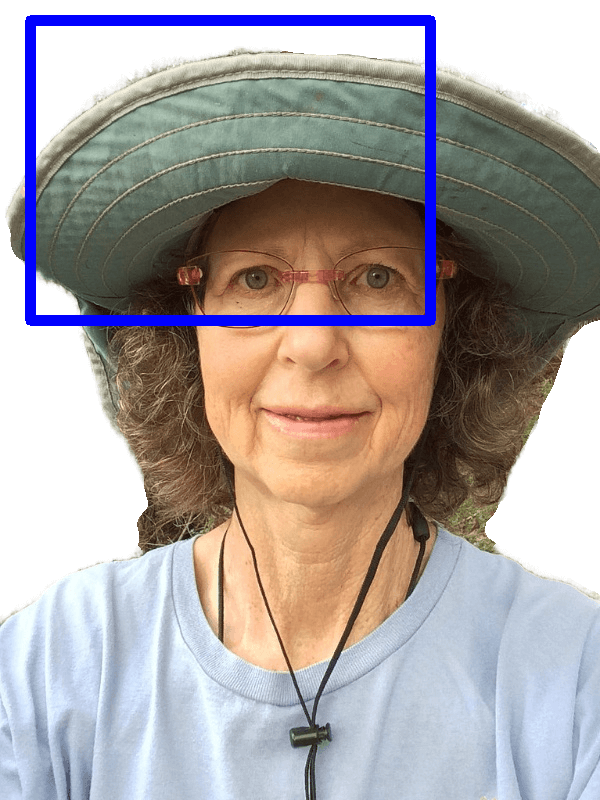}
    \end{minipage}
    }
  \hspace{-3.5mm}
  \caption{\rev{Our method can handle challenging cases like profiles (left part). In addition, we can also enhance the portraits with low-quality initial trimaps or alpha mattes (right part).}}
  \label{fig:result3}\vspace{-3mm}
\end{figure*}

\subsection{Ablation Study}
\label{sec.as}
In this section, we analyze the efficacy of the optimization in different latent spaces and loss functions, under the application of portrait enhancement on Portrait Matting Dataset. We first compare with six variants: 1) optimizing the latent code~$w$ only~while keeping the noise code~$n$~fixed~($w^*,n$); 2) optimizing the noise code~$n$ only ($w,n^*$); 3) without the entropy minimization loss (w/o~$\mathcal{L}_{em}$); 4) without the compositional adversarial loss (w/o~$\mathcal{L}_{ca}$); 5) without the portrait consistency loss (w/o~$\mathcal{L}_{pc}$); 6) without the perceptual loss (w/o~$\mathcal{L}_{pp}$). For variants 1) and 2), we optimize them using all the four losses. Similarly, for variants 3), 4), 5) and 6), we optimize the latent codes $w$ and $n$ simultaneously. \rev{In addition, we also evaluate out method with various hyper-parameter in Eq.~\ref{eq2}, and this results in two variants: 7) Set $\lambda_{1}=1$, $\lambda_{2}=1$, $\lambda_{3}=1$, and $\lambda_{4}=1$ in Eq.~\ref{eq2} (Hyper-param. \#1); 8) Set $\lambda_{1}=5$, $\lambda_{2}=5$, $\lambda_{3}=10$, and $\lambda_{4}=10$ in Eq.~\ref{eq2} (Hyper-param. \#2). Furthermore, for evaluating the influence of different foreground estimation methods, we use MLFgE~\cite{germer2021fast} and Blur-Fusion~\cite{forte2021approximate} instead of Closed-form Matting~\cite{levin2007closed} for the foreground estimation respectively, resulting in two variants: 9) FG w/~\cite{germer2021fast} and 10) FG w/~\cite{forte2021approximate}. 11) We also use the GT matte instead of enhanced matte for the foreground estimation used in Eq.~\ref{fg_est}, and name this variant as (FG w/ GT $\alpha$).}

We take IndexNet as the reference matting model (also served as the baseline). The quantitative comparisons are shown in Tab.~\ref{table:2} and Tab.~\ref{table:3}. We can see that every variant outperforms the baseline on all metrics. This indicates the proposed idea is effective, even is conducted in one latent space with incomplete losses. Meanwhile, the performances of variants~($w^*,n$) and ($w,n^*$) show that the latent code~$w$ contributes more than code~$n$ to enhancement, and optimizing both latent codes~$w$~and~$n$ significantly boost the performance.

Regarding the loss functions, it is unsurprising that the proposed entropy minimization loss ($\mathcal{L}_{em}$) contributes the most to the improvement. This is because $\mathcal{L}_{em}$ is the core component of the latent codes searching process that modifies uncertain and highly-difficult regions. Meanwhile, we observe that deactivating portrait consistency loss (w/o~$\mathcal{L}_{pc}$) and perceptual loss (w/o~$\mathcal{L}_{pp}$) decrease the final performance slightly. Although these two losses are introduced to maintain the fidelity of the original portrait, the input GAN inverted latent codes are able to generate similar portraits. As a result, using one of these losses is sufficient to ensure the fidelity of the original input. \rev{Moreover, both variants (Hyper-param. \#1) and (Hyper-param. \#2) present similar performances with the complete model ($w^*,n^*$), which shows our model is not sensitive to hyper-parameters in Eq.~\ref{eq2}. For the variants (FG w/~\cite{germer2021fast}) and (FG w/~\cite{forte2021approximate}), they report similar results to the complete model ($w^*,n^*$), which indicates that our method is not sensitive to the foreground estimation methods. The variant (FG w/ GT $\alpha$) presents a similar performance with foreground estimation using the enhanced matte ($w^*,n^*$). That shows our method is not restricted by the quality of the foreground layer. It is because 1) compared with the entropy minimization loss, the compositional adversarial loss contributes less to the improvement, and 2) the quality of the composited sample is mainly influenced by the matting quality. For example, composing a GT foreground with a discontinuous alpha matte (like holes) will produce a bad composition quality. Therefore, rather than noticing the insignificant differences in the foreground estimation, the discriminator can easily classify the discontinuity and provides feedback for training a better alpha matte.}

\subsection{\rev{Evaluation on Challenging Samples}}

\rev{In this section, we evaluate our method on challenging samples. We first give the evaluation on profile faces. As shown in the left part of Fig.~\ref{fig:result3}, our method also fixes the undesired holes on the trimap, or correct the misclassified background (see the sample at the bottom-left in Fig.~\ref{fig:result3}). This reveals that the proposed method is not limited to the frontal faces but can handle profiles as well.}

\rev{We also present the improvement on those samples that have poor initial performances in the right part of Fig.~\ref{fig:result3}. For example, the portrait at the bottom-right in Fig.~\ref{fig:result3} contains complicated background, and the original trimap has a large erosion around the hat region. Meanwhile, the original matte also presents discontinuous regions. For those low initial quality cases, we can also remove the erosion on the trimap and improves the quality of matte after optimization, which shows that our method can boost the performance of any input portrait samples.}

\vspace{-1mm}
\section{Conclusion, Limitation, and Future Work}
In this paper, we propose an alternative solution for image matting. We demonstrate that small changes in the input image may significantly enhance the matting performance. These small changes are discovered by optimizing the latent vectors in the latent spaces of GAN models. We tailor four losses to obtain the optimal latent vectors, easing the matting difficulty with minor modifications in the portrait. The proposed method enables two applications, portrait enhancement and generation. The first one can enhance a real portrait such that the performance of existing matting models can be largely improved. The second application generates high-confidence pseudo data, augmenting the representation ability of existing matting models.

\rev{The limitation of our method is that we need to optimize the latent code for each image. This process takes around 3 mins and it might limit our practical usage.} Another limitation is that combining both the portrait enhancement and generation cannot yield better performance, and the portrait enhancement still has the best result. This is because the synthesized data is also the enhanced portraits, training with these images cannot beat the optimized latent code using portrait enhancement. In the future, we aim to discover whether there is a general direction exists in the latent space, such that moving along that direction can produce a matting-specific version.

{
\bibliographystyle{ieee_fullname}
\bibliography{egbib}
}




\end{document}